\documentclass[twoside,11pt]{article}

\usepackage[accepted]{melba}

\usepackage{amsmath,amsfonts}
\usepackage[utf8]{inputenc} 
\usepackage[T1]{fontenc}    
\usepackage{url}            
\usepackage{nicefrac}       
\usepackage{microtype}      
\usepackage{graphicx}
\usepackage[dvipsnames]{xcolor}
\usepackage{algorithm}
\usepackage{dsfont}
\usepackage[noend]{algpseudocode}
\usepackage{tabularx,booktabs,subcaption}
\usepackage{multirow}

\usepackage{amsmath,amsfonts,bm}

\newcommand{\norm}[1]{\left\lVert#1\right\rVert}

\DeclareMathOperator*{\gd}{\mathfrak{d}}
\DeclareMathOperator*{\cL}{\mathcal{L}}

\def\Figref#1{Figure~\ref{#1}}

\def\Algref#1{Algorithm~\ref{#1}}

\def\1{\bm{1}}

\def\rx{{\textnormal{x}}}
\def\ry{{\textnormal{y}}}

\def\vtheta{{\bm{\theta}}}
\def\va{{\bm{a}}}

\def\vg{{\bm{g}}}
\def\vh{{\bm{h}}}

\def\vL{{\bm{L}}}

\def\vp{{\bm{p}}}
\def\vq{{\bm{q}}}

\def\vv{{\bm{v}}}

\def\vx{{\bm{x}}}
\def\vy{{\bm{y}}}
\def\vz{{\bm{z}}}

\DeclareMathAlphabet{\mathsfit}{\encodingdefault}{\sfdefault}{m}{sl}
\SetMathAlphabet{\mathsfit}{bold}{\encodingdefault}{\sfdefault}{bx}{n}

\def\sR{{\mathbb{R}}}

\newcommand{\E}{\mathbb{E}}
\newcommand{\V}{\mathbb{V}}

\newcommand{\softmax}{\mathrm{softmax}}
\newcommand{\relu}{\mathrm{ReLU}}

\newcommand{\KL}{D_{\mathrm{KL}}}

\DeclareMathOperator*{\argmax}{arg\,max}

\newtheorem{assumption}{Assumption}[section]

\newcommand{\de}{{\beta}}

\newcolumntype{Y}{>{\centering\arraybackslash}X}
\newcolumntype{C}[1]{>{\centering\let\newline\\\arraybackslash\hspace{0pt}}m{#1}}

\newcounter{oldtocdepth}
\newcommand{\hidefromtoc}{%
  \setcounter{oldtocdepth}{\value{tocdepth}}%
  \addtocontents{toc}{\protect\setcounter{tocdepth}{-10}}%
}
\newcommand{\unhidefromtoc}{%
  \addtocontents{toc}{\protect\setcounter{tocdepth}{\value{oldtocdepth}}}%
}

\melbaheading{2022:019}{https://www.melba-journal.org/papers/2022:019.html}{2022}{1-61}{01/2022}{07/2022}{Fidon et al}{Perinatal, Preterm and Paediatric Image Analysis (PIPPI) 2021 special issue}{Esra Abaci-Turk, Jana Hutter, Roxane Licandro, Christopher Macgowan, Andrew Melbourne}

\ShortHeadings{Distributionally Robust Deep Learning}{Fidon et al}
\firstpageno{1}

\title{Distributionally Robust Deep Learning \\using Hardness Weighted Sampling}

\author{\name Lucas Fidon \email lucas.fidon@kcl.ac.uk\\  
	\addr School of Biomedical Engineering \& Imaging Sciences, King’s College London, UK
	\AND
	\name Michael Aertsen\\
	\addr Department of Radiology, University Hospitals Leuven, Belgium
	\AND
	\name Thomas Deprest\\
	\addr Department of Radiology, University Hospitals Leuven, Belgium
	\AND
	\name Doaa Emam\\
	\addr Department of Obstetrics and Gynaecology, University Hospitals Leuven, Belgium
	\AND
	\name Fr\'ed\'eric Guffens\\
	\addr Department of Radiology, University Hospitals Leuven, Belgium
	\AND
	\name Nada Mufti\\
	\addr School of Biomedical Engineering \& Imaging Sciences, King’s College London, UK
	\AND
	\name Esther Van Elslander\\
	\addr Department of Radiology, University Hospitals Leuven, Belgium
	\AND
	\name Ernst Schwartz\\
	\addr Department of Biomedical Imaging and Image-guided Therapy, Medical University of Vienna, Austria
	\AND
	\name Michael Ebner\\
	\addr School of Biomedical Engineering \& Imaging Sciences, King’s College London, UK
	\AND
	\name Daniela Prayer\\
	\addr Department of Biomedical Imaging and Image-guided Therapy, Medical University of Vienna, Austria
	\AND
	\name Gregor Kasprian\\
	\addr Department of Biomedical Imaging and Image-guided Therapy, Medical University of Vienna, Austria
	\AND
	\name Anna L. David\\
	\addr Institute for Women's Health, University College London, UK
	\AND
	\name Andrew Melbourne\\
	\addr School of Biomedical Engineering \& Imaging Sciences, King’s College London, UK
	\AND
    \name S\'ebastien Ourselin \\
	\addr School of Biomedical Engineering \& Imaging Sciences, King’s College London, UK
	\AND
	\name Jan Deprest\\
	\addr Department of Obstetrics and Gynaecology, University Hospitals Leuven, Belgium
	\AND
	\name Georg Langs\\
	\addr Department of Biomedical Imaging and Image-guided Therapy, Medical University of Vienna, Austria
	\AND
	\name Tom Vercauteren \email tom.vercauteren@kcl.ac.uk \\
	\addr School of Biomedical Engineering \& Imaging Sciences, King’s College London, UK
}

\begin{document}

\maketitle

\hidefromtoc  
\begin{abstract}
    Limiting failures of machine learning systems is of paramount importance for safety-critical applications.
    In order to improve the robustness of machine learning systems,
    Distributionally Robust Optimization (DRO) has been proposed as a generalization of Empirical Risk Minimization (ERM).
    However, its use in deep learning has been severely restricted due to the relative inefficiency of the optimizers available for DRO in comparison to the wide-spread variants of Stochastic Gradient Descent (SGD) optimizers for ERM.
    We propose SGD with hardness weighted sampling, a principled and efficient optimization method for DRO in machine learning that is particularly suited in the context of deep learning.
    Similar to a hard example mining strategy in practice, the proposed algorithm is straightforward to implement and computationally as efficient as SGD-based optimizers used for deep learning, requiring minimal overhead computation.
    In contrast to typical ad hoc hard mining approaches, we prove the convergence of our DRO algorithm for over-parameterized deep learning networks with $\relu$ activation and finite number of layers and parameters.
    Our experiments on fetal brain 3D MRI segmentation and brain tumor segmentation in MRI demonstrate the feasibility and the usefulness of our approach.
    Using our hardness weighted sampling for training a state-of-the-art deep learning pipeline leads to improved robustness to anatomical variabilities in automatic fetal brain 3D MRI segmentation using deep learning and to improved robustness to the image protocol variations in brain tumor segmentation.
	%
	Our code is available at~\url{https://github.com/LucasFidon/HardnessWeightedSampler}.
\end{abstract}


\section{Introduction}

\begin{figure}[tb!]
    \centering
    \includegraphics[width=\textwidth]{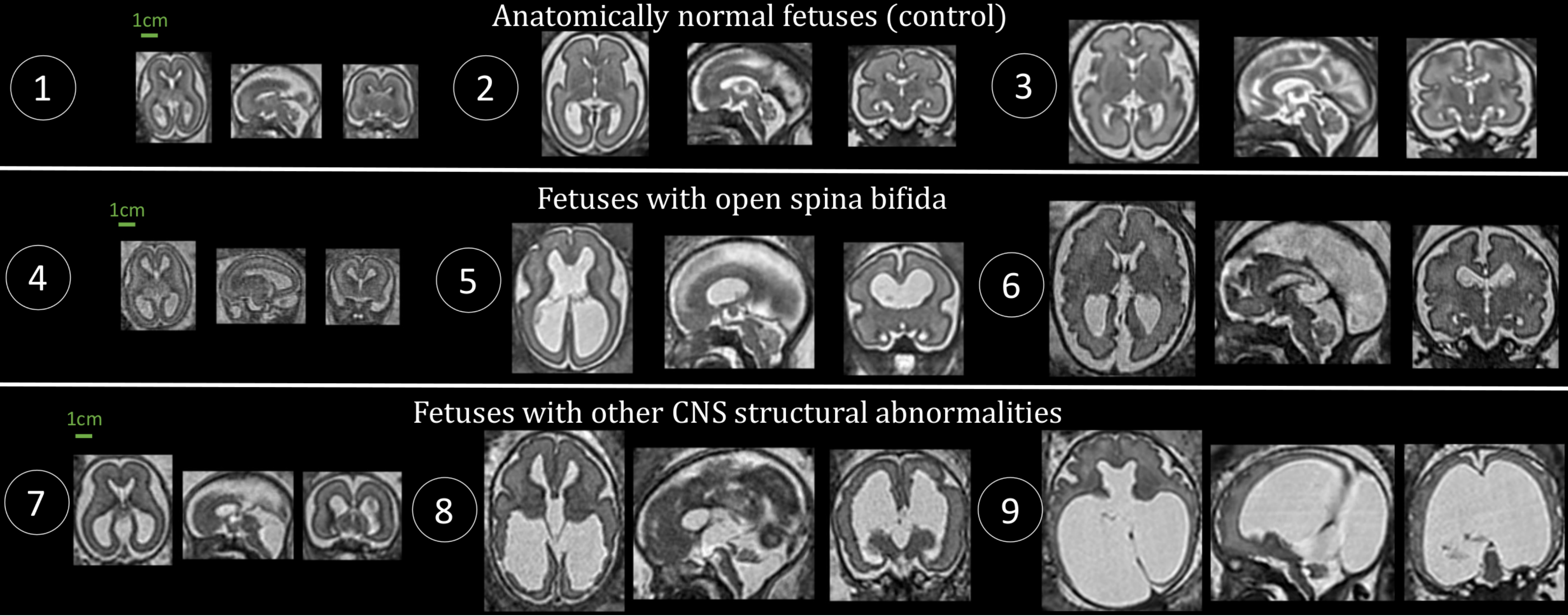}
    \caption{Illustration of the anatomical variability in fetal brain across gestational ages and diagnostics.
    1: Control (22 weeks);  
    2: Control (26 weeks);  
    3: Control (29 weeks);  
    4: Spina bifida (19 weeks);
    5: Spina bifida (26 weeks);
    6: Spina bifida (32 weeks);
    7: Dandy-walker malformation with corpus callosum abnormality (23 weeks);
    8: Dandy-walker malformation with ventriculomegaly and periventricular nodular heterotopia (27 weeks);
    9: Aqueductal stenosis (34 weeks).
    }
    \label{fig:anatomy_variability}
\end{figure}

Datasets used to train deep neural networks typically contain some underrepresented subsets of cases.
These cases are not specifically dealt with by the training algorithms currently used for deep neural networks.
This problem has been referred to as hidden stratification~\citep{oakden2020hidden}.
Hidden stratification has been shown to lead to deep learning models with good average performance but poor performance on underrepresented but clinically relevant subsets of the population~\citep{larrazabal2020gender,oakden2020hidden,puyol2021fairness}.
In \Figref{fig:anatomy_variability} we give an example of hidden stratification in fetal brain MRI. 
The presence of abnormalities associated with diseases with low prevalence~\citep{aertsen2019reliability} exacerbates the anatomical variability of the fetal brain between 18 weeks and 38 weeks of gestation.

While uncovering the issue,
the study of~\cite{oakden2020hidden} does not study the cause or propose a method to mitigate this problem.
In addition, the work of~\cite{oakden2020hidden} is limited to classification.
In standard deep learning pipelines, this hidden stratification is ignored and the model is trained to minimize the mean per-example loss, which corresponds to the standard Empirical Risk Minimization (ERM) problem.
As a result, models trained with ERM are more likely to underperform on those examples from the underrepresented subdomains, seen as \textit{hard examples}.
This may lead to \textit{unfair} AI systems~\citep{larrazabal2020gender,puyol2021fairness}.
For example,
state-of-the-art deep learning models for brain tumor segmentation (currently trained using ERM) underperform for cases with confounding effects, such as low grade gliomas, despite achieving good average and median performance~\citep{bakas2018identifying}.
For safety-critical systems, such as those used in healthcare, this greatly limits their usage as ethics guidelines of regulators such as~\cite{ethics} require AI systems to be technically robust and fair prior to their deployment in hospitals.

Distributionally Robust Optimization (DRO) is a robust generalization of ERM that has been introduced in convex machine learning 
to model the uncertainty in the training data distribution~\citep{chouzenoux2019general,duchi2016statistics,namkoong2016stochastic,rafique2018non}.
Instead of minimizing the mean per-example loss on the training dataset, DRO seeks to optimize for the hardest \emph{weighted} empirical training data distribution around the (uniform) empirical training data distribution.
This suggests a link between DRO and Hard Example Mining.
However, DRO as a generalization of ERM for machine learning still lacks optimization methods that are principled and computationally as efficient as SGD in the non-convex setting of deep learning.
Previously proposed principled optimization methods for DRO consist in alternating between approximate maximization and minimization steps~\citep{jin2019minmax,lin2019gradient,rafique2018non}.
However, they differ from SGD methods for ERM by the introduction of additional hyperparameters for the optimizer such as a second learning rate and a ratio between the number of minimization and maximization steps.
This makes DRO difficult to use as a drop-in replacement for ERM in practice.

In contrast, efficient weighted sampling methods, including Hard Example Mining~\citep{chang2017active,loshchilov2015online,shrivastava2016training} and weighted sampling~\citep{berger2018adaptive,puyol2021fairness}, have been empirically shown to mitigate class imbalance issues and to improve deep embedding learning~\citep{harwood2017smart,suh2019stochastic,wu2017sampling}.
However, even though these works typically start from an ERM formulation, it is not clear how those heuristics formally relate to ERM in theory.
This suggests that bridging the gap between DRO and weighted sampling methods could lead to a principled Hard Example Mining approach, or conversely to more efficient optimization methods for DRO in deep learning.

Given an efficient solver for the inner maximization problem in DRO, DRO could be addressed by maintaining a solution of the inner maximization problem and using a minimization scheme akin to the standard ERM but over an adaptively weighted empirical distribution.
However, even in the case where a closed-form solution is available for the inner maximization problem, it would require performing a forward pass over the entire training dataset at each iteration. This cannot be done efficiently for large datasets.
This suggests identifying an approximate, but practically usable, solution for the inner maximization problem based on a closed-form solution.

From a theoretical perspective, analysis of previous optimization methods for non-convex DRO~\citep{jin2019minmax,lin2019gradient,rafique2018non} made the assumption that the model is either smooth or weakly-convex, but none of those properties are true for deep neural networks with $\relu$ activation functions that are typically used.

In this work, we propose SGD with \textit{hardness weighted sampling}, a novel, principled optimization method for training deep neural networks with DRO and inspired by Hard Example Mining, that is computationally as efficient as SGD for ERM.
Compared to SGD, our method only requires introducing an additional $\softmax$ layer and maintaining a stale per-example loss vector to compute sampling probabilities over the training data.
This work is an extension of our previous preliminary work~\citep{fidon2021distributionally} in which we applied the proposed \textit{hardnes weighted sampler} to distributionally robust fetal brain 3D MRI segmentation and studied the link between DRO and the minimization of percentiles of the per-example loss.
In this extension, we formally introduce our \textit{hardness weighted sampler} and we generalize recent results in the convergence theory of SGD with ERM and over-parameterized deep learning networks with $\relu$ activation functions~\citep{allen2018convergence,allen-zhu19a,cao2019generalization,zou2019improved} to our SGD with hardness weighted sampling for DRO.
This is, to the best of our knowledge, the first convergence result for deep learning networks with $\relu$ trained with DRO.
We also formally link DRO in our method with Hard Example Mining.
As a result, our method can be seen as a principled Hard Example Mining approach.
In terms of experiments, we have extended the evaluation on fetal brain 3D MRI with $69$ additional fetal brain 3D MRIs. We have also added experiments on brain tumor segmentations and experiments on image classification with MNIST as a toy example.
We show that our method outperforms plain SGD in the case of class imbalance, and improves the robustness of a state-of-the-art deep learning pipeline for fetal brain segentation and brain tumor segmentation.
We evaluate the proposed methodology for the automatic segmentation of white matter, ventricles, and cerebellum based on fetal brain 3D T2w MRI.
We used a total of $437$ fetal brain 3D MRIs including anatomically normal fetuses, fetuses with spina bifida aperta, and fetuses with other central nervous system pathologies for gestational ages ranging from $19$ weeks to $40$ weeks.
Our empirical results suggest that the proposed training method based on distributionally robust optimization leads to better percentiles values for abnormal fetuses.
In addition, qualitative results shows that distributionally robust optimization allows to reduce the number of clinically relevant failures of nnU-Net.
For brain tumor segmentation our DRO-based method allows reducing the interquartile range of the Dice scores of $2\%$ for the segmentation of the enhancing tumor and the tumor core regions.

\subsection{Main Mathematical Notations}\label{s:main_notations}
We summarize here the main mathematical notations.
An extended list of notations can be found in Appendix~\ref{s:notations}.
\begin{itemize}
    \item Training dataset: $\{(\vx_i, \vy_i)\}_{i=1}^n$.
    \item $\Delta_n = \left\{\left(p_i\right)_{i=1}^n \in [0,1]^n, \,\,
        \sum_i p_i = 1\right\}$ is a $n$-simplex.
    \item Let $\vq=(q_i) \in \Delta_n$, and $f$ a function, we denote $\E_{\vq}[f(\vx)]:=\sum_{i=1}^n q_i f(\vx_i)$.
    \item Let $\vq \in \Delta_n$, and $f$ a function, we denote $\V_{\vq}[f(\vx)]:=\sum_{i=1}^n q_i\norm{f(\vx_i) - \E_q[f(\vx)]}^2$.
    \item $\vp_{\rm{train}}$ is the uniform training data distribution, i.e. $\vp_{\rm{train}}=\left(\frac{1}{n}\right)_{i=1}^n \in \Delta_n$.
    \item $\cL$ is the per-example loss function.
    \item ERM is short for Empirical Risk Minimization.
    \item DRO is short for Distributionally Robust Optimisation.
\end{itemize}

\section{Related Works}

An optimization method for group-DRO was proposed in~\citep{sagawa2019distributionally}.
In contrast to the formulation of DRO that we study in this paper, their method requires additional labels allowing to identify the underrepresented group in the training dataset.
However, those labels may not be available or may even be impossible to obtain in most applications.
\cite{sagawa2019distributionally} show that, when associated with strong regularization of the weights of the network, their group DRO method can tackle spurious correlations that are known a priori in some classification problems.
It is worth noting that, in contrast, no regularization was necessary in our experiments with MNIST.

Biases of convolutional neural networks applied to medical image classification and segmentation has been studied in the literature.
State-of-the-art deep neural networks for brain tumor segmentation underperform for cases with confounding effects, such as low grade gliomas~\citep{bakas2018identifying}.
It has been shown that scans coming from $15$ different studies can be re-assigned with $73.3\%$ accuracy to their source using a random forest classifier~\citep{wachinger2019quantifying}.
A state-of-the-art deep neural networks for the diagnosis of $14$ thoracic diseases using X-ray trained on a dataset with a gender bias underperform on X-ray of female patients~\citep{larrazabal2020gender}.
And a state-of-the-art deep learning pipeline for cardiac MRI segmentation was found to underperform when evaluated on racial groups that were underrepresented in the training dataset~\citep{puyol2021fairness}.
To mitigate this problem, \cite{puyol2021fairness} proposed to use a stratified batch sampling approach during training that shares similarities with the group-DRO approach mentioned above~\citep{sagawa2019distributionally}.
In contrast to our hardness weighted sampler, their stratified batch sampling approach requires additional labels, such as the racial group, that may not be available for training data. In addition, they do not study the formal relationship between the use of their stratified batch sampling approach and the training optimization problem.

In this work, we focus on DRO with a $\phi$-divergence~\citep{csiszar2004information}.
In this case, the data distributions that are considered in the DRO problem~\eqref{eq:dro} are restricted to sharing the support of the empirical training distribution. In other words, the weights assigned to the training data can change, but the training data itself remains unchanged.
Another popular formulation is DRO with a Wasserstein distance~\citep{chouzenoux2019general,duchi2016statistics,sinha2017certifying,staib2017distributionally}.
In contrast to $\phi$-divergences, using a Wasserstein distance in DRO seeks to apply small data augmentation to the training data to make the deep learning model robust to small deformation of the data, but the sampling weights of the training data distribution typically remains unchanged.
In this sense, DRO with a $\phi$-divergence and DRO with a Wasserstein distance can be considered as orthogonal endeavours.
While we show that DRO with $\phi$-divergence can be seen as a principled Hard Exemple Mining method, it has been shown that DRO with a Wasserstein distance can be seen as a principled adversarial training method~\citep{sinha2017certifying,staib2017distributionally}.

The effect of \textit{multiplicative weighting} during training, rather than \textit{weighted sampling} used in our algorithm, has been studied empirically by
\citep{byrd2019effect} for image classification.
They find that the effect of multiplicative weighting vanishes over training for classification tasks in which we can achieve zero loss on the training dataset.
However, \textit{multiplicative weighting} and \textit{weighted sampling} affect the optimization dynamic in different ways. This may explain why we did not observe this vanishing effect in our experiments on classification and segmentation.
Previous work have also studied empirical and convergence results 
of DRO for linear models~\citep{hu2018does}.

\section{Methods}
\subsection{Background: Deep Learning with Distributionally Robust Optimization}

Standard training procedures in machine learning are based on Empirical Risk Minimization (ERM)~\citep{bottou2018optimization}.
For a neural network $h$ with parameters $\vtheta$, a per-example loss function $\cL$, and a training dataset $\left\{(\vx_i, \vy_i)\right\}_{i=1}^n$, where $\vx_i$ are the inputs and $\vy_i$ are the labels, 
the ERM problem corresponds to
\begin{equation}
    \label{eq:erm_intro}
    \min_{\vtheta} 
    \left\{
    \E_{\textbf{p}_{\rm{train}}} \left[\cL\left(h(\vx; \vtheta), \vy\right)\right]
    = \frac{1}{n} \sum_{i=1}^n \cL \left(h(\vx_i;\vtheta), \vy_i\right)
    \right\}
\end{equation}
where $\textbf{p}_{\rm{train}}$ is the empirical uniform distribution on the training dataset and $\E_{\textbf{p}_{\rm{train}}}$ is the expected value operator as defined in section \ref{s:main_notations}.
When data augmentation is used, the number of samples $n$ can become infinite.
%
For our theoretical results, we suppose that $\textbf{p}_{\rm{train}}$ contains a finite number of examples.
The extension of our \Algref{alg:1} to an infinite number of data augmentations using importance sampling is presented in section~\ref{sec:algo_DRO}.
Optionally, $\cL$ can contain a parameter regularization term that is only a function of $\vtheta$.

The ERM training formulation assumes that $\textbf{p}_{\rm{train}}$ is an unbiased approximation of the true data distribution.
However, this is generally impossible in domains such as medical image computing. This makes models trained with ERM at risk of underperforming on images from parts of the data distribution that are underrepresented in the training dataset.

In contrast, Distributionally Robust Optimization (DRO) is a family of generalization of ERM in which the uncertainty in the training data distribution is modelled by minimizing the worst-case expected loss over an \textit{uncertainty set} of training data distributions~\citep{rahimian2019distributionally}.

In this paper, we consider training deep neural networks with DRO based on a $\phi$-divergence.
We denote 
$\Delta_n := \left\{\left(p_i\right)_{i=1}^n \in [0,1]^n \,\, | \,\,
\sum_{i=1}^n p_i = 1\right\}$
the set of empirical training data probabilities vectors under consideration (i.e. the uncertainty set).
The different probabilities vectors in $\Delta_n$ correspond to all the possible weighting of the training dataset. Every $\textbf{p}=(p_i)_{i=1}^n$ in $\Delta_n$ gives a weight to each training example but keep the examples the same.
We use the following definition of $\phi$-divergence in the remainder of the paper.
%
\begin{definition}[Strong Convexity]
\label{def:strongly_convex}
Let $f : \Omega \rightarrow \sR$ be differentiable on $\Omega$, a convex subset of $\sR$ and $f'$ be the first derivative of $f$.
Let $\rho > 0$, $f$ is $\rho$-strongly convex if for all $x, y \in \Omega,$\\ 
$\phi(y) \geq \phi(x) + \phi'(x)(y - x) + \frac{\rho}{2}(y - x)^2$.
\end{definition}
%
\begin{definition}[$\phi$-divergence]
\label{def:phi_divergence}
Let $\phi : \sR_+ \rightarrow \sR$ 
be two times continuously differentiable on $[0, n]$, $\rho$-strongly convex on $[0, n]$ with $\rho > 0$,
and satisfying
     $\forall z \in \sR,\,\, \phi(z) \geq \phi(1)=0,\,\, \phi'(1) = 0$.
The $\phi$-divergence $D_{\phi}$ is defined as,
for all $\textbf{p}=(p_i)_{i=1}^n, \textbf{q}=(q_i)_{i=1}^n \in \Delta_n$,
\begin{equation}
    D_{\phi}\left(\textbf{q} \Vert \textbf{p}\right)
    = \sum_{i=1}^n p_i \phi\left(\frac{q_i}{p_i}\right)
\end{equation}
\end{definition}
We refer to our example~\ref{ex:softmax} on page~\pageref{ex:softmax} to highlight that the KL divergence is indeed a $\phi$-divergence.

The DRO problem for which we propose an optimizer for training deep neural networks can be formally defined as
\begin{equation}
        \label{eq:dro}
        \min_{\vtheta}\,
        \left\{
        R\left(\vL(h(\vtheta))\right) :=
        \max_{\vq \in \Delta_n}
        \left(
        \E_{\vq} \left[\cL\left(h(\vx; \vtheta), \vy\right)\right]
        - \frac{1}{\de} D_{\phi}\left(\vq \Vert \textbf{p}_{\rm{train}}\right)
        \right)
        \right\}
\end{equation}
where $\textbf{p}_{\rm{train}}$ is the uniform empirical distribution, and $\de > 0$ an hyperparameter.
The choice of $\de$ and $\phi$ controls how the unknown training data distribution $q$ is allowed to differ from $\textbf{p}_{\rm{train}}$.
Here and thereafter, we use the notation 
$\vL(h(\vtheta)) := \left(\cL(h(\vx_i;\vtheta), \vy_i)\right)_{i=1}^n$
to refer to the vector of loss values of the $n$ training samples for the value $\vtheta$ of the parameters of the neural network $h$.
In the remainder of the paper, we will refer to $R$ as the \textit{distributionally robust loss}.

Our analysis of the properties of $R$ in the next sections relies on the Fenchel duality~\citep{moreau1965proximite} and the notion of Fenchel conjugate~\citep{fenchel1949conjugate}.
\begin{definition}[Fenchel Conjugate Function]
    \label{def:fenchel_conjugate}
    Let $f: \sR^m \rightarrow \sR \cup \{+\infty\}$ be a proper function. The Fenchel conjugate of $f$ is defined as
    $\forall \vv \in \sR^m,\,\, f^*(\vv)= \sup_{\vx \in \sR^m} \langle \vv, \vx\rangle - f(\vx)$
    where $\langle \cdot,\cdot\rangle$ is the inner product.
\end{definition}
\subsection{Hardness Weighted Sampling for Distributionally Robust Deep Learning}

In the case where $h$ is a non-convex predictor (such as a deep neural network), existing optimization methods for the DRO problem \eqref{eq:dro} alternate between approximate minimization and maximization steps \citep{jin2019minmax,lin2019gradient,rafique2018non}, requiring the introduction of additional hyperparameters compared to SGD. However, these are difficult to tune in practice and convergence has not been proven for non-smooth deep neural networks such as those with $\relu$ activation functions.

In this section, we present an SGD-like optimization method for training a deep learning model $h$ with the DRO problem \eqref{eq:dro}.
We first highlight, in Section~\ref{sec:sampling_approach_DRO}, mathematical properties that allow us to link DRO with stochastic gradient descent (SGD) combined with an adaptive sampling that we refer to as \textit{hardness weighted sampling}.
In Section~\ref{sec:algo_DRO}, we present our \Algref{alg:1} for distributionally robust deep learning.
Then, in Section~\ref{sec:theoretical_results}, we present theoretical convergence results for our hardness weighted sampling.

\subsubsection{A sampling approach to Distributionally Robust Optimization}\label{sec:sampling_approach_DRO}
The goal of this subsection is to show that a stochastic approximation of the gradient of the \textit{distributionally robust loss} can be obtained by using a weighted sampler.
This result is a first step towards our \Algref{alg:1} for efficient training with the \textit{distributionally robust loss} presented in the next subsection.


To reformulate $R$ as an unconstrained optimization problem over $\sR^n$ (rather than constraining it \textcolor{black}{to the $n$-simplex} $\Delta_n$), we define
\begin{equation}
    \label{eq:G_def}
    \begin{aligned}
    \forall \textbf{p} \in \sR^n,\quad 
             G(\textbf{p}) = \frac{1}{\de} D_{\phi}(\textbf{p} \Vert \textbf{p}_{\rm{train}}) + \delta_{\Delta_n}(\textbf{p})
    \end{aligned}
\end{equation}
where $\delta_{\Delta_n}$ is the characteristic function of the \textcolor{black}{to the $n$-simplex $\Delta_n$ which is a} closed convex set, i.e.
\begin{equation}
    \forall \textbf{p} \in \sR^n,\quad \delta_{\Delta_n}(\textbf{p})=\left\{
\begin{array}{cl}
    0 & \text{if } \textbf{p} \in \Delta_n \\
    +\infty & \text{otherwise}
\end{array}
\right.
\end{equation}
The distributionally robust loss $R$ in \eqref{eq:dro} can now be rewritten using the Fenchel conjugate function $G^*$ of $G$.
This allows us to obtain regularity properties for $R$.
\begin{lemma}[Regularity of $R$]
\label{lemma:R_property}
    If $\phi$ satisfies Definition \ref{def:phi_divergence} 
    (i.e. can be used for a $\phi$-divergence), then $G$ and $R$ satisfy the following:
    \begin{equation}
        \label{eq:strong_convexity_G}
        G \text{ is} \left(\frac{n\rho}{\de}\right)\text{-strongly convex}
    \end{equation}
    \begin{equation}
        \label{eq:link_R_and_G}
        \forall \vtheta, \quad 
        R(\vL(h(\vtheta)))
        =
        \max_{\textbf{q} \in \sR^n} 
            \Big(
            \langle \vL(h(\vtheta)), \textbf{q}\rangle - G(\textbf{q})
            \Big)
        = G^*\left(\vL(h(\vtheta))\right)
    \end{equation}
    \begin{equation}
        \label{eq:gradient_Lip_R}
        R \text{ is } \left(\frac{\de}{n\rho}\right)\text{-gradient Lipschitz continuous.}
    \end{equation}
\end{lemma}
Equation \eqref{eq:link_R_and_G} follows from Definition \ref{def:fenchel_conjugate}. Proofs of \eqref{eq:strong_convexity_G} and \eqref{eq:gradient_Lip_R} can be found in Appendix~\ref{s:regularity_R}.
According to \eqref{eq:strong_convexity_G}, the optimization problem \eqref{eq:link_R_and_G} is strictly convex and admits a unique solution in $\Delta_n$, which we denote as
\begin{equation}
    \label{eq:hardness_weighted_proba}
    \bar{\textbf{p}}(\vL(h(\vtheta))) = \argmax_{\textbf{q} \in \sR^n} 
        \left(
        \langle \vL(h(\vtheta)), \textbf{q}\rangle - G(\textbf{q})
        \right)
\end{equation}

Thanks to those properties, we can now show the following lemma that is essential for the theoretical foundation of our \Algref{alg:1}.
Equation \eqref{eq:p} states that the gradient of the distributionally robust loss $R$ is a weighted sum of the the gradients of the per-example losses (i.e. the gradients computed by the backpropagation algorithm in deep learning)
with the weights given by the empirical distribution $\bar{\textbf{p}}(\vL(\vh(\vtheta)))$.
We further show that straightforward analytical formulas exist for $\bar{\textbf{p}}$, and give an example of such probability distribution for the Kullback-Leibler (KL) divergence.
\begin{lemma}[Stochastic Gradient of the Distributionally Robust Loss]
\label{lemma:robust_loss_sg}
For all $\vtheta$, we have
\begin{equation}
    \label{eq:p}
    \begin{aligned}
        \nabla_\vtheta (R \circ \vL \circ ~h)(\vtheta) 
        & = \E_{\textcolor{black}{\bar{\textbf{p}}(\vL(h(\vtheta)))}}\left[\nabla_\vtheta \cL\left(h(\vx; \vtheta), \vy\right)\right]
    \end{aligned}
\end{equation}
\end{lemma}
The proof is found in Appendix~\ref{s:robust_loss_sg}.
We now provide a closed-form formula for $\bar{\textbf{p}}$ given $\cL(h(\vtheta))$ for the KL divergence as the choice of $\phi$-divergence. 
%
\begin{example}
    \label{ex:softmax}
    For $\phi: z \mapsto z\log(z) - z + 1$, $D_{\phi}$ is the Kullback-Leibler (KL) divergence:
    \begin{equation}
        D_{\phi}(\textbf{q} \Vert \textbf{p}) = \KL (\textbf{q} \Vert \textbf{p}) = \sum_{i=1}^n q_i\log\left(\frac{q_i}{p_i}\right)
    \end{equation}
    In this case, we have (see Appendix~\ref{s:proof_softmax} for a proof)
    \begin{equation}\label{eq:KL_softmax}
        \bar{\textbf{p}}(\vL(h(\vtheta))) 
        = \softmax\left(\de \vL(h(\vtheta))\right)
    \end{equation}
\end{example}

\subsubsection{Proposed Efficient Algorithm for Distributionally Robust Deep Learning}\label{sec:algo_DRO}
We now describe our algorithm for training deep neural networks with DRO using our hardness weighted sampling.

\begin{algorithm}[htb!]
\caption{Training procedure for DRO with Hardness Weighted Sampling.
\textcolor{black}{Additional operations as compared to standard training algorithms are highlighted in} \textcolor{blue}{blue}.
}
\label{alg:1}
\begin{algorithmic}[1]
\Require{$\left\{(\vx_i, \vy_i)\right\}_{i=1}^n$: training dataset with $n>0$ the number of training samples.}
\Require{$b\in \{1,\ldots, n\}$: batch size.}
\Require{$\cL$: (any) smooth per-example loss 
function \textcolor{black}{(e.g. cross entropy loss, Dice loss)}}.
\Require{$\de > 0$: robustness parameter defining the distributionally robust optimization problem.}
\Require{$\vtheta_{0}$: initial parameter vector for the model $h$ to train.}
\Require{$\vL_{init}$: initial stale per-example loss values vector.}
\State{$t \leftarrow 0$}\Comment{\textcolor{black}{initialize the time step}}
\State{\textcolor{blue}{$\vL \leftarrow \vL_{init}$}}\Comment{\textcolor{black}{initialize the vector of stale loss values}}
\While{$\vtheta_t$ has not converged}
    \State{\textcolor{blue}{$\vp_t \leftarrow \softmax(\de \vL)$}
    }\Comment{online estimation of the \textit{hardness weights}}
    
    \State{\textcolor{blue}{$I \sim \vp_t$}
    }\Comment{hardness weighted sampling}
    
    \If{importance sampling is not used}
    \State{
    \textcolor{blue}{$\forall i \in I,\,\, w_i = 1$}
    }
    \Else{}
    \State{
    \textcolor{blue}{
    $\forall i \in I,\,\,
    w_{i} \leftarrow 
    \exp\left(\de(\cL(h(\vx_i;\vtheta), \vy_i) - L_{i})\right)$
    }
    }\Comment{importance sampling weights}
    \State{
    \textcolor{blue}{
    $\forall i \in I,\,\,
    w_{i} \leftarrow \textup{clip}\left(w_i, [w_{min}, w_{max}]\right)$
    }
    }\Comment{clip the weights for stability}
    \EndIf
    
    \State{
    \textcolor{blue}{
    $\forall i \in I,\,\, 
    L_i \leftarrow \cL(h(\vx_i;\vtheta), \vy_i)$
    }
    }\Comment{update the vector of stale loss values}
    
    \State{$\vg_t \leftarrow \frac{1}{b}\sum_{i \in I}
    \textcolor{blue}{ w_i}
    \nabla_{\vtheta}\cL(h(\vx_i;\vtheta_{t}), \vy_i)$ 
    }
    
    \State{$\vtheta_{t+1} \leftarrow \vtheta_{t} 
    - \eta \,\vg_t$}\Comment{\textcolor{black}{SGD step or any other optimizer (e.g. SGD momentum, Adam)}}
\EndWhile
\State{\textbf{Output:} $\vtheta_t$}
\end{algorithmic}
\end{algorithm}

Equation \eqref{eq:p} implies that $\nabla_\vtheta \cL(h(\vx_i;\vtheta), \vy_i)$ is an unbiased estimator of the gradient of the distributionally robust loss gradient when $i$ is sampled with respect to $\bar{\textbf{p}}(\vL(h(\vtheta)))$.
This suggests that the distributionally robust loss can be minimized efficiently by SGD by sampling mini-batches with respect to $\bar{\textbf{p}}(\vL(h(\vtheta)))$ at each iteration.
However, even though closed-form formulas were provided in Example~\ref{ex:softmax} for $\bar{\textbf{p}}$,
evaluating exactly $\vL(h(\vtheta))$, i.e. doing one forward pass on the whole training dataset at each iteration, is computationally prohibitive for large training datasets.

In practice, we propose to use a stale version of the vector of per-example loss values
by maintaining an online history of the loss values of the examples seen during training $\left(\cL(h(\vx_i;\vtheta^{(t_i)}), \vy_i)\right)_{i=1}^n$,
where for all $i$, $t_i$ is the last iteration at which the per-example loss of example $i$ has been computed.
Using the Kullback-Leibler divergence as $\phi$-divergence, this leads to the SGD with hardness weighted sampling algorithm proposed in \Algref{alg:1}.

When data augmentation is used, an infinite number of training examples is virtually available.
In this case, we keep one stale loss value per example irrespective of any augmentation as an approximation of the loss for this example under any augmentation.

Importance sampling is often used when sampling with respect to a desired distribution cannot be done exactly~\citep{kahn1953methods}.
In \Algref{alg:1}, an up-to-date estimation of the per-example losses (or equivalently the hardness weights) in a batch is only available \emph{after} sampling and evaluation through the network. Importance sampling can be used to compensate for the difference between the initial and the updated stale losses within this batch.
We propose to use importance sampling in steps 9-10 of \Algref{alg:1} and highlight that this is especially useful to deal with data augmentation. Indeed, in this case, the stale losses for the examples in the batch are expected to be less accurate as they were estimated under a different augmentation.
For efficiency, we use the following approximation
$
w_i = \frac{p_i^{new}}{p_i^{old}} \approx \exp\left(\de(\cL(h(\vx_i;\vtheta), \vy_i) - L_{i})\right)
$
where we have neglected the typically small change in the denominator of the $\softmax$. More details are given in Appendix~\ref{appendix:importance_sampling}.
%
To tackle the typical instabilities that can arise when using importance sampling~\citep{owen2000safe}, the importance weights are clipped.

%
Compared to standard SGD-based training optimizers for the mean loss, our algorithm requires only an additional $\softmax$ operation per iteration and to store an additional vector of scalars of size $n$ (number of training examples), thereby making it well suited for deep learning applications.
\textcolor{black}{
The computational time and memory overheads are studied in section~\ref{sec:efficiency}.
}

\textcolor{black}{
For the convergence theorem, the stopping criteria is $\norm{\nabla_{\vtheta} (R\circ \vL \circ h)(\vtheta)} \leq \epsilon$.
However, in our experiments, a fixed number of iterations is used as implemented in the state-of-the-art method nnU-Net~\cite{isensee2021nnu}.
}

\subsection{Overview of Theoretical Results}\label{sec:theoretical_results}
In this section, we present convergence guarantees for \Algref{alg:1} in the framework of over-parameterized deep learning.
We further demonstrate properties of our hardness weighted sampling that allow to clarify its link with Hard Example Mining and with the minimization of percentiles of the per-sample loss on the training data distribution.

\subsubsection{Convergence of SGD with Hardness Weighted Sampling for Over-parameterized Deep Neural Networks with \texorpdfstring{$\relu$}{$\relu$}}\label{s:convergence_main_text}
%
Convergence results for over-parameterized deep learning have recently been proposed in~\citep{allen-zhu19a}.
Their work gives convergence guarantees for deep neural networks $h$ with any activation functions (including $\relu$), and with any (finite) number of layers $L$ and parameters $m$, under the assumption that $m$ is large enough.
In our work, we extend the convergence theory developed by~\citep{allen-zhu19a} for ERM and SGD to DRO using the proposed SGD with hardness weighted sampling and stale per-example loss vector (as stated in~\Algref{alg:1}).
The proof in Appendix~\ref{s:proof_stale_loss_history} deals with the challenges raised by the non-linearity of $R$ with respect to the per-sample stale loss and the non-uniform dynamic sampling used in \Algref{alg:1}.

\begin{theorem}[Convergence of \Algref{alg:1} for neural networks with $\relu$]
\label{th:convergence_dro}
Let $\cL$ be a smooth per-example loss function, $b \in \{1,\ldots,n\}$ be the batch size, and $\epsilon > 0$.
If the number of parameters $m$ is large enough, and the learning rate is small enough, then, with high probability over the randomness of the initialization and the mini-batches, \Algref{alg:1} (without importance sampling) guarantees 
$\norm{\nabla_{\vtheta} (R\circ \vL \circ h)(\vtheta)} \leq \epsilon$ after a finite number of iterations.
\end{theorem}

A detailed description of the assumption for this theorem is described in Appendix~\ref{th:conv_sgd_stale_loss_history} and its proof can be found in Appendix~\ref{s:proof_stale_loss_history}.
Our proof does not cover the case where importance sampling is used. However, our empirical results suggest that convergence guarantees still hold with importance sampling.

\subsubsection{Link between Hardness Weighted Sampling and Hard Example Mining}
%
In this section, we discuss the relationship between the proposed hardness weighted sampling for DRO and Hard Example Mining.
The following result shows that using the proposed \textit{hardness weighted sampler} the hard training examples, those training examples with relatively high values of the loss, are sampled with higher probability.

\begin{theorem}
\label{th:hard_example_mining}
Let a $\phi$-divergence that satisfies Definition~\ref{def:phi_divergence}, and $\textbf{L} = \left(L_i\right)_{i=1}^n \in \sR^n$ a vector of loss values for the examples $\{\vx_1, \ldots, \vx_n\}$.
The proposed hardness weighted sampling probabilities vector $\bar{\textbf{p}}\left(\textbf{L}\right) = \left(\bar{p}_i\left(\textbf{L}\right)\right)_{i=1}^n$
defined as in \eqref{eq:hardness_weighted_proba} verifies:
\begin{enumerate}
    \item For all $i \in \{1, \ldots, n\}$, $\bar{p}_i$ is an increasing function of $L_i$.
    \item For all $i \in \{1, \ldots, n\}$, $\bar{p}_i$ is an non-increasing function of any $L_j$ for $j \neq i$.
\end{enumerate}
\end{theorem}
See Appendix~\ref{s:hard_example_mining} for the proof.
The second part of Theorem \ref{th:hard_example_mining} implies that as the loss of an example diminishes, the sampling probabilities of all the other examples increase.
As a result, the proposed SGD with hardness weighted sampling balances exploitation (i.e. sampling the identified \textit{hard examples}) and exploration (i.e. sampling any example to keep the record of \textit{hard examples} up to date).
Heuristics to enforce this trade-off are often used in Hard Example Mining methods~\citep{berger2018adaptive,harwood2017smart,wu2017sampling}.

\subsubsection{Link between DRO and the Minimization of a Loss Percentile}
In this section, we show that the DRO problem \eqref{eq:dro} using the KL divergence is equivalent to a relaxation of the minimization of the per-example loss percentile shown thereafter in equation \eqref{eq:perc}.

Instead of the average per-example loss \eqref{eq:erm_intro}, for robustness,
one might be more interested in minimizing
the
percentile
$l_{\alpha}$ at $\alpha$ (e.g. 5\%)
of the per-example loss function.
Formally, this corresponds to the minimization problem
\begin{equation}
    \label{eq:perc}
        \min_{\vtheta,\, l_{\alpha}} \quad l_{\alpha} \qquad
        \textrm{such that}
        \qquad  
        p_{\rm{train}}\left(
            \cL \left(h(\vx;\vtheta), \vy\right) \geq l_{\alpha}
            \right) \leq \alpha
\end{equation}
where $p_{\rm{train}}$ is the empirical distribution defined by the training dataset.
%
In other words,
if $\alpha=0.05$, the optimal $l_{\alpha}^*(\vtheta)$ of \eqref{eq:perc} for a given value set of parameters $\vtheta$ is the value of the loss such that the per-example loss function is worse than $l_{\alpha}^*(\vtheta)$ $5\%$ of the time.
As a result, training the deep neural network using \eqref{eq:perc} corresponds to minimizing the percentile of the per-example loss function $l_{\alpha}^*(\vtheta)$.

Unfortunately, the minimization problem \eqref{eq:perc} cannot be solved directly using stochastic gradient descent to train a deep neural network.
We now propose a tractable upper bound for $l_{\alpha}^*(\vtheta)$ and show that it can be solved in practice using distributionally robust optimization.

The Chernoff bound~\citep{chernoff1952measure} applied to the per-example loss function and the empirical training data distribution states that for all $l_{\alpha}$ and $\de>0$
\begin{equation}
    p_{\rm{train}}\left(
            \cL \left(h(\vx;\vtheta), \vy\right) \geq l_{\alpha}
            \right) 
    \leq 
        \frac{\exp\left(-\de l_{\alpha}\right)}{n} 
        \sum_{i=1}^n \exp\left(\de \cL \left(h(\vx_i;\vtheta), \vy_i\right)\right)
\end{equation}
To link this inequality to the minimization problem \eqref{eq:perc}, we set $\de>0$ and
\begin{align}
    \hat{l}_{\alpha}(\vtheta) &= \frac{1}{\de} \log\left(
        \frac{1}{\alpha n}
        \sum_{i=1}^n \exp\left(\de \cL \left(h(\vx_i;\vtheta), \vy_i\right)\right)
    \right)
\end{align}
In this case, we have
\begin{equation}
    p_{\rm{train}}\left(
            \cL \left(h(\vx;\vtheta), \vy\right) \geq \hat{l}_{\alpha}(\vtheta)
            \right) 
    \leq \alpha =
        \frac{\exp\left(-\de \hat{l}_{\alpha}(\vtheta)\right)}{n} 
        \sum_{i=1}^n \exp\left(\de \cL \left(h(\vx_i;\vtheta), \vy_i\right)\right)
\end{equation}
$\hat{l}_{\alpha}(\vtheta)$ is therefore an upper bound for the optimal $l^*_{\alpha}(\vtheta)$ in equation \eqref{eq:perc}, independently to the value of $\vtheta$.
Equation \eqref{eq:perc} can therefore be relaxed by 
\begin{equation}
    \label{eq:expvar}
    \min_{\vtheta} \frac{1}{\de} \log\left(
        \sum_{i=1}^n \exp\left(\de \cL \left(h(\vx_i;\vtheta), \vy_i\right)\right)
    \right)
\end{equation}
where $\de>0$ is a hyperparameter,
and where
the term $\frac{1}{\de} \log\left(\frac{1}{\alpha n}\right)$ was dropped as being independent of $\vtheta$.
%
While in \eqref{eq:expvar}, $\alpha$ does not appear in the optimization problem directly anymore, $\de$ essentially acts as a substitute for $\alpha$.
The higher the value of $\de$, the higher weights the per-example losses with a high value will have in \eqref{eq:expvar}.

We give a proof in Appendix~\ref{s:proof_dro_and_percentile}
that \eqref{eq:expvar} is equivalent to solving the following DRO problem
\begin{equation}
    \label{eq:dro2}
    \min_{\vtheta}\, \max_{\vq \in \Delta_n}
        \left(
        \sum_{i=1}^n q_i \cL\left(h(\vx_i; \vtheta), \vy_i\right)
        - \frac{1}{\de} D_{KL}\left(\vq\, \biggr\Vert\, 
        \textcolor{black}{\vp}_{\rm{train}}
        \right)
        \right)
\end{equation}
This is a special case of the DRO problem \eqref{eq:dro} where $\phi$ is chosen as the KL-divergence and it corresponds to the setting of \Algref{alg:1}.

\section{Experiments}
In this section, we experiments with the proposed \textit{hardness weighted sampler} for DRO as implemented in the proposed \Algref{alg:1}.
In the subsection~\ref{s:mnist}, we give a toy example with the task of automatic classification of digits in the case where the digit $3$ is underrepresented in the training dataset.
And in subsection~\ref{s:medical_image_segmentation}, we report the results of our experiments on two medical image segmentation tasks: fetal brain segmentation using 3D MRI, and brain tumor segmentation using 3D MRI.

\subsection{Toy Example: MNIST Classification with a Class Imbalance}\label{s:mnist}
The goal of this subsection is to illustrate key benefits of training a deep neural network using
DRO in comparison to ERM when a part of the sample distribution is underrepresented in the training dataset.
We take the MNIST dataset~\citep{lecun1998mnist} as a toy example, in which the task is to automatically classify images representing digits between $0$ and $9$.
In addition, we verify the ability of our \Algref{alg:1} to train a deep neural network for DRO and illustrates the behaviour of SGD with hardness weighted sampling for different values of $\de$.

\paragraph*{Material:}
We create a bias between training and testing data distribution of MNIST~\citep{lecun1998mnist} by keeping only $1\%$ of the digits $3$ in the training dataset, while the testing dataset remains unchanged.

For our experiments on MNIST, we used a Wide Residual Network (WRN)~\citep{zagoruyko2016wide}.
The family of WRN models has proved to be very efficient and flexible, achieving state-of-the-art accuracy on several dataset.
More specifically, we used WRN-$16$-$1$~\citep[section 2.3]{zagoruyko2016wide}.
For the optimization we used a learning rate of $0.01$.
No momentum or weight decay were used.
No data augmentation was used.
For DRO no importance sampling was used.
We used a GPU NVIDIA GeForce GTX 1070 with 8GB of memory for the experiments on MNIST.

\begin{figure}[bth!]
    \centering
    \includegraphics[width=0.98\linewidth]{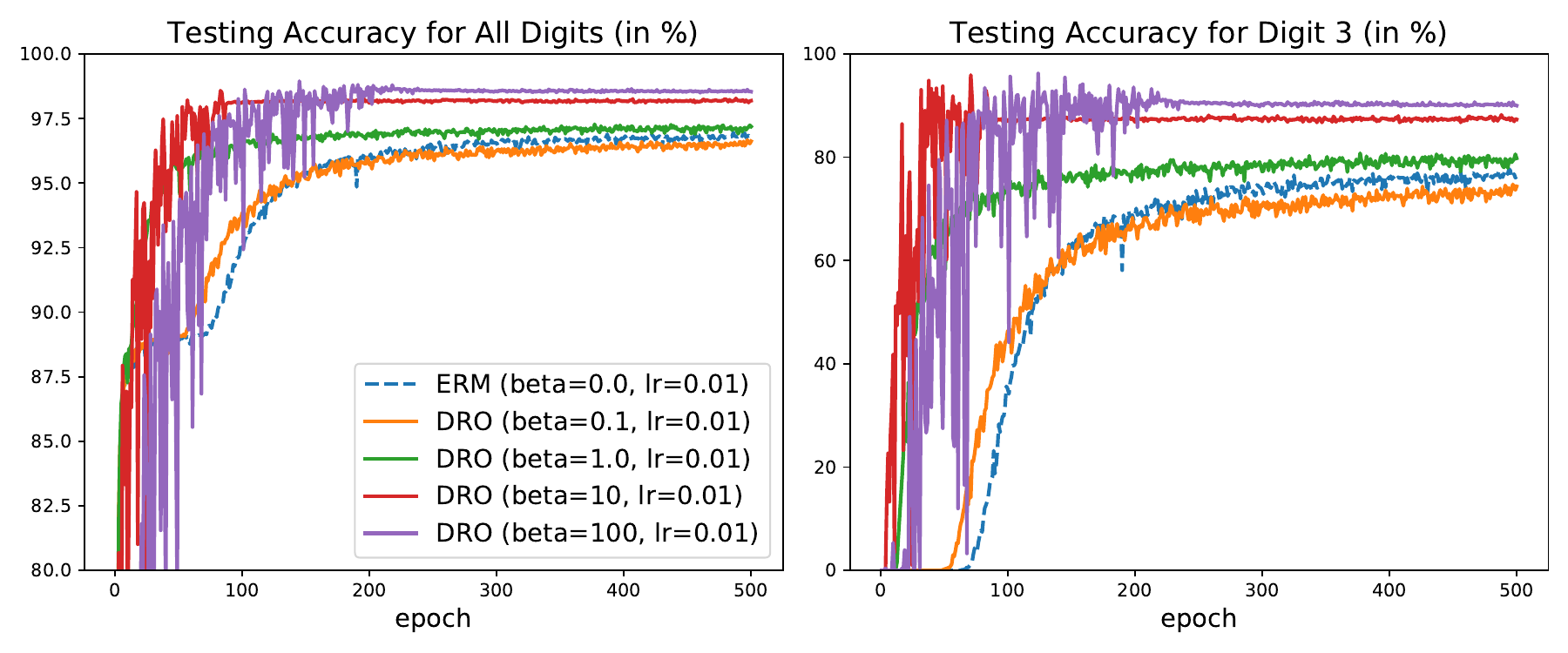}
    \includegraphics[width=0.98\linewidth,trim=0cm 0cm 1cm 0cm]{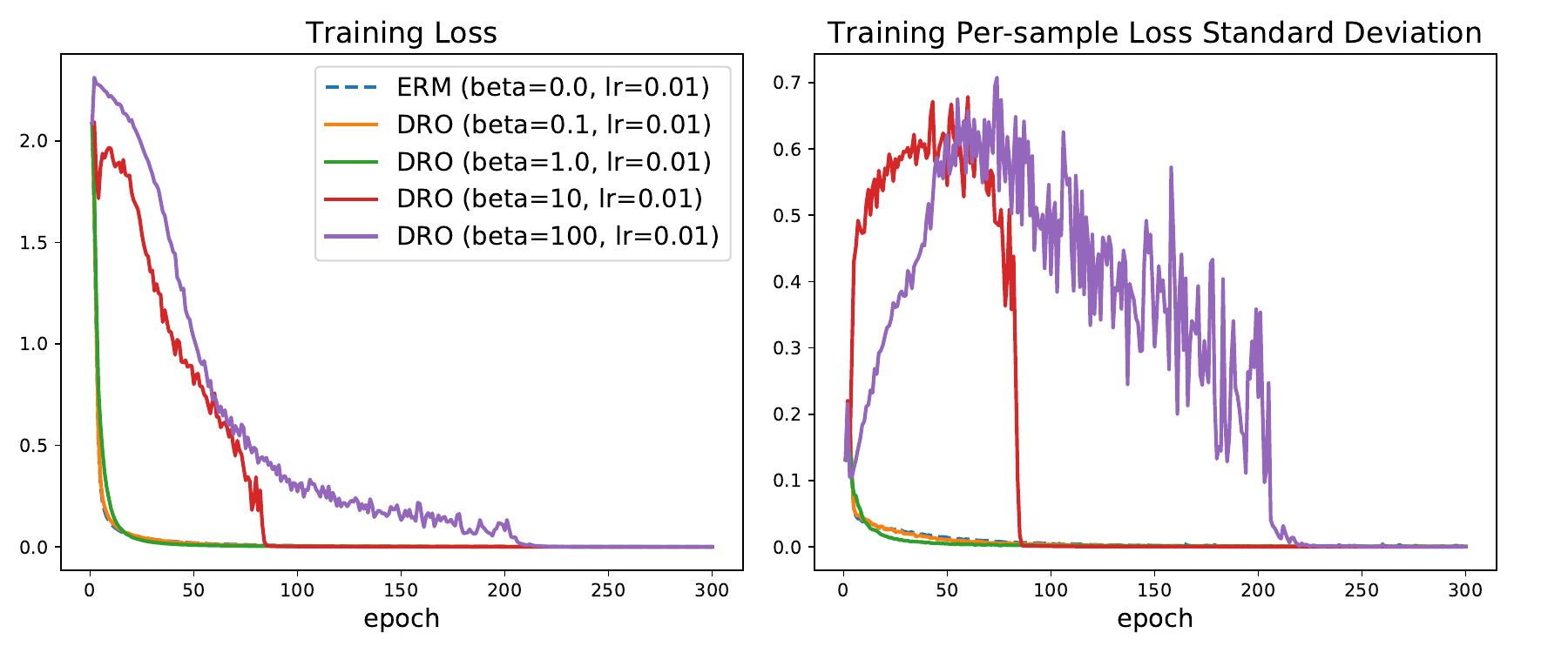}
    \caption{\label{fig:mnist_1percent_all}
        \textbf{Experiments on MNIST.}
        We compare the learning curves at testing (top panels) and at training (bottom panels) for ERM with SGD (\textcolor{blue}{blue}) and DRO with our SGD with hardness weighted sampling for different values of $\de$
        (\textcolor{orange}{$\de=0.1$}, \textcolor{green}{$\de=1$}, \textcolor{red}{$\de=10$},
        \textcolor{violet}{$\de=100$}).
        The models are trained on an imbalanced MNIST dataset (only $1\%$ of the digits $3$ kept for training) and evaluated on the original MNIST testing dataset.
        }
\end{figure}

\paragraph*{Results:}
Our experiment suggests that DRO and ERM lead to different optima.
Indeed, DRO for $\beta=10$ outperforms ERM by more than $15\%$ of accuracy on the underrepresented class, as illustrated in \Figref{fig:mnist_1percent_all}.
This suggests that DRO is more robust than ERM to domain gaps between the training and the testing dataset.
In addition, \Figref{fig:mnist_1percent_all} suggests that DRO with our SGD with hardness weighted sampling can converge faster than ERM with SGD.

Furthermore, the variations of learning curves with $\de$ shown in \Figref{fig:mnist_1percent_all} are consistent with our theoretical insight.
As $\de$ decreases to $0$, the learning curve of DRO with our \Algref{alg:1} converges to the learning curve of ERM with SGD.

For large values of $\de$ (here $\de \geq 10$), instabilities appear before convergence in the \textbf{testing learning curves}, as illustrated in the top panels of \Figref{fig:mnist_1percent_all}.
However, the bottom left panel of \Figref{fig:mnist_1percent_all} shows that the \textbf{training loss curves} for $\de \geq 10$ were stable there. 
We also observe that during iterations where instabilities appear on the \textbf{testing set}, the standard deviation of the per-example loss on the \textbf{training set} is relatively high (i.e. the hardness weighted probability is further away from the uniform distribution).
This suggests that the apparent instabilities on the \textbf{testing set} are related to differences between the distributionally robust loss and the mean loss.
\subsection{Medical Image Segmentation}\label{s:medical_image_segmentation}

In this section, we illustrate the application of \Algref{alg:1} to improve the robustness of deep learning methods for medical image segmentation.
We first discuss the specificities of applying the proposed \textit{hardness weighted sampling} to medical image segmentation in relation to the use of patch-based sampling.
We evaluated the proposed method on two applications:
fetal brain 3D MRI segmentation using the FeTA dataset and a private dataset,
and brain tumor multi-sequence MRI segmentation using the BraTS 2019 dataset~\citep{bakas2017HGG,bakas2017LGG}.

\subsubsection{Hardness Weighted Sampler with Large Images}\label{sec:patch_sampling}
In medical image segmentation, the image used as input of the deep neural network are typically large 3D volumes.
For this reason, state-of-the-art deep learning pipelines use patch-based sampling rather than full-volume sampling during training with ERM ~\citep{isensee2021nnu} as described in subsection~\ref{sec:material}.

This raised the question of what is the training distribution $p_{\rm{train}}$ in the ERM \eqref{eq:erm_intro} and DRO \eqref{eq:dro} optimization problems.
Here, since the patches are large enough to cover most of the brains, we consider that patches are good approximation of the whole volumes and $p_{\rm{train}}$ is the distribution of the full volumes.
Therefore, in the hardness weighted sampler of \Algref{alg:1}, we have only one weight per full volume.

In the case the full volumes are too large to be well covered by the patches, one can divide each full volume into a finite number of subvolumes prior to training.
For example, for chest CT, one can divide the volumes into left and right lungs~\citep{tilborghs2020comparative}.

\subsubsection{Material}\label{sec:material}

\paragraph{Fetal Brain Dataset.}
\begin{table}[bt]
	\centering
	\caption{
	\textbf{Training and Testing Fetal Drain 3D MRI Dataset Details.}
	Other Abn: brain structural abnormalities other than spina bifida.
	There is no overlap of subjects between training and testing.
	}
	\begin{tabularx}{\textwidth}{ *{5}{Y}}
		\toprule
		Train/Test & Origin & Condition & Volumes & Gestational age (in weeks)\\
		\midrule
		Training & Atlas& Control & 18 & [21,\,38]\\
		Training & FeTA & Control & 5 & [22,\,28]\\
		Training & UHL & Control & 116 & [20,\,35]\\
		Training & UHL & Spina Bifida & 28 & [22,\,34]\\
		Training & UHL & Other Abn & 10 & [23,\,35]\\
		\midrule
		Testing & FeTA & Control & 31 & [20,\,34]\\
		Testing & FeTA & Spina Bifida & 38 & [21,\,31]\\
		Testing & FeTA & Other Abn & 16 & [20,\,34]\\
		Testing & UHL & Control & 76 & [22,\,37]\\
		Testing & UHL and MUV & Spina Bifida & 74 & [19,\,35]\\
		Testing & UHL & Other Abn & 25 & [21, 40]\\
	\bottomrule
	\end{tabularx}
	\label{tab:data}
\end{table}
\textcolor{black}{
A total of $177$ (resp. $260$) fetal brain 3D MRIs were used for training (resp. testing).
Origin, condition, and gestational ages for the training and testing datasets are summarized in Table \ref{tab:data}.
}

We used the 18 control fetal brain 3D MRIs of the spatio-temporal fetal brain atlas\footnote{\url{http://crl.med.harvard.edu/research/fetal_brain_atlas/}}~\citep{gholipour2017normative} for gestational ages ranging from $21$ weeks to $38$ weeks.
We also used $80$ volumes from the publicly available FeTA MICCAI challenge dataset\footnote{DOI: 10.7303/syn25649159}~\citep{payette2021automatic,payette2022fetal}
and the $10$ 3D MRIs from the testing set of the first release of the FeTA dataset for which manual segmentations are not publicly available.
For those 3D MRIs, manual segmentations and corrections of the segmentations were performed by authors MA and LF to reduce the variability against the published segmentation guidelines that was released with the FeTA dataset~\citep{payette2021automatic}.
Part of those corrections were performed as part of our previous work~\citep{fidon2021label,fidon2021partial} and are publicly available\footnote{DOI: 10.5281/zenodo.5148611}.
Brain masks for the FeTA data were obtained via affine registration using two fetal brain atlases\footnote{DOI: 10.7303/syn25887675}~\citep{fidon2021atlas,gholipour2017normative}.

In addition, we used $329$ 3D MRIs from a private dataset.
All images in the private dataset were part of routine clinical care and were acquired at University Hospital Leuven (UHL) and Medical University of Vienna (MUW)
due to congenital malformations seen on ultrasound.
In total, 
$102$ cases with spina bifida aperta,
$35$ cases with other central nervous system pathologies,
and 
$192$ cases with other malformations, though with normal brain, and referred as controls,
were included.
The gestational age at MRI ranged from $19$ weeks to $40$ weeks.
Some of those 3D MRIs and their manual segmentations were used in previous studies~\citep{emam2021longitudinal,fidon2021atlas,fidon2021label,mufti2021cortical}.
We have started to make fetal brain T2w 3D MRIs publicly available\footnote{\url{https://www.cir.meduniwien.ac.at/research/fetal/}}.
For each study, at least three orthogonal T2-weighted HASTE series of the fetal brain were collected on a $1.5$T scanner using an echo time of $133$ms, a repetition time of $1000$ms, with no slice overlap nor gap, pixel size $0.39$mm to $1.48$mm, and slice thickness $2.50$mm to $4.40$mm.
A radiologist attended all the acquisitions for quality control.

The reconstructed fetal brain 3D MRIs were obtained using \texttt{NiftyMIC}~\citep{ebner2020automated} 
a state-of-the-art super resolution and reconstruction algorithm. The volumes were all reconstructed to a resolution of $0.8$ mm isotropic and registered to a fetal brain atlas~\citep{gholipour2017normative}.
The 2D MRIs were also corrected for image intensity bias field as implemented in \texttt{NiftyMIC}.
Our pre-processing improves the resolution, and removes motion between neighboring slices and motion artefacts present in the original 2D slices~\citep{ebner2020automated}.
It also facilitates the manual delineation of the fetal brain structures compared to the original 2D slices.
We used volumetric brain masks to mask the tissues outside the fetal brain.
Those brain masks were obtained using the automatic segmentation methods described in~\citep{ebner2020automated,ranzini2021monaifbs}.

The labelling protocol used for white matter, intra-axial CSF, and cerebellum is the same as in~\citep{payette2021automatic}.
We use the term \textit{intra-axial CSF} rather than \textit{ventricular system} because in addition to the lateral ventricles, third ventricle, and forth ventricle, it also contains the cavum septum pellucidum and the cavum vergae that are not part of the ventricular system~\citep{tubbs2011cavum}.
The three tissue types were segmented for our private dataset by DE, EVE, FG, LF, MA, NM, and TD under the supervision of MA a paediatric radiologist specialized in fetal brain anatomy, who quality controlled and corrected all manual segmentations.

\paragraph{Brain Tumor Dataset.}
We have used the BraTS 2019 dataset because it is the last edition of the BraTS challenge for which information about the image acquisition center is available at the time of writing.
The dataset contains the same four MRI sequences (T1, ceT1, T2, and FLAIR) for 448 cases, corresponding to patients with either a high-grade Gliomas or a low-grade Gliomas.
All the cases were manually segmented for peritumoral edema, enhancing tumor, and non-enhancing tumor core using the same labeling protocol~\citep{menze2014multimodal,bakas2018identifying,bakas2017advancing}.
We split the 323 cases of the BraTS 2019 \textit{training} dataset into 268 for training and 67 for validation.
In addition, the BraTS 2019 \textit{validation} dataset that contains 125 cases was used for testing.

\begin{figure}[h!]
    \setlength{\lineskip}{0pt}
    \centering
    \includegraphics[width=\textwidth,trim=0cm 7.2cm 0.2cm 3.1cm,clip]{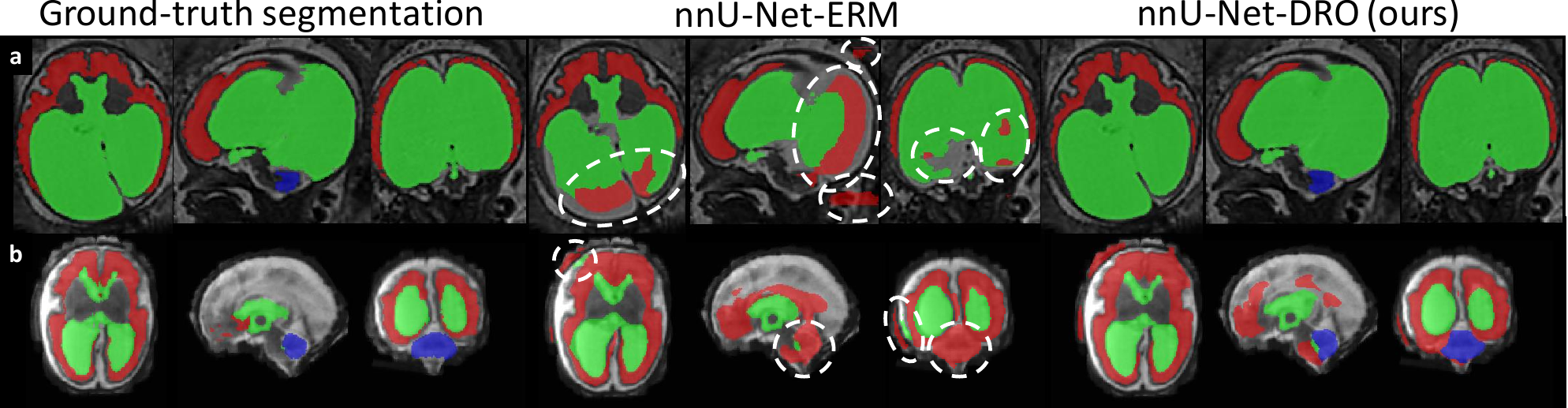}
    \includegraphics[width=\textwidth,trim=0cm 0cm 0cm 1.5cm,clip]{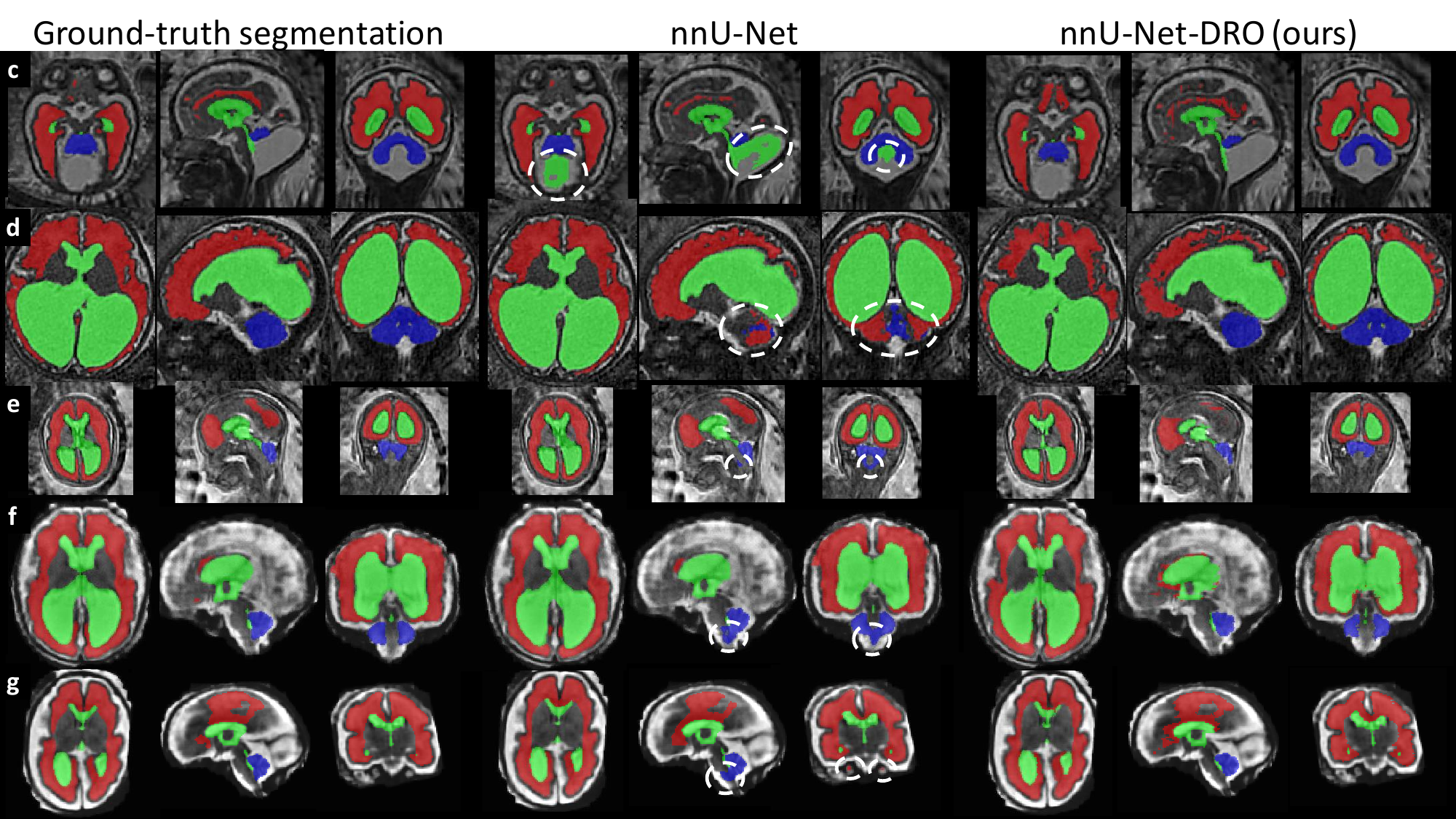}
    \caption{\textbf{Qualitative Results for Fetal Brain 3D MRI Segmentation using DRO.}
    \textcolor{black}{
    We have highlighted in white areas with severe violation of the anatomy by nnU-Net-ERM. Most of them are avoided by our nnU-Net-DRO.
    nnU-Net-ERM and nnU-Net-DRO differ only by the use of the hardness weighted sampler for the latter.
    }
    a) Fetus with aqueductal stenosis (34 weeks).
    b) Fetus with spina bifida aperta (27 weeks).
    c) Fetus with Blake's pouch cyst (29 weeks).
    d) Fetus with tuberous sclerosis complex (34 weeks).
    e) Fetus with spina bifida aperta (22 weeks).
    f) Fetus with spina bifida aperta (31 weeks).
    g) Fetus with spina bifida aperta (28 weeks).
    For cases a) and b), nnU-Net-ERM~\citep{isensee2021nnu} misses completely the cerebellum and achieves poor segmentation for the white matter and the ventricles.
    %
    For case c), a large part of the Blake's pouch cyst is wrongly included in the ventricular system segmentation by nnU-Net-ERM. This is not the case for the proposed nnU-Net-DRO.
    For case d), nnU-Net-ERM fails to segment the cerebellum correctly and a large part of the cerebellum is segmented as part of the white matter. In contrast, our nnU-Net-DRO correctly segment cerebellum and white matter for this case.
    For cases e) f) and g), 
    \textcolor{black}{
    nnU-Net-ERM wrongly included parts of the brainstem in the cerebellum segmentation.
    }
    nnU-Net-DRO does not make this mistake.
    We emphasise that the segmentation of the cerebellum for spina bifida aperta is essential for studying and evaluating the effect of surgery in-utero.
    }
    \label{fig:res_qualitative_fetal}
\end{figure}

\paragraph{Deep Learning Pipeline.}
The deep learning pipeline used was based on nnU-Net~\citep{isensee2021nnu}, which is a generic deep learning pipeline for medical image segmentation, that has been shown to outperform other deep learning pipelines on 23 public datasets without the need to manually tune the loss function or the deep neural network architecture.
Specifically, we used nnU-Net version 2 in 3D-full-resolution mode which is the recommended mode for isotropic 3D MRI data and the code is publicly available at~\url{https://github.com/MIC-DKFZ/nnUNet}.

Like most deep learning pipelines in the literature, nnU-Net is based on ERM.
For clarity, in the following we will sometimes refer to the unmodified nnU-Net as nnU-Net-ERM.

The meta-parameters used for the deep learning pipeline used were determined automatically using the heuristics developed in nnU-Net~\citep{isensee2021nnu}.
The 3D CNN selected for the brain tumor data is based on 3D U-Net~\citep{cciccek20163d} with 5 (resp. 6) levels for fetal brain segmentation (resp. brain tumor segmentation) and 32 features after the first convolution that are multiplied by 2 at each level with a maximum set at 320.
The 3D CNN uses leaky $\relu$ activation, instance normalization~\citep{ulyanov2016instance}, max-pooling downsampling operations and linear upsampling with learnable parameters.
In addition, the network is trained using the addition of the mean Dice loss and the cross entropy, and deep supervision~\citep{lee2015deeply}.
The default optimization step is SGD with a momentum of $0.99$ and Nesterov update, a batch size of 4 (resp. 2) for fetal brain segmentation (resp. brain tumor segmentation), and a decreasing learning rate defined for each epoch $t$ as
\[
\eta_t = 0.01 \times \left(1 - \frac{t}{t_{max}}\right)^{0.9}
\]
where $t_{max}$ is the maximum number of epochs fixed as $1000$.
Note that in nnU-Net, one epoch is defined as equal to 250 batches, irrespective of the size of the training dataset.
A patch size of $96 \times 112 \times 96$ (resp. $128 \times 192 \times 128$) was selected for fetal brain segmentation (resp. brain tumor segmentation), which is not sufficient to fit the whole brain of all the cases. As a result, a patch-based approach is used as often in medical image segmentation applications.
A large number of data augmentation methods are used: random cropping of a patch, random zoom, gamma intensity augmentation, multiplicative brightness, random rotations, random mirroring along all axes, contrast augmentation, additive Gaussian noise, Gaussian blurring and simulation of low resolution.
nnU-Net automatically splits the training data into 5 folds $80\%$ training/$20\%$ validation.
For the experiments on brain tumor segmentation, only the networks corresponding to the first fold were trained.
For the experiments on fetal brain segmentation, 5 models were trained, one for each fold, and the predicted class probability maps of the 5 models are averaged at inference to improve robustness~\citep{isensee2021nnu}.
GPUs NVIDIA Tesla V100-SXM2 with 16GB of memory were used for the experiments.
Training each network took from 4 to 6 days.

Our only modifications of the nnU-Net pipeline is the addition of our hardness weighted sampling when "DRO" is indicated and for some experiments we modified the optimization update rule as indicated in Table~\ref{tab:models_results}.
Our implementation of the nnU-Net-DRO training procedure is publicly available at \url{https://github.com/LucasFidon/HardnessWeightedSampler}.
If "ERM" is indicated and nothing is indicated about the optimization update rule, it means that nnU-Net~\citep{isensee2021nnu} is used without any modification.

\begin{table}[t]
	\centering
	\caption{\textbf{Evaluation of Distribution Robustness with Respect to the Pathology (260 3D MRIs).}
	\textbf{nnU-Net-ERM} is the unmodified nnU-Net pipeline~\citep{isensee2021nnu} in which Empirical Risk Minimization (ERM) is used.
	\textbf{nnU-Net-DRO} is the nnU-Net pipeline modified to use the proposed \textit{hardness weighted sampler} and in which Distributionally Robust Optimization (DRO) is therefore used.
	\textcolor{red}{\bf WM}: White matter, 
	\textcolor{ForestGreen}{\bf In-CSF}: Intra-axial CSF, 
	\textcolor{blue}{\bf Cer}: Cerebellum.
	IQR: interquartile range,
	$\textbf{p}_{X}$: $X^{\textrm{th}}$ percentile of the Dice score distribution in percentage.
    Best values are in bold \textcolor{black}{and improvements of at least $5$ points of percentage are highlighted.}
	}
	\begin{tabularx}{\textwidth}{c c c *{6}{Y}}
		\toprule
        \multicolumn{1}{c}{} & \multicolumn{1}{c}{} & \multicolumn{1}{c}{}
        & \multicolumn{6}{c}{Dice Score ($\%$)} \\
        \cmidrule(lr){4-9} 
		\multicolumn{1}{c}{\bf CNS} &
		\multicolumn{1}{c}{\bf Method} &
		\multicolumn{1}{c}{\bf ROI} &
		Mean & Median & IQR & $\textbf{p}_{25}$ & $\textbf{p}_{10}$ & $\textbf{p}_5$ \\ 
		\midrule
		\textbf{Controls} & nnU-Net-ERM &
		        \textcolor{red}{\bf WM} & 
		        $\bf94.4$ & 95.2 & \bf 2.8 & \bf 93.3 & \bf 91.5 & \colorbox{Goldenrod}{\bf 90.6} \\
		(107 volumes) & (baseline) & 
		        \textcolor{ForestGreen}{\bf In-CSF} &
		        90.3 & 92.4 & 6.4 & 87.8 & 80.7 & 79.0 \\
		 & & 
		        \textcolor{blue}{\bf Cer} &
		        \bf 95.7 & 97.0 & 3.4 & \bf 94.2 & 91.3 & \bf 90.4 \\
	        \cmidrule(lr){2-9}
	     & nnU-Net-DRO &
		        \textcolor{red}{\bf WM} & 
		        \bf 94.4 & \bf95.3 & 3.0 & 93.2 & 91.1 & 90.1 \\
		 & (ours) & 
		        \textcolor{ForestGreen}{\bf In-CSF} &
		        \bf 90.4 & \bf 92.7 & \bf 6.2 & \bf 87.9 & \bf 81.7 & \bf 79.1 \\
		 & & 
		        \textcolor{blue}{\bf Cer} &
		        \bf 95.7 & \bf 97.1 & \bf 3.3 & \bf 94.2 & \bf 91.4 & 90.1 \\
		 \cmidrule(lr){1-9}
	     \textbf{Spina Bifida} & nnU-Net-ERM &
		        \textcolor{red}{\bf WM} & 
		        89.6 & 92.1 & 4.1 & 89.5 & 80.6 & 73.8 \\
		 (112 volumes) & (baseline) & 
		        \textcolor{ForestGreen}{\bf In-CSF} &
		        91.4 & 93.9 & \bf 6.4 & 89.6 & \bf 86.9 & \bf 83.7 \\
		 & & 
		        \textcolor{blue}{\bf Cer} &
		        76.8 & 87.8 & 11.1 & 80.4 & 15.8 & \bf 0.0 \\
		    \cmidrule(lr){2-9}
		    & nnU-Net-DRO &
		        \textcolor{red}{\bf WM} & 
		        \colorbox{Goldenrod}{\bf 90.1} & \bf 92.2 & \bf 4.0 & \bf 89.9 & \colorbox{Goldenrod}{\bf 81.6} & \colorbox{Goldenrod}{\bf 74.8}\\
		 & (ours)  & 
		        \textcolor{ForestGreen}{\bf In-CSF} &
		        $\bf 91.6$ & \bf 94.1 & \bf 6.4 & \bf 90.0 & 86.7 & 83.6 \\
		 & & 
		        \textcolor{blue}{\bf Cer} &
		        \colorbox{Goldenrod}{\bf 77.8} & \bf 87.9 & \colorbox{Goldenrod}{\bf 9.7} & \colorbox{Goldenrod}{\bf 82.0} & \colorbox{Goldenrod}{\bf 43.3} & \bf 0.0 \\
		 \cmidrule(lr){1-9}
	     \textbf{Other Abn.} & nnU-Net-ERM &
		        \textcolor{red}{\bf WM} & 
		        90.3 & \bf 92.6 & \bf 4.6 & 90.1 & 88.0 & 71.6 \\
		 (41 volumes) & (baseline) & 
		        \textcolor{ForestGreen}{\bf In-CSF} &
		        87.4 & 87.9 & 10.4 & 82.7 & 77.7 & 75.9 \\
		 & & 
		        \textcolor{blue}{\bf Cer} &
		        90.4 & 92.8 & \bf5.4 & \bf90.7 & \bf87.5 & 81.4 \\
		 \cmidrule(lr){2-9}
	     & nnU-Net-DRO &
		        \textcolor{red}{\bf WM} & 
		        \bf 90.4 & \bf 92.6 & 4.7 & \bf 90.2 & \bf 88.2 & \colorbox{Goldenrod}{\bf 73.5} \\
		 & (ours) & 
		        \textcolor{ForestGreen}{\bf In-CSF} &
		        \bf 87.9 & \bf 88.1 & \colorbox{Goldenrod}{\bf 9.5} & \colorbox{Goldenrod}{\bf 83.3} & \colorbox{Goldenrod}{\bf 80.4} & \colorbox{Goldenrod}{\bf 77.7} \\
		 & & 
		        \textcolor{blue}{\bf Cer} &
		        \colorbox{Goldenrod}{\bf 91.3} & \bf 93.0 & 5.5 & \bf 90.7 & \bf 87.5 & \colorbox{Goldenrod}{\bf 82.7}\\
    %
	\bottomrule
	\end{tabularx}
	\label{tab:models_results}
\end{table}

\paragraph{Hyper-parameters of the Hardness Weighted Sampler.}
For brain tumor segmentation, we tried the values $\{10, 100, 1000\}$ of $\de$ with or without importance sampling. Using $\de=100$ with importance sampling lead to the best mean dice score on the validation split of the training dataset.
For fetal brain segmentation, we tried only $\beta=100$ with importance sampling.
When importance sampling is used, the clipping values $w_{min}=0.1$ and $w_{max}=10$ are always used.
No other values of $w_{max}$ and $w_{min}$ have been tested.
%

\paragraph{Metrics.}
We evaluate the quality of the automatic segmentations using the Dice score~\citep{dice1945measures,fidon2017generalised}.
We are particularly interested in measuring the statistical risk of the results as a way to evaluate the robustness of the different methods.

In the BraTS challenge, this is usually measured using the interquartile range (IQR) which is the difference between the percentiles at $75\%$ and $25\%$ of the the metric values~\citep{bakas2018identifying}.
We therefore reported the mean, the median and the IQR of the Dice score in Table~\ref{tab:models_results_brats}.
For fetal brain segmentation, in addition to the mean, median, and IQR, we also report the percentiles of the Dice score at $25\%$, $10\%$, and $5\%$.
In Table~\ref{tab:models_results}, we report those quantities for the Dice scores of the three tissue types white matter, intra-axial CSF, and cerebellum.

For each method, nnU-Net is trained 5 times using different train/validation splits and different random initializations.
The 5 same splits, computed randomly, are used for the two methods.
The results for fetal brain 3D MRI segmentation in Table~\ref{tab:models_results} are for the ensemble of the 5 3D U-Nets.
Ensembling is known to increase the robustness of deep learning methods for segmentation~\citep{isensee2021nnu}.
It also makes the evaluation less sensitive to the random initialization and to the stochastic optimization.

\begin{table}[bt]
	\centering
	\caption{\textbf{Dice Score Evaluation on the BraTS 2019 Online Validation Set (125 cases).}
	Metrics were computed using the BraTS online evaluation platform (\url{https://ipp.cbica.upenn.edu/}).
	ERM: Empirical Risk Minimization,
	DRO: Distributionally Robust Optimization,
	SGD: plain SGD (no momentum used),
	Nesterov: SGD with Nesterov momentum,
	IQR: Interquartile range.
	The best values \textcolor{black}{overall} are in bold \textcolor{black}{and improvements of at least $5$ points of percentage when comparing ERM and DRO for the same optimizer are highlighted.}
	}
	\begin{tabularx}{\textwidth}{c c *{9}{Y}}
		\toprule
        \textbf{Optim.}
        & \textbf{Optim.}
        & \multicolumn{3}{c}{Enhancing Tumor} 
        & \multicolumn{3}{c}{Whole Tumor}
        & \multicolumn{3}{c}{Tumor Core}\\
    \cmidrule(lr){3-5} \cmidrule(lr){6-8} \cmidrule(lr){9-11}
		\textbf{problem} & \textbf{update} & 
		Mean & Median & IQR & Mean & Median & IQR & Mean & Median & IQR\\
	\midrule
	    ERM & SGD
	     & 71.3 & 86.0 & 20.9
	     & 90.4 & 92.3 & 6.1
	     & 80.5 & 88.8 & 17.5\\
	\cmidrule(lr){1-11}
	    DRO & SGD
	     & \colorbox{Goldenrod}{72.3} & \colorbox{Goldenrod}{87.2} & \colorbox{Goldenrod}{19.1}
	     & 90.5 & \bf92.6 & 6.0
	     & \colorbox{Goldenrod}{82.1} & \colorbox{Goldenrod}{89.7} & \colorbox{Goldenrod}{15.2}\\
	\midrule
	\midrule
		ERM & Nesterov
		 & 73.0 & 87.1 & 15.6
		 & \bf90.7 & \bf92.6 & \colorbox{Goldenrod}{\bf5.4}
		 & 83.9 & \colorbox{Goldenrod}{\bf90.5} & 14.3\\
	\cmidrule(lr){1-11}
        DRO & Nesterov
         & \colorbox{Goldenrod}{\bf74.5} & \bf87.3 & \colorbox{Goldenrod}{\bf13.8}
		 & 90.6 & \bf92.6 & 5.9
		 & \bf84.1 & 90.0 & \colorbox{Goldenrod}{\bf12.5}\\
	\bottomrule
	\end{tabularx}
	\label{tab:models_results_brats}
\end{table}

\begin{figure}[bt]
    \centering
    \includegraphics[width=\linewidth]{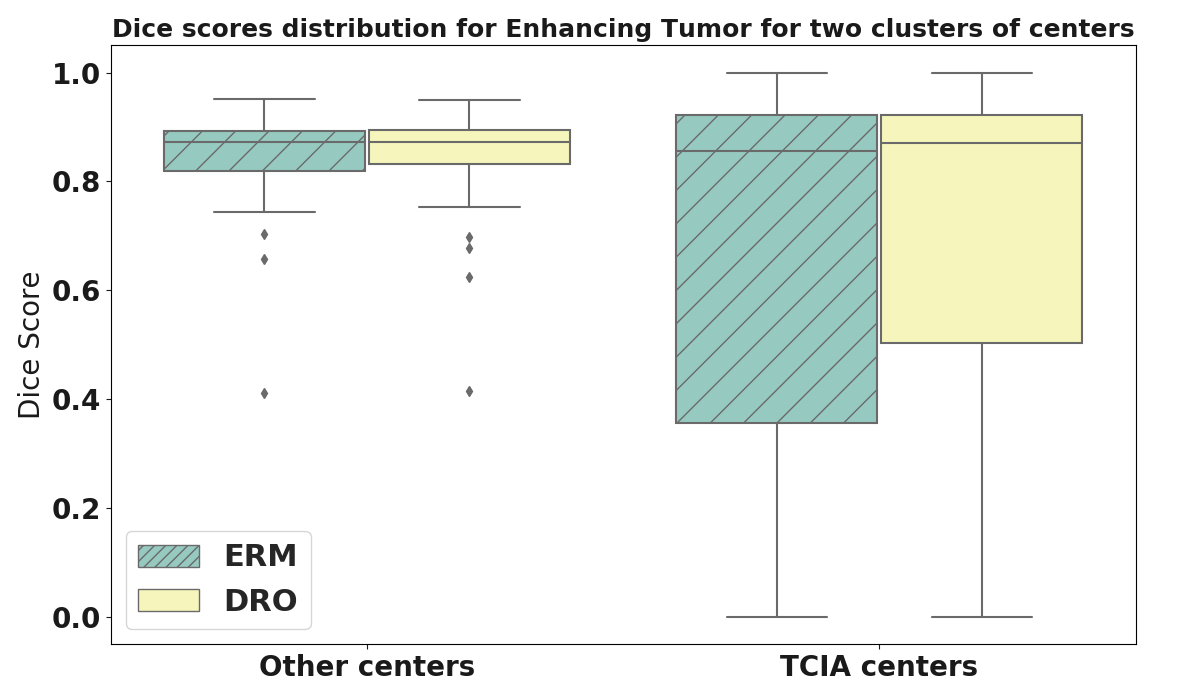}
    \caption{\label{fig:difference_centers}
        \textbf{Dice scores distribution on the BraTS 2019 validation dataset for cases from a center of TCIA (76 cases) and cases from other centers (49 cases).}
        This shows that the lower interquartile range of DRO for the enhancing tumor comes specifically from a lower number of poor segmentations on cases coming from The Cancer Imaging Archive (TCIA).
        This suggests that DRO can deal with some of the confounding biases present in the training dataset, and lead to a model that is more fair.
        }
\end{figure}

\paragraph*{Results.}
The quantitative comparison of nnU-Net-ERM and nnU-Net-DRO on fetal brain 3D MRI segmentation for the three different central nervous system conditions control, spina bifida, and other abnormalities can be found in Table~\ref{tab:models_results}.

For spina bifida and other brain abnormalities, the proposed nnU-Net-DRO achieves same or higher mean Dice scores than nnU-Net-ERM~\citep{isensee2021nnu} with $+0.5$ percentage points (pp) for white matter and $+1$pp for the cerebellum of spina bifida cases and $+0.9$pp for the cerebellum for other abnormalities.
In addition, nnU-Net-DRO achieves comparable (at most $0.1$pp of difference) or lower IQR than nnU-Net-ERM with $-1.4$pp for the cerebellum of spina bifida cases and $-0.9$pp for the intra-axial CSF of cases with other abnormalities.
For controls, the mean, median, and IQR of the Dice scores of nnU-Net-DRO and nnU-Net-ERM differ by less than $0.2$pp for the three tissue types.
This suggests that nnU-Net-DRO is more robust to anatomical variabilities associated with abnormal brains, while retaining the same segmentation performance on neurotypical cases.

In terms of median Dice score, nnU-Net-DRO and nnU-Net-ERM differ by less than $0.3$pp for all tissue types and conditions.
Therefore the differences in terms of mean Dice scores mentioned above are not due to improved segmentation in the middle of the Dice score performance distribution.

The comparison of the percentiles at $25\%$, $10\%$, and $5\%$ of the Dice score allows us to compare methods at the tail of the Dice scores distribution where segmentation methods reach their worst-case performance.
For spina bifida, nnU-Net-DRO achieves higher values of percentiles than nnU-Net-ERM for the white matter ($+1.0$pp for $\textbf{p}_{10}$ and $+1.0$pp for $\textbf{p}_{5}$), 
and for the cerebellum ($+1.6$pp for $\textbf{p}_{25}$ and $+27.5$pp for $\textbf{p}_{10}$).
And for other brain abnormalities, nnU-Net-DRO achieves higher values of percentiles than nnU-Net-ERM for the white matter ($+1.9$pp for $\textbf{p}_{5}$), 
for the intra-axial CSF ($+0.6$pp for $\textbf{p}_{25}$, $+2.3$pp for $\textbf{p}_{10}$ and $+1.8$pp for $\textbf{p}_{5}$), 
and for the cerebellum ($+1.3$pp for $\textbf{p}_{5}$).
All the other percentile values differ by less than $0.5$pp of Dice score between the two methods.
This suggests that nnU-Net-DRO achieves better worst case performance than nnU-Net-ERM for abnormal cases.
However, both methods have a percentile at $5\%$ of the Dice score equal to $0$ for the cerebellum of spina bifida cases. This indicates that both methods completely miss the cerebellum for spina bifida cases in $5\%$ of the cases.

As can be seen in the qualitative results of \Figref{fig:res_qualitative_fetal}, there are cases for which nnU-Net-ERM predicts an empty cerebellum segmentation while nnU-Net-DRO achieves satisfactory cerebellum segmentation.
There were no cases for which the converse was true.
However, there were also spina bifida cases for which both methods failed to predict the cerebellum.
Robust segmentation of the cerebellum for spina bifida is particularly relevant for the evaluation of fetal brain surgery for spina bifida aperta~\citep{aertsen2019reliability,danzer2020fetal,sacco2019fetal}.
\textcolor{black}{
All the spina bifida 3D MRIs with missing cerebellum in the automatic segmentations were 3D MRIs from the FeTA dataset~\cite{payette2021automatic} and represented brains of fetuses with spina bifida before they were operated on.
The cerebellum is more difficult to detect using MRI before surgery as compared to early or late after surgery~\citep{aertsen2019reliability,danzer2007fetal}.
No 3D MRI with the combination of those two factors were present in the training dataset (Table.~\ref{tab:data}).
This might explain why DRO did not help improving the segmentation quality for those cases.
DRO aims at improving the performance on subgroups that were underrepresented in the training dataset, not subgroups that were not represented at all.
}

In Table~\ref{tab:models_results}, it is worth noting that overall the Dice score values decrease for the white matter and the cerebellum between controls and spina bifida and abnormal cases.
It was expected due to the higher anatomical variability in pathological cases.
However, the Dice score values for the ventricular system tend to be higher for spina bifida cases than for controls.
This can be attributed to the large proportion of spina bifida cases with enlarged ventricles because the Dice score values tend to be higher for larger regions of interest.

For our experiments on brain tumor segmentation, Table~\ref{tab:models_results_brats} summarizes the performance of training nnU-Net using ERM or using DRO.
Here, we experiment with two SGD-based optimizers.
For both ERM and DRO, the optimization update rule used was either plain SGD without momentum (SGD), or SGD with a Nesterov momentum equal to $0.99$ (Nesterov).
Especially, for the latter, this implies that step 12 of \Algref{alg:1} is modified to use SGD with Nesterov momentum. It was also the case for our experiments on fetal brain 3D MRI segmentation.
For DRO, the results presented here are for $\de=100$ and using importance sampling (step 6 of \Algref{alg:1}).

As illustrated in Table~\ref{tab:models_results_brats}, for both ERM and DRO, the use of SGD with Nesterov momentum outperforms plain-SGD for all metrics and all regions of interest.
This result was expected for ERM, for which it is common practice in the deep learning literature to use SGD with a momentum.
Our results here suggest that the benefit of using a momentum with SGD is retained for DRO.

For both optimizers, DRO outperforms ERM in terms of IQR for the enhancing tumor and the tumor core by approximately $2$pp of Dice score, and in terms of mean Dice score for the enhancing tumor by $1$pp for the plain-SGD and $1.5$pp for SGD with Nesterov momentum.
For plain-SGD, DRO also outpermforms ERM in terms of mean Dice score for the tumor core by $1.6$pp.
The IQR is the global statistic used in the BraTS challenge to measure the level of robustness of a method~\citep{bakas2018identifying}.
In addition, \Figref{fig:difference_centers} shows that the lower IQR of DRO for the enhancing tumor comes specifically from a lower number of poor segmentations on cases coming from The Cancer Imaging Archive (TCIA).
This suggests that DRO can deal with some of the confounding biases present in the training dataset, and lead to a model that is more fair with respect to the acquisition center of the MRI.

Since the same improvements are observed independently of the optimization update rule used.
This suggests that in practice \Algref{alg:1} still converges when a momentum is used, even if Theorem~\ref{th:convergence_dro} was only demonstrated to hold for plain-SGD.

The value $\de=100$ and the use of importance sampling was selected based on the mean Dice score on the validation split of the training dataset.
Results for $\de \in \{10, 100, 1000\}$ with Nesterov momentum and with or without importance sampling can be found in Appendix~\ref{s:more_results_brats} Table~\ref{tab:models_all_results}.
The tendency described previously still holds true for the enhancing tumor for $\de$ equal to $10$ or $100$ with and without importance sampling.
The mean Dice score is improved by $0.4$pp to $2.3$pp
and the IQR is reduced by $1.3$pp to $2.3$pp for the four DRO models as compared to the ERM model.
For the tumor core with $\de=100$ mean and IQR are improved over ERM with and without importance sampling.
However, for $\de=10$ with importance sampling there was a loss of performance as compared to ERM for the whole tumor.
This problem was not observed with $\de=10$ without importance sampling. For the other models with $\de$ equal to $10$ or $100$ similar Dice score performance similar to the one ERM was observed for the whole tumor.
This suggests that overall the use of ERM or DRO does not affect the segmentation performance of the whole tumor.
One possible explanation of this is that Dice scores for the whole tumor are already high for almost all cases when ERM is used with a low IQR.
In addition, DRO and the \textit{hardness weighted sampler} are sensitive to the loss function, here the mean-class Dice loss plus cross entropy loss.
In the case of brain tumor segmentation, we hypothesise that the loss function is more sensitive to the segmentation performance for the tumor core and the enhancing tumor than for the whole tumor.

When $\de$ becomes too large ($\de=1000$) a decrease of the mean and median Dice score for all regions is observed as compared to ERM.
In this case, DRO tends towards the maximization of the worst-case example only which appears to be unstable using our \Algref{alg:1}.
For all values of $\de$ the use of importance sampling, as described in steps 6-8 of \Algref{alg:1}, improves the IQR of the Dice scores for the enhancing tumor and the tumor core.
We therefore recommend to use \Algref{alg:1} with importance sampling.

\textcolor{black}{
\subsection{Computational Time and Memory Overhead of \Algref{alg:1}}\label{sec:efficiency}
}

\begin{table}[bt]
	\centering
	\caption{
	\textcolor{black}{
	\textbf{Estimated Computational Time and Memory Overhead of the hardness weighted sampler in \Algref{alg:1}}.
	The times (in seconds) are estimated using a batch size of $2$ and $\de=100$ and by taking the average sampling time over $10,000$ sampling operations for each number of samples.
	It is worth noting that the sampling operations are computed on the CPUs as in most deep learning pipeline.
	The time and memory overhead of the proposed hardness weighted sampler is negligible for training datasets with up to 1 million samples.
	}
	}
	\begin{tabularx}{\textwidth}{c *{6}{Y}}
	\toprule
	    \# Samples 
	     & $10^2$ & $10^3$ 
	     & $10^4$ & $10^5$ & $10^6$ 
	     & $10^7$ \\
	\midrule
	    Time (in sec) 
	     & $1.3\times 10^{-4}$ & $1.5\times 10^{-4}$
	     & $2.6\times 10^{-4}$ & $2.4\times 10^{-3}$ & $2.1\times 10^{-2}$ 
	     & $1.8\times 10^{-1}$ \\
	\midrule
		Memory (in MB) 
		 & $7.6 \times 10^{-4}$ & $7.6\times 10^{-3}$
		 & $7.6\times 10^{-2}$ & $7.6\times 10^{-1}$ & 7.6 
		 & 76.3 \\
	\bottomrule
	\end{tabularx}
	\label{tab:efficiency}
\end{table}

The main additional computational cost \textcolor{black}{in} \Algref{alg:1} is due to the hardness weighted sampling in steps 4 and 5 that is dependent on the number $n$ of training examples.
\textcolor{black}{
In Table~\ref{tab:efficiency}, we have computed the computational time and memory overhead of the hardness weighted sampler for different sizes of the training dataset.
We have computed that additional time required is less than $0.5$ second and the additional memory less than $100$ MB for up to $n=10^7$ using a batch size of $2$ and the function \texttt{random.choice} of Numpy version $1.21.1$.
The times were estimated using $12$ Intel(R) Core(TM) i7-8750H CPU @ 2.20GHz.
The additional time and memory that occurs due to the proposed hardness weighted sampling is therefore negligible for all the datasets used in practice in medical image segmentation.
}
For our brain tumor segmentation training set of n=268 volumes and a batch size of 2, the additional memory usage of Algorithm 1 is only 2144 bytes of memory (one float array of size n) and the additional computational time is approximately $10^{-4}$ seconds per iteration using the Python library numpy, i.e. approximately $0.005 \%$ of the total duration of an iteration.
The size of the training dataset for fetal brain 3D MRI segmentation being lower, the additional memory usage and the additional computational time are even lower than for brain tumor segmentation.
\textcolor{black}{
We have made available a python script in our \href{https://github.com/LucasFidon/HardnessWeightedSampler}{GitHub repository} that allows to easily compute the additional time and memory occurring because of the hardness weighted sampler for any number of samples and batch size.
}
\section{Discussion and Conclusion}

In this paper, we have shown that efficient training of deep neural networks with Distributionally Robust Optimization (DRO) with a $\phi$-divergence is possible.

The proposed \textit{hardness weighted sampler} for training a deep neural network with Stochastic Gradient Descent (SGD) for DRO is as straightforward to implement, and as computationally efficient as SGD for Empirical Risk Minimization (ERM).
It can be used for deep neural networks with any activation function (including $\relu$), and with any per-example loss function.
We have shown that the proposed approach can formally be described as a principled Hard Example Mining strategy (Theorem~\ref{th:hard_example_mining}) and is related to minimizing the percentile of the per-example loss distribution \eqref{eq:perc}.
In addition, we prove the convergence of our method for over-parameterized deep neural networks (Theorem~\ref{th:convergence_dro}).
Thereby, extending the convergence theory of deep learning of~\cite{allen-zhu19a}.
This is, to the best of our knowledge, the first convergence result for training a deep neural network based on DRO.

In practice, we have shown that our hardness weighted sampling method can be easily integrated in a state-of-the-art deep learning framework for medical image segmentation.
Interestingly, the proposed algorithm remains stable when SGD with momentum is used.
\textcolor{black}{
The hardness weighted sampling has one hyperparameter $\de > 0$.
Our experiments suggest that similar values of $\de$ lead to improve robustness in different applications.
We hypothesize that good values of $\de$ are of the order of the inverse of the standard deviation of the vector of per-volume (stale) losses during the training epochs that precede convergence.
}

The high anatomical variability of the developing fetal brain across gestational ages and pathologies hampers the robustness of deep neural networks trained by maximizing the average per-volume performance.
Specifically, it limits the generalization of deep neural networks to abnormal cases for which few cases are available during training.
In this paper, we propose to mitigate this problem by training deep neural networks using Distributionally Robust Optimization (DRO) with the proposed hardness weighted sampling.
We have validated the proposed training method on a multi-centric dataset of $437$ fetal brain T2w 3D MRIs with various diagnostics.
nnU-Net trained with DRO achieved improved segmentation results for pathological cases as compared to the unmodified nnU-Net, while achieving similar segmentation performance for the neurotypical cases.
Those results suggest that nnU-Net trained with DRO is more robust to anatomical variabilities than the original nnU-Net that is trained with ERM.
In addition, we have performed experiments on the open-source multiclass brain tumor segmentation dataset BraTS~\citep{bakas2018identifying}.
Our results on BraTS suggests that DRO can help improving the robustness of deep neural network for segmentation to variations in the acquisition protocol of the images.

However, we have also found in our experiments that all deep learning models, either trained with ERM or DRO, failed in some cases.
For example, the models evaluated all missed the cerebellum in at least $5\%$ of the spina bifida aperta cases.
As a result, while our results do suggest that DRO with our method can improve the robustness of deep neural networks for segmentation, they also show that DRO alone with our method does not provide a guarantee of robustness.
DRO with a $\phi$-divergence reweights the examples in the training dataset but cannot account for subsets of the true distribution that are not represented at all in the training dataset.
We investigate this problem in our following work~\citep{fidon2022dempster}.

We have shown that the additional computational cost of the proposed hardness weighted sampling is small enough to be negligible in practice and requires less than one second for up to $n=10^8$ examples.
The proposed \Algref{alg:1} is therefore as computationally efficient as state-of-the-art deep learning pipeline for medical image segmentation.
%
%
However, when data augmentation is used, an infinite number of training examples is virtually available.
We mitigate this problem using importance sampling and only one probability per non-augmented example.
We found that importance sampling led to improved segmentation results.

We have also illustrated in our experiments that reporting the mean and standard deviation of the Dice score is not enough to evaluate the robustness of deep neural networks for medical image segmentation.
A stratification of the evaluation is required to assess for which subgroups of the population and for which image protocols a deep learning model for segmentation can be safely used.
In addition, not all improvements of the mean and standard deviation of the Dice score are equally relevant as they can result from improvements of either the best or the worst segmentation cases.
Regarding the robustness of automatic segmentation methods across various conditions, one is interested in improvements of segmentation metrics in the tail of the distribution that corresponds to the worst segmentation cases.
%
%
To this end, one can report the interquartile range (IQR) and measures of risk such as percentiles.


\acks{
This project has received funding from the European Union's Horizon 2020 research and innovation program under the Marie Sk{\l}odowska-Curie grant agreement TRABIT No 765148;
Wellcome [203148/Z/16/Z; WT101957], EPSRC [NS/A000049/1; NS/A000027/1].
Tom Vercauteren is supported by a Medtronic / RAEng Research Chair [RCSRF1819\textbackslash7\textbackslash34].
Data used in this publication were obtained as part of the RSNA-ASNR-MICCAI Brain Tumor Segmentation (BraTS) Challenge project through Synapse ID (syn25829067).
}

%
\ethics{The work follows appropriate ethical standards in conducting research and writing the manuscript, following all applicable laws and regulations regarding treatment of human subjects.}

\coi{S\'ebastien Ourselin is co-founder of Brainminer and non-executive director at Hypervision Surgical.
Tom Vercauteren is chief scientific officer at Hypervision Surgical.
Michael Ebner is chief executive officer at Hypervision Surgical.
Georg Langs is chief scientist and co-founder at Contextflow.
}

\bibliography{main}

\begin{thebibliography}{67}
\providecommand{\natexlab}[1]{#1}
\providecommand{\url}[1]{\texttt{#1}}
\expandafter\ifx\csname urlstyle\endcsname\relax
  \providecommand{\doi}[1]{doi: #1}\else
  \providecommand{\doi}{doi: \begingroup \urlstyle{rm}\Url}\fi

\bibitem[Aertsen et~al.(2019)Aertsen, Verduyckt, De~Keyzer, Vercauteren,
  Van~Calenbergh, De~Catte, Dymarkowski, Demaerel, and
  Deprest]{aertsen2019reliability}
M~Aertsen, J~Verduyckt, F~De~Keyzer, T~Vercauteren, F~Van~Calenbergh,
  L~De~Catte, S~Dymarkowski, P~Demaerel, and J~Deprest.
\newblock Reliability of {MR} imaging--based posterior fossa and brain stem
  measurements in open spinal dysraphism in the era of fetal surgery.
\newblock \emph{American Journal of Neuroradiology}, 40\penalty0 (1):\penalty0
  191--198, 2019.

\bibitem[Allen-Zhu et~al.(2019{\natexlab{a}})Allen-Zhu, Li, and
  Song]{allen-zhu19a}
Zeyuan Allen-Zhu, Yuanzhi Li, and Zhao Song.
\newblock A convergence theory for deep learning via over-parameterization.
\newblock In \emph{ICML}, pages 242--252, 2019{\natexlab{a}}.

\bibitem[Allen-Zhu et~al.(2019{\natexlab{b}})Allen-Zhu, Li, and
  Song]{allen2018convergence}
Zeyuan Allen-Zhu, Yuanzhi Li, and Zhao Song.
\newblock On the convergence rate of training recurrent neural networks.
\newblock In \emph{Advances in Neural Information Processing Systems 32}, pages
  6676--6688. Curran Associates, Inc., 2019{\natexlab{b}}.

\bibitem[Bakas et~al.(2017{\natexlab{a}})Bakas, Akbari, Sotiras, Bilello,
  Rozycki, Kirby, Freymann, Farahani, and Davatzikos]{bakas2017HGG}
Spyridon Bakas, Hamed Akbari, Aristeidis Sotiras, Michel Bilello, Martin
  Rozycki, Justin~S Kirby, John~B Freymann, Keyvan Farahani, and Christos
  Davatzikos.
\newblock Segmentation labels and radiomic features for the pre-operative scans
  of the {TCGA-GBM} collection.
\newblock \emph{The Cancer Imaging Archive}, 2017{\natexlab{a}}.
\newblock \doi{10.7937/K9/TCIA.2017.KLXWJJ1Q}.

\bibitem[Bakas et~al.(2017{\natexlab{b}})Bakas, Akbari, Sotiras, Bilello,
  Rozycki, Kirby, Freymann, Farahani, and Davatzikos]{bakas2017LGG}
Spyridon Bakas, Hamed Akbari, Aristeidis Sotiras, Michel Bilello, Martin
  Rozycki, Justin~S Kirby, John~B Freymann, Keyvan Farahani, and Christos
  Davatzikos.
\newblock Segmentation labels and radiomic features for the pre-operative scans
  of the {TCGA-LGG} collection.
\newblock \emph{The Cancer Imaging Archive}, 2017{\natexlab{b}}.
\newblock \doi{10.7937/K9/TCIA.2017.GJQ7R0EF}.

\bibitem[Bakas et~al.(2017{\natexlab{c}})Bakas, Akbari, Sotiras, Bilello,
  Rozycki, Kirby, Freymann, Farahani, and Davatzikos]{bakas2017advancing}
Spyridon Bakas, Hamed Akbari, Aristeidis Sotiras, Michel Bilello, Martin
  Rozycki, Justin~S Kirby, John~B Freymann, Keyvan Farahani, and Christos
  Davatzikos.
\newblock Advancing the cancer genome atlas glioma {MRI} collections with
  expert segmentation labels and radiomic features.
\newblock \emph{Scientific data}, 4:\penalty0 170117, 2017{\natexlab{c}}.

\bibitem[Bakas et~al.(2018)Bakas, Reyes, Jakab, Bauer, Rempfler, Crimi,
  Shinohara, Berger, Ha, Rozycki, et~al.]{bakas2018identifying}
Spyridon Bakas, Mauricio Reyes, Andras Jakab, Stefan Bauer, Markus Rempfler,
  Alessandro Crimi, Russell~Takeshi Shinohara, Christoph Berger, Sung~Min Ha,
  Martin Rozycki, et~al.
\newblock Identifying the best machine learning algorithms for brain tumor
  segmentation, progression assessment, and overall survival prediction in the
  {BRATS} challenge.
\newblock \emph{arXiv preprint arXiv:1811.02629}, 2018.

\bibitem[Berger et~al.(2018)Berger, Eoin, Cardoso, and
  Ourselin]{berger2018adaptive}
Lorenz Berger, Hyde Eoin, M~Jorge Cardoso, and S{\'e}bastien Ourselin.
\newblock An adaptive sampling scheme to efficiently train fully convolutional
  networks for semantic segmentation.
\newblock In \emph{Annual Conference on Medical Image Understanding and
  Analysis}, pages 277--286. Springer, 2018.

\bibitem[Bottou et~al.(2018)Bottou, Curtis, and
  Nocedal]{bottou2018optimization}
L{\'e}on Bottou, Frank~E Curtis, and Jorge Nocedal.
\newblock Optimization methods for large-scale machine learning.
\newblock \emph{Siam Review}, 60\penalty0 (2):\penalty0 223--311, 2018.

\bibitem[Byrd and Lipton(2019)]{byrd2019effect}
Jonathon Byrd and Zachary Lipton.
\newblock What is the effect of importance weighting in deep learning?
\newblock In \emph{ICML}, pages 872--881, 2019.

\bibitem[Cao and Gu(2020)]{cao2019generalization}
Yuan Cao and Quanquan Gu.
\newblock Generalization error bounds of gradient descent for learning
  overparameterized deep relu networks.
\newblock In \emph{AAAI}, 2020.

\bibitem[Chang et~al.(2017)Chang, Learned-Miller, and
  McCallum]{chang2017active}
Haw-Shiuan Chang, Erik Learned-Miller, and Andrew McCallum.
\newblock Active bias: Training more accurate neural networks by emphasizing
  high variance samples.
\newblock In \emph{Advances in Neural Information Processing Systems}, pages
  1002--1012, 2017.

\bibitem[Chernoff et~al.(1952)]{chernoff1952measure}
Herman Chernoff et~al.
\newblock A measure of asymptotic efficiency for tests of a hypothesis based on
  the sum of observations.
\newblock \emph{The Annals of Mathematical Statistics}, 23\penalty0
  (4):\penalty0 493--507, 1952.

\bibitem[Chouzenoux et~al.(2019)Chouzenoux, G{\'e}rard, and
  Pesquet]{chouzenoux2019general}
Emilie Chouzenoux, Henri G{\'e}rard, and Jean-Christophe Pesquet.
\newblock General risk measures for robust machine learning.
\newblock \emph{Foundations of Data Science}, 1:\penalty0 249, 2019.

\bibitem[{\c{C}}i{\c{c}}ek et~al.(2016){\c{C}}i{\c{c}}ek, Abdulkadir, Lienkamp,
  Brox, and Ronneberger]{cciccek20163d}
{\"O}zg{\"u}n {\c{C}}i{\c{c}}ek, Ahmed Abdulkadir, Soeren~S Lienkamp, Thomas
  Brox, and Olaf Ronneberger.
\newblock {3D U-Net}: learning dense volumetric segmentation from sparse
  annotation.
\newblock In \emph{International conference on medical image computing and
  computer-assisted intervention}, pages 424--432. Springer, 2016.

\bibitem[Csisz{\'a}r et~al.(2004)Csisz{\'a}r, Shields,
  et~al.]{csiszar2004information}
Imre Csisz{\'a}r, Paul~C Shields, et~al.
\newblock Information theory and statistics: A tutorial.
\newblock \emph{Foundations and Trends{\textregistered} in Communications and
  Information Theory}, 1\penalty0 (4):\penalty0 417--528, 2004.

\bibitem[Danzer et~al.(2007)Danzer, Johnson, Bebbington, Simon, Wilson,
  Bilaniuk, Sutton, and Adzick]{danzer2007fetal}
Enrico Danzer, Mark~P Johnson, Michael Bebbington, Erin~M Simon, R~Douglas
  Wilson, Larrissa~T Bilaniuk, Leslie~N Sutton, and N~Scott Adzick.
\newblock Fetal head biometry assessed by fetal magnetic resonance imaging
  following in utero myelomeningocele repair.
\newblock \emph{Fetal diagnosis and therapy}, 22\penalty0 (1):\penalty0 1--6,
  2007.

\bibitem[Danzer et~al.(2020)Danzer, Joyeux, Flake, and
  Deprest]{danzer2020fetal}
Enrico Danzer, Luc Joyeux, Alan~W Flake, and Jan Deprest.
\newblock Fetal surgical intervention for myelomeningocele: lessons learned,
  outcomes, and future implications.
\newblock \emph{Developmental Medicine \& Child Neurology}, 62\penalty0
  (4):\penalty0 417--425, 2020.

\bibitem[Dice(1945)]{dice1945measures}
Lee~R Dice.
\newblock Measures of the amount of ecologic association between species.
\newblock \emph{Ecology}, 26\penalty0 (3):\penalty0 297--302, 1945.

\bibitem[Duchi et~al.(2016)Duchi, Glynn, and Namkoong]{duchi2016statistics}
John Duchi, Peter Glynn, and Hongseok Namkoong.
\newblock Statistics of robust optimization: A generalized empirical likelihood
  approach.
\newblock \emph{arXiv preprint arXiv:1610.03425}, 2016.

\bibitem[Ebner et~al.(2020)Ebner, Wang, Li, Aertsen, Patel, Aughwane,
  Melbourne, Doel, Dymarkowski, De~Coppi, et~al.]{ebner2020automated}
Michael Ebner, Guotai Wang, Wenqi Li, Michael Aertsen, Premal~A Patel, Rosalind
  Aughwane, Andrew Melbourne, Tom Doel, Steven Dymarkowski, Paolo De~Coppi,
  et~al.
\newblock An automated framework for localization, segmentation and
  super-resolution reconstruction of fetal brain {MRI}.
\newblock \emph{NeuroImage}, 206:\penalty0 116324, 2020.

\bibitem[Emam et~al.(2021)Emam, Aertsen, Van~der Veeken, Fidon, Patkee,
  Kyriakopoulou, De~Catte, Russo, Demaerel, Vercauteren,
  et~al.]{emam2021longitudinal}
Doaa Emam, Michael Aertsen, Lennart Van~der Veeken, Lucas Fidon, Prachi Patkee,
  Vanessa Kyriakopoulou, Luc De~Catte, Francesca Russo, Philippe Demaerel, Tom
  Vercauteren, et~al.
\newblock Longitudinal evaluation of brain development in fetuses with
  congenital diaphragmatic hernia on mri: an original research study.
\newblock 2021.

\bibitem[{European Commission}(2019)]{ethics}
{European Commission}.
\newblock Ethics guidelines for trustworthy {AI}.
\newblock Report, {European Commission}, 2019.

\bibitem[Fenchel(1949)]{fenchel1949conjugate}
Werner Fenchel.
\newblock On conjugate convex functions.
\newblock \emph{Canadian Journal of Mathematics}, 1\penalty0 (1):\penalty0
  73--77, 1949.

\bibitem[Fidon et~al.(2017)Fidon, Li, Garcia-Peraza-Herrera, Ekanayake,
  Kitchen, Ourselin, and Vercauteren]{fidon2017generalised}
Lucas Fidon, Wenqi Li, Luis~C Garcia-Peraza-Herrera, Jinendra Ekanayake, Neil
  Kitchen, S{\'e}bastien Ourselin, and Tom Vercauteren.
\newblock Generalised {W}asserstein dice score for imbalanced multi-class
  segmentation using holistic convolutional networks.
\newblock In \emph{International MICCAI Brainlesion Workshop}, pages 64--76.
  Springer, 2017.

\bibitem[Fidon et~al.(2021{\natexlab{a}})Fidon, Aertsen, Emam, Mufti, Guffens,
  Deprest, Demaerel, David, Melbourne, Ourselin, et~al.]{fidon2021label}
Lucas Fidon, Michael Aertsen, Doaa Emam, Nada Mufti, Fr{\'e}d{\'e}ric Guffens,
  Thomas Deprest, Philippe Demaerel, Anna~L David, Andrew Melbourne,
  S{\'e}bastien Ourselin, et~al.
\newblock Label-set loss functions for partial supervision: Application to
  fetal brain {3D MRI} parcellation.
\newblock \emph{arXiv preprint arXiv:2107.03846}, 2021{\natexlab{a}}.

\bibitem[Fidon et~al.(2021{\natexlab{b}})Fidon, Aertsen, Mufti, Deprest, Emam,
  Guffens, Schwartz, Ebner, Prayer, Kasprian,
  et~al.]{fidon2021distributionally}
Lucas Fidon, Michael Aertsen, Nada Mufti, Thomas Deprest, Doaa Emam,
  Fr{\'e}d{\'e}ric Guffens, Ernst Schwartz, Michael Ebner, Daniela Prayer,
  Gregor Kasprian, et~al.
\newblock Distributionally robust segmentation of abnormal fetal brain {3D
  MRI}.
\newblock In \emph{Uncertainty for Safe Utilization of Machine Learning in
  Medical Imaging, and Perinatal Imaging, Placental and Preterm Image
  Analysis}, pages 263--273. Springer, 2021{\natexlab{b}}.

\bibitem[Fidon et~al.(2021{\natexlab{c}})Fidon, Aertsen, Shit, Demaerel,
  Ourselin, Deprest, and Vercauteren]{fidon2021partial}
Lucas Fidon, Michael Aertsen, Suprosanna Shit, Philippe Demaerel, S{\'e}bastien
  Ourselin, Jan Deprest, and Tom Vercauteren.
\newblock Partial supervision for the {FeTA} challenge 2021.
\newblock \emph{arXiv preprint arXiv:2111.02408}, 2021{\natexlab{c}}.

\bibitem[Fidon et~al.(2021{\natexlab{d}})Fidon, Viola, Mufti, David, Melbourne,
  Demaerel, Ourselin, Vercauteren, Deprest, and Aertsen]{fidon2021atlas}
Lucas Fidon, Elizabeth Viola, Nada Mufti, Anna David, Andrew Melbourne,
  Philippe Demaerel, Sebastien Ourselin, Tom Vercauteren, Jan Deprest, and
  Michael Aertsen.
\newblock A spatio-temporal atlas of the developing fetal brain with spina
  bifida aperta.
\newblock \emph{Open Research Europe}, 2021{\natexlab{d}}.

\bibitem[Fidon et~al.(2022)Fidon, Aertsen, Kofler, Bink, David, Deprest, Emam,
  Guffens, Jakab, Kasprian, et~al.]{fidon2022dempster}
Lucas Fidon, Michael Aertsen, Florian Kofler, Andrea Bink, Anna~L David, Thomas
  Deprest, Doaa Emam, Fr{\'e}d{\'e}ric Guffens, Andr{\'a}s Jakab, Gregor
  Kasprian, et~al.
\newblock A {Dempster-Shafer} approach to trustworthy {AI} with application to
  fetal brain {MRI} segmentation.
\newblock \emph{arXiv preprint arXiv:2204.02779}, 2022.

\bibitem[Gholipour et~al.(2017)Gholipour, Rollins, Velasco-Annis, Ouaalam,
  Akhondi-Asl, Afacan, Ortinau, Clancy, Limperopoulos, Yang,
  et~al.]{gholipour2017normative}
Ali Gholipour, Caitlin~K Rollins, Clemente Velasco-Annis, Abdelhakim Ouaalam,
  Alireza Akhondi-Asl, Onur Afacan, Cynthia~M Ortinau, Sean Clancy, Catherine
  Limperopoulos, Edward Yang, et~al.
\newblock A normative spatiotemporal {MRI} atlas of the fetal brain for
  automatic segmentation and analysis of early brain growth.
\newblock \emph{Scientific reports}, 7\penalty0 (1):\penalty0 1--13, 2017.

\bibitem[Harwood et~al.(2017)Harwood, Kumar, Carneiro, Reid, Drummond,
  et~al.]{harwood2017smart}
Ben Harwood, BG~Kumar, Gustavo Carneiro, Ian Reid, Tom Drummond, et~al.
\newblock Smart mining for deep metric learning.
\newblock In \emph{Proceedings of the IEEE International Conference on Computer
  Vision}, pages 2821--2829, 2017.

\bibitem[Hiriart-Urruty and Lemar{\'e}chal(2013)]{hiriart2013convex}
Jean-Baptiste Hiriart-Urruty and Claude Lemar{\'e}chal.
\newblock \emph{Convex analysis and minimization algorithms I: Fundamentals},
  volume 305.
\newblock Springer science \& business media, 2013.

\bibitem[Hu and et~al(2018)]{hu2018does}
Weihua Hu and et~al.
\newblock Does distributionally robust supervised learning give robust
  classifiers?
\newblock In \emph{ICML}, 2018.

\bibitem[Isensee et~al.(2021)Isensee, Jaeger, Kohl, Petersen, and
  Maier-Hein]{isensee2021nnu}
Fabian Isensee, Paul~F Jaeger, Simon~AA Kohl, Jens Petersen, and Klaus~H
  Maier-Hein.
\newblock nn{U-N}et: a self-configuring method for deep learning-based
  biomedical image segmentation.
\newblock \emph{Nature Methods}, 18\penalty0 (2):\penalty0 203--211, 2021.

\bibitem[Jin et~al.(2019)Jin, Netrapalli, and Jordan]{jin2019minmax}
Chi Jin, Praneeth Netrapalli, and Michael~I Jordan.
\newblock Minmax optimization: Stable limit points of gradient descent ascent
  are locally optimal.
\newblock \emph{arXiv preprint arXiv:1902.00618}, 2019.

\bibitem[Kahn and Marshall(1953)]{kahn1953methods}
Herman Kahn and Andy~W Marshall.
\newblock Methods of reducing sample size in {Monte} {Carlo} computations.
\newblock \emph{Journal of the Operations Research Society of America},
  1\penalty0 (5):\penalty0 263--278, 1953.

\bibitem[Larrazabal et~al.(2020)Larrazabal, Nieto, Peterson, Milone, and
  Ferrante]{larrazabal2020gender}
Agostina~J Larrazabal, Nicol{\'a}s Nieto, Victoria Peterson, Diego~H Milone,
  and Enzo Ferrante.
\newblock Gender imbalance in medical imaging datasets produces biased
  classifiers for computer-aided diagnosis.
\newblock \emph{Proceedings of the National Academy of Sciences}, 117\penalty0
  (23):\penalty0 12592--12594, 2020.

\bibitem[LeCun(1998)]{lecun1998mnist}
Yann LeCun.
\newblock The {MNIST} database of handwritten digits.
\newblock \emph{http://yann. lecun. com/exdb/mnist/}, 1998.

\bibitem[Lee et~al.(2015)Lee, Xie, Gallagher, Zhang, and Tu]{lee2015deeply}
Chen-Yu Lee, Saining Xie, Patrick Gallagher, Zhengyou Zhang, and Zhuowen Tu.
\newblock Deeply-supervised nets.
\newblock In \emph{Artificial intelligence and statistics}, pages 562--570,
  2015.

\bibitem[Lin et~al.(2019)Lin, Jin, and Jordan]{lin2019gradient}
Tianyi Lin, Chi Jin, and Michael~I Jordan.
\newblock On gradient descent ascent for nonconvex-concave minimax problems.
\newblock \emph{arXiv preprint arXiv:1906.00331}, 2019.

\bibitem[Loshchilov and Hutter(2016)]{loshchilov2015online}
Ilya Loshchilov and Frank Hutter.
\newblock Online batch selection for faster training of neural networks.
\newblock \emph{ICLR Workshop}, 2016.

\bibitem[Menze et~al.(2014)Menze, Jakab, Bauer, Kalpathy-Cramer, Farahani,
  Kirby, Burren, Porz, Slotboom, Wiest, et~al.]{menze2014multimodal}
Bjoern~H Menze, Andras Jakab, Stefan Bauer, Jayashree Kalpathy-Cramer, Keyvan
  Farahani, Justin Kirby, Yuliya Burren, Nicole Porz, Johannes Slotboom, Roland
  Wiest, et~al.
\newblock The multimodal brain tumor image segmentation benchmark (brats).
\newblock \emph{IEEE transactions on medical imaging}, 34\penalty0
  (10):\penalty0 1993--2024, 2014.

\bibitem[Moreau(1965)]{moreau1965proximite}
Jean-Jacques Moreau.
\newblock Proximit{\'e} et dualit{\'e} dans un espace hilbertien.
\newblock \emph{Bulletin de la Soci{\'e}t{\'e} math{\'e}matique de France},
  93:\penalty0 273--299, 1965.

\bibitem[Mufti et~al.(2021)Mufti, Aertsen, Ebner, Fidon, Patel, Rahman,
  Brackenier, Ekart, Fernandez, Vercauteren, et~al.]{mufti2021cortical}
Nada Mufti, Michael Aertsen, Michael Ebner, Lucas Fidon, Premal Patel, Muhamad
  Bin~Abdul Rahman, Yannick Brackenier, Gregor Ekart, Virginia Fernandez, Tom
  Vercauteren, et~al.
\newblock Cortical spectral matching and shape and volume analysis of the fetal
  brain pre-and post-fetal surgery for spina bifida: a retrospective study.
\newblock \emph{Neuroradiology}, pages 1--14, 2021.

\bibitem[Namkoong and Duchi(2016)]{namkoong2016stochastic}
Hongseok Namkoong and John~C Duchi.
\newblock Stochastic gradient methods for distributionally robust optimization
  with f-divergences.
\newblock In \emph{Advances in Neural Information Processing Systems}, pages
  2208--2216, 2016.

\bibitem[Oakden-Rayner et~al.(2020)Oakden-Rayner, Dunnmon, Carneiro, and
  R{\'e}]{oakden2020hidden}
Luke Oakden-Rayner, Jared Dunnmon, Gustavo Carneiro, and Christopher R{\'e}.
\newblock Hidden stratification causes clinically meaningful failures in
  machine learning for medical imaging.
\newblock In \emph{Proceedings of the ACM conference on health, inference, and
  learning}, pages 151--159, 2020.

\bibitem[Owen and Zhou(2000)]{owen2000safe}
Art Owen and Yi~Zhou.
\newblock Safe and effective importance sampling.
\newblock \emph{Journal of the American Statistical Association}, 95\penalty0
  (449):\penalty0 135--143, 2000.

\bibitem[Payette et~al.(2021)Payette, de~Dumast, Kebiri, Ezhov, Paetzold, Shit,
  Iqbal, Khan, Kottke, Grehten, et~al.]{payette2021automatic}
Kelly Payette, Priscille de~Dumast, Hamza Kebiri, Ivan Ezhov, Johannes~C
  Paetzold, Suprosanna Shit, Asim Iqbal, Romesa Khan, Raimund Kottke, Patrice
  Grehten, et~al.
\newblock An automatic multi-tissue human fetal brain segmentation benchmark
  using the fetal tissue annotation dataset.
\newblock \emph{Scientific Data}, 8\penalty0 (1):\penalty0 1--14, 2021.

\bibitem[Payette et~al.(2022)Payette, Li, de~Dumast, Licandro, Ji, Siddiquee,
  Xu, Myronenko, Liu, Pei, et~al.]{payette2022fetal}
Kelly Payette, Hongwei Li, Priscille de~Dumast, Roxane Licandro, Hui Ji,
  Md~Mahfuzur~Rahman Siddiquee, Daguang Xu, Andriy Myronenko, Hao Liu, Yuchen
  Pei, et~al.
\newblock Fetal brain tissue annotation and segmentation challenge results.
\newblock \emph{arXiv preprint arXiv:2204.09573}, 2022.

\bibitem[Puyol-Ant{\'o}n et~al.(2021)Puyol-Ant{\'o}n, Ruijsink, Piechnik,
  Neubauer, Petersen, Razavi, and King]{puyol2021fairness}
Esther Puyol-Ant{\'o}n, Bram Ruijsink, Stefan~K Piechnik, Stefan Neubauer,
  Steffen~E Petersen, Reza Razavi, and Andrew~P King.
\newblock Fairness in cardiac mr image analysis: An investigation of bias due
  to data imbalance in deep learning based segmentation.
\newblock In \emph{International Conference on Medical Image Computing and
  Computer-Assisted Intervention}, pages 413--423. Springer, 2021.

\bibitem[Rafique et~al.(2018)Rafique, Liu, Lin, and Yang]{rafique2018non}
Hassan Rafique, Mingrui Liu, Qihang Lin, and Tianbao Yang.
\newblock Non-convex min-max optimization: Provable algorithms and applications
  in machine learning.
\newblock \emph{arXiv preprint arXiv:1810.02060}, 2018.

\bibitem[Rahimian and Mehrotra(2019)]{rahimian2019distributionally}
Hamed Rahimian and Sanjay Mehrotra.
\newblock Distributionally robust optimization: A review.
\newblock \emph{arXiv preprint arXiv:1908.05659}, 2019.

\bibitem[Ranzini et~al.(2021)Ranzini, Fidon, Ourselin, Modat, and
  Vercauteren]{ranzini2021monaifbs}
Marta Ranzini, Lucas Fidon, S{\'e}bastien Ourselin, Marc Modat, and Tom
  Vercauteren.
\newblock {MONAI}fbs: {MONAI}-based fetal brain {MRI} deep learning
  segmentation.
\newblock \emph{arXiv preprint arXiv:2103.13314}, 2021.

\bibitem[Sacco et~al.(2019)Sacco, Ushakov, Thompson, Peebles, Pandya, De~Coppi,
  Wimalasundera, Attilakos, David, and Deprest]{sacco2019fetal}
Adalina Sacco, Fred Ushakov, Dominic Thompson, Donald Peebles, Pranav Pandya,
  Paolo De~Coppi, Ruwan Wimalasundera, George Attilakos, Anna~Louise David, and
  Jan Deprest.
\newblock Fetal surgery for open spina bifida.
\newblock \emph{The Obstetrician \& Gynaecologist}, 21\penalty0 (4):\penalty0
  271, 2019.

\bibitem[Sagawa et~al.(2020)Sagawa, Koh, Hashimoto, and
  Liang]{sagawa2019distributionally}
Shiori Sagawa, Pang~Wei Koh, Tatsunori~B Hashimoto, and Percy Liang.
\newblock Distributionally robust neural networks for group shifts: On the
  importance of regularization for worst-case generalization.
\newblock \emph{ICLR}, 2020.

\bibitem[Shrivastava et~al.(2016)Shrivastava, Gupta, and
  Girshick]{shrivastava2016training}
Abhinav Shrivastava, Abhinav Gupta, and Ross Girshick.
\newblock Training region-based object detectors with online hard example
  mining.
\newblock In \emph{Proceedings of the IEEE conference on computer vision and
  pattern recognition}, pages 761--769, 2016.

\bibitem[Sinha et~al.(2018)Sinha, Namkoong, and Duchi]{sinha2017certifying}
Aman Sinha, Hongseok Namkoong, and John Duchi.
\newblock Certifying some distributional robustness with principled adversarial
  training.
\newblock \emph{ICLR}, 2018.

\bibitem[Staib and Jegelka(2017)]{staib2017distributionally}
Matthew Staib and Stefanie Jegelka.
\newblock Distributionally robust deep learning as a generalization of
  adversarial training.
\newblock In \emph{{NIPS} workshop on Machine Learning and Computer Security},
  2017.

\bibitem[Suh et~al.(2019)Suh, Han, Kim, and Lee]{suh2019stochastic}
Yumin Suh, Bohyung Han, Wonsik Kim, and Kyoung~Mu Lee.
\newblock Stochastic class-based hard example mining for deep metric learning.
\newblock In \emph{Proceedings of the IEEE Conference on Computer Vision and
  Pattern Recognition}, pages 7251--7259, 2019.

\bibitem[Tilborghs et~al.(2020)Tilborghs, Dirks, Fidon, Willems, Eelbode,
  Bertels, Ilsen, Brys, Dubbeldam, Buls, et~al.]{tilborghs2020comparative}
Sofie Tilborghs, Ine Dirks, Lucas Fidon, Siri Willems, Tom Eelbode, Jeroen
  Bertels, Bart Ilsen, Arne Brys, Adriana Dubbeldam, Nico Buls, et~al.
\newblock Comparative study of deep learning methods for the automatic
  segmentation of lung, lesion and lesion type in {CT} scans of {COVID-19}
  patients.
\newblock \emph{arXiv preprint arXiv:2007.15546}, 2020.

\bibitem[Tubbs et~al.(2011)Tubbs, Krishnamurthy, Verma, Shoja, Loukas,
  Mortazavi, and Cohen-Gadol]{tubbs2011cavum}
R~Shane Tubbs, Sanjay Krishnamurthy, Ketan Verma, Mohammadali~M Shoja, Marios
  Loukas, Martin~M Mortazavi, and Aaron~A Cohen-Gadol.
\newblock Cavum velum interpositum, cavum septum pellucidum, and cavum vergae:
  a review.
\newblock \emph{Child's Nervous System}, 27\penalty0 (11):\penalty0 1927--1930,
  2011.

\bibitem[Ulyanov et~al.(2016)Ulyanov, Vedaldi, and
  Lempitsky]{ulyanov2016instance}
Dmitry Ulyanov, Andrea Vedaldi, and Victor Lempitsky.
\newblock Instance normalization: The missing ingredient for fast stylization.
\newblock \emph{arXiv preprint arXiv:1607.08022}, 2016.

\bibitem[Wachinger et~al.(2019)Wachinger, Becker, Rieckmann, and
  P{\"o}lsterl]{wachinger2019quantifying}
Christian Wachinger, Benjamin~Gutierrez Becker, Anna Rieckmann, and Sebastian
  P{\"o}lsterl.
\newblock Quantifying confounding bias in neuroimaging datasets with causal
  inference.
\newblock In \emph{International Conference on Medical Image Computing and
  Computer-Assisted Intervention}, pages 484--492. Springer, 2019.

\bibitem[Wu et~al.(2017)Wu, Manmatha, Smola, and Krahenbuhl]{wu2017sampling}
Chao-Yuan Wu, R~Manmatha, Alexander~J Smola, and Philipp Krahenbuhl.
\newblock Sampling matters in deep embedding learning.
\newblock In \emph{Proceedings of the IEEE International Conference on Computer
  Vision}, pages 2840--2848, 2017.

\bibitem[Zagoruyko and Komodakis(2016)]{zagoruyko2016wide}
Sergey Zagoruyko and Nikos Komodakis.
\newblock Wide residual networks.
\newblock In \emph{Proceedings of the British Machine Vision Conference
  (BMVC)}, pages 87.1--87.12. BMVA Press, 2016.

\bibitem[Zou and Gu(2019)]{zou2019improved}
Difan Zou and Quanquan Gu.
\newblock An improved analysis of training over-parameterized deep neural
  networks.
\newblock In \emph{Advances in Neural Information Processing Systems 32}, pages
  2055--2064. Curran Associates, Inc., 2019.

\end{thebibliography}

\unhidefromtoc 

\newpage
\appendix 

\tableofcontents
\newpage

\section{Summary of the Notations used in the Proofs}\label{s:notations}

For the ease of reading the proofs we first summarize our notations.

\subsection{Probability Theory Notations}
\begin{itemize}
    \item $\Delta_n = \left\{\left(p_i\right)_{i=1}^n \in [0,1]^n, \,\,
        \sum_i p_i = 1\right\}$
    \item Let $\vq=(q_i) \in \Delta_n$, and $f$ a function, we denote $\E_{\vq}[f(\vx)]:=\sum_{i=1}^n q_i f(\vx_i)$.
    \item Let $\vq \in \Delta_n$, and $f$ a function, we denote $\V_{\vq}[f(\vx)]:=\sum_{i=1}^n q_i\norm{f(\vx_i) - \E_q[f(\vx)]}^2$.
    \item $\vp_{\rm{train}}$ is the uniform training data distribution, i.e. $\vp_{\rm{train}}=\left(\frac{1}{n}\right)_{i=1}^n \in \Delta_n$.
\end{itemize}

\subsection{Machine Learning Notations}
\begin{itemize}
    \item n is the number of training examples.
    \item d is the dimension of the output.
    \item $\gd$ is the dimension of the input.
    \item m is the number of nodes in each layer.
    \item Training data: $\{(\vx_i, \vy_i)\}_{i=1}^n$, where for all $i\in \{1,\ldots,n\}$, $\vx_i \in \sR^{\gd}$ and $\vy_i \in \sR^d$.
    \item $h: \vx \mapsto \vy$ is the predictor (deep neural network).
    \item $\vtheta$ is the set of parameters of the predictor.
    \item For all $i$, $h_i: \vtheta \mapsto h(\vx_i;\vtheta)$ is the output of the network for example $i$ as a function of $\vtheta$.
    \item $\mathcal{L}$ is the \textcolor{black}{per-example loss} function.
    \item $\mathcal{L}_i: \vv \mapsto \mathcal{L}(\vv, \vy_i)$ is the \textcolor{black}{per-example loss} function for example $i$.
    \item We denote by $\vL$ the vector-valued function $\vL: (\vv_i)_{i=1}^n \mapsto (\cL_i(\vv_i))_{i=1}^n$.
    \item $b \in \{1, \ldots, n\}$ is the batch size.
    \item $\eta > 0$ is the learning rate.
    \item ERM is short for Empirical Risk Minimization.
\end{itemize}

\subsection{Distributionally Robust Optimisation Notations}

\begin{itemize}
    \item Forall $\vtheta$, $R(\vL(h(\vtheta))) 
        = \max_{\vq \in \Delta_n} \E_{\vq} \left[\cL\left(h(\vx; \vtheta), \vy\right)\right]
        - \frac{1}{\de} D_{\phi}(\vq \Vert \vp_{\rm{train}})$ is the \textbf{Distributionally Robust Loss} evaluated at $\vtheta$, where $\de>0$ is the parameter that adjusts the distributionally robustness.
        For short, we also used the terms \textbf{distributionally robust loss} or just \textbf{robust loss} for $R(\vL(h(\vtheta)))$.
    \item DRO is short for Distributionally Robust Optimisation.
\end{itemize}

\subsection{Miscellaneous}
By abuse of notation, and similarly to \citep{allen-zhu19a}, we use the Bachmann-Landau notations to hide constants that do not depend on our main hyper-parameters.
Let $f$ and $g$ be two scalar functions, we note:
\[
\left\{
    \begin{array}{ccccc}
        f \leq O(g)      & \iff & \exists c > 0 & \textup{ s.t. } & f \leq c g\\
        f \geq \Omega(g) & \iff & \exists c > 0 & \textup{ s.t. } & f \geq c g\\
        f = \Theta(g)    & \iff & \exists c_1 > 0 \textup{ and } \exists c_2 > c_1 & \textup{ s.t. } & c_1 g \leq f \leq c_2 g\\
    \end{array}
\right.
\]

\section{Evaluation of the Influence of $\de$ on the Segmentation Performance for BraTS}\label{s:more_results_brats}

\begin{table}[H]
	\centering
	\caption{\textbf{Detailed evaluation on the BraTS 2019 online validation set (125 cases).}
	All the models in this table were trained using the default \textbf{SGD with Nesterov momentum} of nnU-Net~\citep{isensee2021nnu}.
	Dice scores were computed using the BraTS online plateform for evaluation \protect\url{https://ipp.cbica.upenn.edu/}.
	ERM: Empirical Risk Minimization,
	DRO: Distributionally Robust Optimization,
	IS: Importance Sampling is used,
	IQR: Interquartile range.
    The best values are in bold.
	}
	\begin{tabularx}{\textwidth}{c *{9}{Y}}
		\toprule
        \textbf{Optimization}
        & \multicolumn{3}{c}{Enhancing Tumor} 
        & \multicolumn{3}{c}{Whole Tumor}
        & \multicolumn{3}{c}{Tumor Core}\\
        \cmidrule(lr){2-4} \cmidrule(lr){5-7} \cmidrule(lr){8-10}
		\textbf{problem} & Mean & Median & IQR & Mean & Median & IQR & Mean & Median & IQR\\ 
	\midrule
		ERM 
		 & 73.0 & 87.1 & 15.6
		 & 90.7 & 92.6 & \bf5.4
		 & 83.9 & 90.5 & 14.3\\
	\cmidrule(lr){1-10}
        DRO $\de=10$
         & 74.6 & 86.8 & 14.1
         & \bf90.8 & \bf93.0 & 5.9
         & 83.4 & 90.7 & 14.5\\
    \cmidrule(lr){1-10}
        DRO $\de=10$ IS
         & \bf75.3 & 86.0 & \bf13.3
         & 90.0 & 91.9 & 7.0
         & 82.8 & 89.1 & 14.3\\
	\cmidrule(lr){1-10}
        DRO $\de=100$ 
         & 73.4 & 86.7 & 14.3
         & 90.6 & 92.6 & 6.2
         & \bf84.5 & \bf90.9 & 13.7\\
	\cmidrule(lr){1-10}
        DRO $\de=100$ IS
         & 74.5 & \bf87.3 & 13.8
		 & 90.6 & 92.6 & 5.9
		 & 84.1 & 90.0 & \bf12.5\\
	\cmidrule(lr){1-10}
        DRO $\de=1000$
         & 74.5 & 84.2 & 33.0
         & 89.5 & 91.8 & 5.9
         & 71.1 & 87.2 & 41.1\\
    \cmidrule(lr){1-10}
        DRO $\de=1000$ IS
         & 72.2 & 85.7 & 15.0
         & 90.3 & 92.2 & 6.3
         & 81.1 & 89.4 & 15.1\\
	\bottomrule
	\end{tabularx}
	\label{tab:models_all_results}
\end{table}
\section{Importance Sampling Approximation in \Algref{alg:1}}\label{appendix:importance_sampling}

In this section, we give additional details about the approximation made in the computation of the importance weights (step 9 of \Algref{alg:1}).

Let $\vtheta$ be the parameters of the neural network $h$, $\vL=\left(L_i\right)_{i=1}^n$ be the stale per-example loss vector, and let $i$ be an index in the current batch $I$.

We start from the definition of the importance weight $w_i$ for example $i$ and use the formula for the hardness weighted sampling probabilities of Example~\ref{ex:softmax}.

\begin{equation}
    \begin{aligned}
        w_i &= \frac{p_i^{new}}{p_i^{old}}\\
            &= \frac{\exp\left(\de L_i^{new}\right)}{\exp\left(\de L_i^{new}\right) + \sum_{j \neq i} \exp\left(\de L_j^{old}\right)}
            \times
            \frac{\sum_{j=1}^n \exp\left(\de L_j^{old}\right)}{\exp\left(\de L_i^{old}\right)}\\
            & \approx \exp\left(\de (L_i^{new} - L_{i}^{old})\right)
    \end{aligned}
\end{equation}

where we have assumed that the two sums of exponentials are approximately equal.
\section{Proof of Example \ref{ex:softmax}: Formula of the Sampling Probabilities for the KL Divergence}\label{s:proof_softmax}

We give here a simple proof of the formula of the sampling probabilities for the KL divergence as $\phi$-divergence (i.e. $\phi: z \mapsto z \log(z) - z + 1$)
\[
\forall \vtheta, \quad 
\bar{p}(\vL(h(\vtheta))) 
= \softmax\left(\de \vL(h(\vtheta))\right)
\]

\paragraph*{Proof:}
For any $\vtheta$, the distributionally robust loss for the KL divergence at $\vtheta$ is given by
\[
\begin{aligned}
    R \circ \vL \circ h (\vtheta)
    &= \max_{\vq \in \Delta_n}
    \left(
    \sum_{i=1}^n q_i \textcolor{black}{\mathcal{L}_i} \circ ~h_i (\vtheta)
    - \frac{1}{\de}\sum_{i=1}^n q_i\log\left(n q_i\right)
    \right)\\
    &= \max_{\vq \in \Delta_n}
    \sum_{i=1}^n \left(q_i \textcolor{black}{\mathcal{L}_i} \circ ~h_i (\vtheta)
    - \frac{1}{\de} q_i\log\left(n q_i\right)
    \right)
\end{aligned}
\]
where we have used that $\frac{1}{p_{\rm{train},i}} = n$ inside the $\log$ function.
To simplify the notations, let us denote 
$\vv=(v_i)_{i=1}^n
:=\left(\cL_i \circ ~h_i (\vtheta)\right)_{i=1}^n$,
and $\bar{\vp} = (\bar{p}_i)_{i=1}^n := \bar{\textbf{p}}(\vL(h(\vtheta)))$.
Thus $\bar{\textbf{p}}(\vL(\vh(\vtheta)))$ is, by definition, solution of the optimization problem
\begin{equation}
    \label{eq:optim_softmax_1}
    \textcolor{black}{\max}_{\vq \in \Delta_n}
    \sum_{i=1}^n \left(q_i v_i
    - \frac{1}{\de} q_i\log\left(n q_i\right)
    \right)
\end{equation}
First, let us remark that the function $q \mapsto \sum_{i=1}^n q_i\log\left(n q_i\right)$ is strictly convex on the non empty closed convex set $\Delta_n$ as a sum of strictly convex functions.
This implies that the optimization \eqref{eq:optim_softmax_1} has a unique solution and as a result $\textcolor{black}{\bar{\textbf{p}}}(\vL(h(\vtheta)))$ is well defined.

We now reformulate the optimization problem \eqref{eq:optim_softmax_1} as a convex smooth constrained optimization problem by writing the condition $\vq \in \Delta_n$ as constraints.
\begin{equation}
    \label{eq:optim_softmax_2}
    \begin{aligned}
        \textcolor{black}{\max}_{\vq \in \sR^n_+}&
    \sum_{i=1}^n \left(q_i v_i
    - \frac{1}{\de} q_i\log\left(n q_i\right)
    \right)\\
    \text{s.t.}& \sum_{i=1}^n q_i =1 
    \end{aligned}
\end{equation}

There exists a Lagrange multiplier $\lambda \in \sR$, such that the solution $\bar{p}$ of \eqref{eq:optim_softmax_2} is characterized by
\begin{equation}
    \label{eq:optim_softmax_3}
    \left\{
    \begin{aligned}
        \forall i \in \{1, \ldots, n\},\quad & v_i
    - \frac{1}{\de} \left(
        \log\left(n \bar{p}_i\right) + 1
    \right) + \lambda = 0\\
    & \sum_{i=1}^n \bar{p}_i =1 
    \end{aligned}
    \right.
\end{equation}
Which we can rewrite as
\begin{equation}
    \label{eq:optim_softmax_4}
    \left\{
    \begin{aligned}
        \forall i \in \{1, \ldots, n\},\quad & 
        \bar{p}_i = \frac{1}{n} \exp\left(
        \de \left(v_i + \lambda\right) - 1
        \right)\\
    & \frac{1}{n} \sum_{i=1}^n \exp\left(
        \de \left(v_i + \lambda\right) - 1
        \right) = 1 
    \end{aligned}
    \right.
\end{equation}

The last equality gives
\[
    \exp\left(\de \lambda - 1\right) 
        = \frac{n}{\sum_{i=1}^n \exp\left(\de v_i\right)}
\]
And by replacing in the formula of the $\bar{p}_i$
\[
\begin{aligned}
    \forall i \in \{1, \ldots, n\},\quad
        \bar{p}_i =&
        \frac{1}{n} 
        \exp\left(\de v_i\right)
        \exp\left(
        \de \lambda - 1
        \right)\\
        =&\frac{\exp\left(\de v_i\right)}{{\sum_{j=1}^n \exp\left(\de v_j\right)}}
\end{aligned}
\]
Which corresponds to
$
\bar{\vp} = \softmax\left(\de \vv\right)
\,\,\blacksquare
$

\section{Proof of Lemma \ref{lemma:R_property}: Regularity Properties of R}\label{s:regularity_R}

For the ease of reading, let us first recall that given a $\phi$-divergence $D_{\phi}$ that satisfies Definition~\ref{def:phi_divergence}, we have defined in \eqref{eq:dro}
\begin{equation}
    \begin{aligned}
        R:\,\,&\sR^n \rightarrow \sR\\
             & \vv \,\mapsto \max_{\vq \in \Delta_n} \sum_i q_i v_i
        - \frac{1}{\de} D_{\phi}(\vq \Vert \vp_{\rm{train}})
    \end{aligned}
\end{equation}
And in \eqref{eq:G_def}
\begin{equation}
    \begin{aligned}
        G:\,\,&\sR^n \rightarrow \sR\\
             & \vp \,\mapsto \frac{1}{\de} D_{\phi}(\vp \Vert \vp_{\rm{train}}) + \delta_{\Delta_n}(\vp)
    \end{aligned}
\end{equation}
where $\delta_{\Delta_n}$ is the characteristic function of the \textcolor{black}{to the $n$-simplex $\Delta_n$ which is a} closed convex set, i.e.
\begin{equation}
    \forall \vp \in \sR^n,\,\, \delta_{\Delta_n}(\vp)=\left\{
\begin{array}{cl}
    0 & \text{if } \vp \in \Delta_n \\
    +\infty & \text{otherwise}
\end{array}
\right.
\end{equation}

We now prove Lemma \ref{lemma:R_property} on the regularity of $R$.

\begin{lemma}[Regularity of R -- Restated from Lemma \ref{lemma:R_property}]
\label{lemma:R_property_re}
    Let $\phi$ that satisfies Definition~\ref{def:phi_divergence}, $G$ and $R$ satisfy
    \begin{equation}
        \label{eq:strong_convexity_G_re}
        G \text{ is} \left(\frac{n\rho}{\de}\right)~\text{-strongly convex}
    \end{equation}
    \begin{equation}
        \label{eq:link_R_and_G_re}
        R(\vL(h(\vtheta)))
        =
        \max_{\vq \in \sR^n} 
            \left(
            \langle \vL(h(\vtheta)), \vq\rangle - G(\vq)
            \right)
        = G^*\left(\vL(h(\vtheta))\right)
    \end{equation}
    \begin{equation}
        \label{eq:gradient_Lip_R_re}
        R \text{ is } \left(\frac{\de}{n\rho}\right)~\text{-gradient Lipschitz continuous.}
    \end{equation}
\end{lemma}

\paragraph*{Proof:}
$\phi$ is $\rho$-strongly convex on $[0,n]$ so
\begin{equation}
	\label{eq:strong-conv}
	\forall x,y \in [0, n]^2, \forall \lambda \in [0,1],
	\phi\left(\lambda x + (1 - \lambda)y \right) \leq
	\lambda \phi(x) + (1-\lambda)\phi(y) - \frac{\rho \lambda(1-\lambda)}{2} |y-x|^2
\end{equation}
Let $\vp=\left(p_i\right)_{i=1}^n$, $\vq=\left(q_i\right)_{i=1}^n \in \Delta_n$, and $\lambda \in [0,1]$,
using \eqref{eq:strong-conv} and the convexity of $\delta_{\Delta_n}$, we obtain:
\begin{equation}
	\begin{aligned}
	G\left(\lambda \vp + (1 - \lambda)\vq \right) &
	= \frac{1}{\de n} \sum_{i=1}^n \phi\left(n\lambda p_i + n(1 - \lambda)q_i \right) + \delta_{\Delta_n}\left(\lambda \vp + (1 - \lambda)\vq \right)\\
	&\leq \lambda G(\vp) + (1 - \lambda)G(\vq) - \frac{1}{\de n}\sum_{i=1}^n\frac{\rho \lambda(1-\lambda)}{2} |np_i-nq_i|^2\\
	&\leq \lambda G(\vp) + (1 - \lambda)G(\vq) - \frac{n \rho}{\de}\frac{\lambda(1 - \lambda)}{2} \norm{\vp - \vq}^2
	\end{aligned}
\end{equation}

This proves that $G$ is $\frac{n\rho}{\de}$-strongly convex.

\textcolor{black}{
$R=G^*$ is convex, and since $G$ is closed and convex, $R^*=\left(G^*\right)^*=G$~\citep{hiriart2013convex}.
}
%
We obtain \eqref{eq:link_R_and_G_re} using Definition \ref{def:fenchel_conjugate}.

We now show that $R$ is Frechet differentiable on $\sR^n$.
Let $\vv \in \sR^n$.

$G$ is strongly-convex, so in particular $G$ is strictly convex.
This implies that the following optimization problem has a unique solution that we denote $\hat{\vp}(\vv)$.
\begin{equation}
    \label{eq:max_pb_R_grad_lip}
    \hat{\vp}(\vv) := 
    \argmax_{\vq \in \sR^n} 
            \left(
            \langle \vv, \vq\rangle - G(\vq)
            \right)
\end{equation}

In addition, \textcolor{black}{using the notion of subderivative of convex functions \citep[Definition 4.1.5 p.39]{hiriart2013convex}, we have}
\[
\begin{aligned}
    \hat{\vp} \in \Delta_n \text{ solution of \eqref{eq:max_pb_R_grad_lip} }
    & \Longleftrightarrow 0 \in \vv - \partial G(\hat{\vp})\\
    & \Longleftrightarrow \vv \in \partial G(\hat{\vp})\\
    & \Longleftrightarrow \hat{\vp} \in \partial G^*(\vv) \\
    & \Longleftrightarrow \hat{\vp} \in \partial R(\vv)
\end{aligned}
\]
where we have used \citep[Proposition 6.1.2 p.39]{hiriart2013convex} for the third equivalence, and \eqref{eq:link_R_and_G_re} for the last equivalence.

As a result, $\partial R(\vv)=\{\hat{\vp}(\vv)\}$.
\textcolor{black}{T}his implies that $R$ admit a gradient at $\vv$, and
\begin{equation}
    \label{eq:R_differentiable}
    \nabla_{\vv} R(\vv) = \hat{\vp}(\vv)
\end{equation}

Since this holds for any $\vv \in \sR^n$, we deduce that $R$ is \textcolor{black}{Fr\'echet} differentiable on $\sR^n$. $\blacksquare$

We are now ready to show that $R$ is $\frac{\de}{n\rho}$-gradient Lipchitz continuous by using the following lemma \citep[Theorem 6.1.2 p.280]{hiriart2013convex}.

\begin{lemma}
    A necessary and sufficient condition for a convex function $f\,: \sR^n \rightarrow \sR$ to be $c$-strongly convex on a convex set $C$ is that for all $x_1,x_2 \in C$
    \[
    \langle
    s_2 - s_1, x_2 - x_1
    \rangle
    \geq c \norm{x_2 - x_1}^2 \quad \text{for all } s_i \in \partial f(x_i), i=1,2.
    \]
\end{lemma}
Using this lemma for $f=G$, $c=\frac{n\rho}{\de}$, and $C=\Delta_n$, we obtain:

For all $\vp_1, \vp_2 \in \Delta_n$, for all $\vv_1 \in \partial G(\vp_1)$, $\vv_2 \in \partial G(\vp_2)$,
\[
\langle
\vv_2 - \vv_1, \vp_2 - \vp_1
\rangle
\geq \frac{n\rho}{\de} \norm{\vp_2 - \vp_1}^2
\]
In addition, for $i \in \{1,\,2\}$, $\vv_i \in \partial G(\vp_i) \Longleftrightarrow \vp_i \in \partial R(\vv_i)= \{\nabla_{\vv} R(\vv_i)\}$.

And using Cauchy Schwarz inequality
\[
\norm{\vv_2 - \vv_1} \norm{\vp_2 - \vp_1}
\geq
\langle
\vv_2 - \vv_1, \vp_2 - \vp_1
\rangle
\]
We conclude that
\[
\frac{n\rho}{\de} \norm{\nabla_{\vv} R(\vv_2) - \nabla_{\vv} R(\vv_1)} \leq \norm{\vv_2 - \vv_1}
\]
Which implies that $R$ is $\frac{\de}{n\rho}$-gradient Lipchitz continuous. $\blacksquare$
\section{Proof of Lemma \ref{lemma:robust_loss_sg}: Formula of the Distributionally Robust Loss Gradient}\label{s:robust_loss_sg}
We prove Lemma \ref{lemma:robust_loss_sg} that we restate here for the ease of reading.
\begin{lemma}[Stochastic Gradient of the DRO Loss -- Restated from Lemma \ref{lemma:robust_loss_sg}]
\label{lemma:robust_loss_sg_re}
For all $\vtheta$, we have
    \begin{equation}
        \label{eq:p_re}
        \bar{p}(\vL(h(\vtheta))) = \nabla_{\vv} R(\vL(h(\vtheta)))
    \end{equation}
    \begin{equation}
        \label{eq:sg_re}
        \nabla_\vtheta (R \circ \vL \circ h)(\vtheta) 
        = \E_{\bar{\vp}(\cL(\vh(\vtheta)))}\left[\nabla_\vtheta \cL\left(h(\rx; \vtheta), y\right)\right]
    \end{equation}
\end{lemma}
where $\nabla_\vv R$ is the gradient of $R$ with respect to its input.

\paragraph*{Proof:}
For a given $\vtheta$,
equality \eqref{eq:p_re} is a special case of \eqref{eq:R_differentiable} for $\vv = \cL(\vh(\vtheta))$.

Then using the chain rule and \eqref{eq:p_re},
\[
\begin{aligned}
    \nabla_{\vtheta} (R \circ \vL \circ h)(\vtheta)
    &= \sum_{i=1}^n \frac{\partial R}{\partial v_i}(\vL \circ h(\vtheta))) \nabla_{\vtheta} (\textcolor{black}{\mathcal{L}_i} \circ h_i)(\vtheta)\\
    &= \sum_{i=1}^n \bar{p}_i(\vL(h(\vtheta))) \nabla_{\vtheta} (\textcolor{black}{\mathcal{L}_i} \circ h_i)(\vtheta)\\
    &= \E_{\bar{\vp}(\cL(\vh(\vtheta)))}\left[\nabla_\vtheta \cL\left(h(\rx; \vtheta), y\right)\right]
\end{aligned}
\]
Which concludes the proof. $\blacksquare$
\section{Proof of Theorem \ref{th:hard_example_mining}: Distributionally Robust Optimization as Principled Hard Example Mining}\label{s:hard_example_mining}

In this section, we demonstrate that the proposed hardness weighted sampling can be interpreted as a principled hard example mining method.

Let $D_{\phi}$ an $\phi$-divergence satisfying Definition~\ref{def:phi_divergence},
and $\vv=\left(v_i\right)_{i=1}^n \in \sR^n$.
$\vv$ will play the role of a generic loss vector.

$\phi$ is strongly convex, and $\Delta_n$ is closed and convex, so the following optimization problem has one and only one solution
\begin{equation}
    \max_{\vp=\left(p_i\right)_{i=1}^n \in \Delta_n} \langle \vv, \vp \rangle - \frac{1}{\de n}\sum_{i=1}^n \phi(n p_i)
\end{equation}
Making the constraints associated with $p \in \Delta_n$ explicit, this can be rewritten as
\begin{equation}
\begin{aligned}
    \max_{\vp=\left(p_i\right)_{i=1}^n \in \sR^n} &\langle \vv, \vp \rangle - \frac{1}{\de n}\sum_{i=1}^n \phi(n p_i)\\
    \text{ s.t. } &\forall i \in \{1, \ldots, n\},\,\, p_i \geq 0\\
    \text{ s.t. } &\sum_{i=1}^n p_i = 1
\end{aligned}
\end{equation}
There exists KKT multipliers $\lambda \in \sR$ and $\forall i,\, \mu_i \geq 0$ such that the solution $\bar{\vp}=\left(\bar{p}_i\right)_{i=1}^n$ satisfies
\begin{equation}
\label{eq:hard_proof}
    \left\{
    \begin{aligned}
        \forall i \in \{1, \ldots, n\}, \quad& v_i - \frac{1}{\de}\phi'(n\bar{p}_i) + \lambda - \mu_i = 0\\
        \forall i \in \{1, \ldots, n\},\quad& \mu_i p_i = 0\\
        \forall i \in \{1, \ldots, n\},\quad& p_i \geq 0\\
        & \sum_{i=1}^n \bar{p}_i = 1
    \end{aligned}
    \right.
\end{equation}
Since $\phi$ is continuously differentiable and strongly convex, we have $\left(\phi'\right)^{-1} = \left(\phi^*\right)'$, where $\phi^*$ is the Fenchel conjugate of $\phi$~\citep[see][Proposition 6.1.2]{hiriart2013convex}.
As a result, \eqref{eq:hard_proof} can be rewritten as
\begin{equation}
\label{eq:hard_proof_1}
    \left\{
    \begin{aligned}
        \forall i \in \{1, \ldots, n\}, \quad& \bar{p}_i = \frac{1}{n}\left(\phi^*\right)'\left(\de(v_i + \lambda - \mu_i)\right)\\
        \forall i \in \{1, \ldots, n\},\quad& \mu_i p_i = 0\\
        \forall i \in \{1, \ldots, n\},\quad & p_i \geq 0\\
        & \frac{1}{n}\sum_{i=1}^n \left(\phi^*\right)'\left(\de(v_i + \lambda - \mu_i)\right) = 1
    \end{aligned}
    \right.
\end{equation}
We now show that the KKT multipliers are uniquely defined.
\paragraph{\textbf{The $\mu_i$'s are uniquely defined by $\vv$ and $\lambda$}:}\ \\
Since $\forall i \in \{1, \ldots, n\},\,\, \mu_i p_i = 0,\,\, p_i \geq 0$ and $\mu_i \geq 0$,
for all $\forall i \in \{1, \ldots, n\}$, either $p_i=0$ or $\mu_i=0$.
In the case $p_i=0$ and using \eqref{eq:hard_proof_1} it comes $\left(\phi^*\right)'\left(\de(v_i + \lambda - \mu_i)\right)=0$.

According to Definition~\ref{def:phi_divergence}, $\phi$ is strongly convex and continuously differentiable, so $\phi'$ and $(\phi^*)'=(\phi')^{-1}$ are continuous and strictly increasing functions.
As a result, it exists a unique $\mu_i$ (dependent to $\vv$ and $\lambda$) such that:
\[
\left(\phi^*\right)'\left(\de(v_i + \lambda - \mu_i)\right)=0
\]
And \eqref{eq:hard_proof_1} can be rewritten as
\begin{equation}
\label{eq:hard_proof_2}
    \left\{
    \begin{aligned}
        \forall i \in \{1, \ldots, n\}, \quad& \bar{p}_i = \relu\left( \frac{1}{n}\left(\phi^*\right)'\left(\de(v_i + \lambda)\right)
        \right)
        =\frac{1}{n}\relu\left(\left(\phi^*\right)'\left(\de(v_i + \lambda)\right)
        \right)\\
        & \frac{1}{n}\sum_{i=1}^n \relu\left(\left(\phi^*\right)'\left(\de(v_i + \lambda)\right)\right) = 1\\
    \end{aligned}
    \right.
\end{equation}

\paragraph{\textbf{The KKT multiplier $\lambda$ is uniquely defined by $\vv$ and a continuous function of $\vv$}:}\ \\
Let $\lambda \in \sR$ that satisfies \eqref{eq:hard_proof_2}.
We have $\frac{1}{n}\sum_{i=1}^n \relu\left(\left(\phi^*\right)'\left(\de(v_i + \lambda)\right)\right) = 1$.
So there exists at least one index $i_0$ such that
\[
\relu\left(\left(\phi^*\right)'\left(\de(v_{i_0} + \lambda)\right)\right)=\left(\phi^*\right)'\left(\de(v_{i_0} + \lambda)\right) \geq 1
\]
Since $(\phi^*)^{-1}$ is continuous and striclty increasing, 
$\lambda'\mapsto \relu\left(\left(\phi^*\right)'\left(\de(v_{i_0} + \lambda')\right)\right)$ is continuous and strictly increasing on a neighborhood of $\lambda$.
In addition $\relu$ is continuous and increasing,
so for all $i \in \{1,\ldots,n\}$, 
$\lambda'\mapsto \relu\left(\left(\phi^*\right)'\left(\de(v_{i} + \lambda')\right)\right)$ is a continuous and increasing function.

As a result, $\lambda'\mapsto \frac{1}{n}\sum_{i=1}^n \relu\left(\left(\phi^*\right)'\left(\de(v_i + \lambda')\right)\right)$ is a continuous function that is increasing on $\sR$, and strictly increasing on a neighborhood of $\lambda$.
This implies that $\lambda$ is uniquely defined by $\vv$, and that $\vv \mapsto \lambda(\vv)$ is continuous.

\subsection{Link between Hard Weighted Sampling and Hard Example Mining}

For any pseudo loss vector $\vv=\left(v_i\right)_{i=1}^n \in \sR^n$, there exists a unique KKT multiplier $\lambda$ and a unique $\bar{\vp}$ that satisfies \eqref{eq:hard_proof_2}, so we can define the mapping:
\begin{equation}
    \begin{aligned}
        \bar{\vp}:\,\,\sR^n &\rightarrow \Delta_n\\
                    \vv & \mapsto \bar{\vp}(\vv;\lambda(\vv))\\
    \end{aligned}
\end{equation}
where for all $\vv$, $\lambda(\vv)$ is the unique $\lambda \in \sR$ satisfying \eqref{eq:hard_proof_2}.

We will now demonstrate that each $\bar{\vp}_{i_0}(\vv)$ for $i_0 \in \{1,\ldots, n\}$ is an increasing function of $v_i$ and a decreasing function of the $v_i$ for $i\neq i_0$.
Without loss of generality we assume $i_0=1$.

Let $\vv=\left(v_i\right)_{i=1}^n \in \sR^n$, and $\epsilon >0$.
Let us define $\vv'=\left(v_i'\right)_{i=1}^n \in \sR^n$, such that $v_1'=v_1+\epsilon$ and $\forall i \in \{2,\ldots, n\},\,\, v_i'=v_i$.
Similarly as in the proof of the uniqueness of $\lambda$ above, we can show that there exists $\eta > 0$ such that the function
\[
F:\lambda'\mapsto \frac{1}{n}\sum_{i=1}^n \relu\left(\left(\phi^*\right)'\left(\de(v_i + \lambda')\right)\right)
\]
is continuous and strictly increasing on $[\lambda(\vv)-\eta, \lambda(\vv)+\eta]$, and $F(\lambda(\vv))=1$.

$v \mapsto \lambda(\vv)$ is continuous, so for $\epsilon$ small enough $\lambda(\vv') \in [\lambda(v)-\eta, \lambda(\vv)+\eta]$.

Let us now prove by contradiction that $\lambda(\vv') \leq \lambda(\vv)$. Therefore,
let us assume that $\lambda(\vv') > \lambda(\vv)$. Then, as $\relu \circ \left(\phi^*\right)'$ is an increasing function and $F$ is strictly increasing on $[\lambda(\vv)-\eta, \lambda(\vv)+\eta]$, and $\epsilon > 0$ we obtain
\[
\begin{aligned}
    1 &= \frac{1}{n}\sum_{i=1}^n \relu\left(\left(\phi^*\right)'\left(\de(v_i' + \lambda(\vv'))\right)\right)\\
    & \geq \frac{1}{n}\sum_{i=1}^n \relu\left(\left(\phi^*\right)'\left(\de(v_i + \lambda(\vv'))\right)\right)\\
    & \geq F(\lambda(\vv'))\\
    & > F(\lambda(\vv))\\
    & >1
\end{aligned}
\]
which is a contradiction.
As a result
\begin{equation}
    \label{eq:hard_proof_3}
    \lambda(\vv') \leq \lambda(\vv)
\end{equation}

Using equations \eqref{eq:hard_proof_2} and \eqref{eq:hard_proof_3}, and the fact that $\relu \circ \left(\phi^*\right)'$ is an increasing function, we obtain for all $i \in \{2, \ldots, n\}$
\begin{equation}
\begin{aligned}
    \bar{p}_i(\vv') 
 &= \frac{1}{n}\relu\left(\left(\phi^*\right)'\left(\de(v_i' + \lambda(\vv'))\right)\right)\\
 &= \frac{1}{n}\relu\left(\left(\phi^*\right)'\left(\de(v_i + \lambda(\vv'))\right)\right)\\
 & \leq \frac{1}{n}\relu\left(\left(\phi^*\right)'\left(\de(v_i + \lambda(\vv))\right)\right)\\
 & \leq \bar{p}_i(\vv)
\end{aligned}
\end{equation}
In addition
\[
\sum_{i=1}^n \bar{p}_i(\vv') = 1 = \sum_{i=1}^n \textcolor{black}{\bar{p}}_i(\vv)
\]
So necessarily
\begin{equation}
    \textcolor{black}{\bar{p}_{i_0}}(\vv') \geq \textcolor{black}{\bar{p}_{i_0}}(\vv)
\end{equation}
This holds for any $i_0$ and any $\vv$, which concludes the proof. $\blacksquare$

\section{Proof of Equivalence between \eqref{eq:expvar} and \eqref{eq:dro2}: Link between DRO and Percentile Loss}\label{s:proof_dro_and_percentile}

In the DRO optimization problem of equation \eqref{eq:dro2}, the optimal $\vq$ for any $\vtheta$ has the closed-form formula as shown in Appendix~\ref{s:proof_softmax}
\begin{equation*}
    \forall \vtheta,\quad
    \vq^*\left(\vtheta\right) = \softmax\left(
        \left(\de \cL\left(h(\vx_i; \vtheta), \vy_i\right)\right)_{i=1}^n
    \right)
\end{equation*}
By injecting this in equation \eqref{eq:dro2}, we obtain
\begin{align*}
    \min_{\vtheta}\,& \max_{\vq \in \Delta_n}
        \left(
        \sum_{i=1}^n q_i \cL\left(h(\vx_i; \vtheta), \vy_i\right)
        - \frac{1}{\de} D_{KL}\left(\vq\, \biggr\Vert\, \frac{1}{n}\mathbf{1}\right)
        \right)\\
    = \min_{\vtheta}\,&
        \left(
        \sum_{i=1}^n q^*_i(\vtheta) \cL\left(h(\vx_i; \vtheta), \vy_i\right)
        - \frac{1}{\de} 
            \sum_{i=1}^n q^*_i(\vtheta)
                \log\left(\frac{
                    \exp\left(\de \cL\left(h(\vx_i; \vtheta), \vy_i\right)\right)}{
                    \frac{1}{n}\sum_{j=1}^n\exp\left(\de \cL\left(h(\vx_j; \vtheta), \vy_j\right)\right)}\right)
        \right)\\
    = \min_{\vtheta}\,&
        \left(
        \sum_{i=1}^n q^*_i(\vtheta) \cL\left(h(\vx_i; \vtheta), \vy_i\right)
        - \sum_{i=1}^n q^*_i(\vtheta)
                \frac{1}{\de} \log\left(\exp\left(\de \cL\left(h(\vx_i; \vtheta), \vy_i\right)\right)\right)
        \right.\\
    & + \frac{1}{\de}
        \left.
            \left(\sum_{i=1}^n q^*_i(\vtheta)\right) \times
            \log\left(
            \frac{1}{n}\sum_{j=1}^n\exp\left(\de \cL\left(h(\vx_j; \vtheta), \vy_j\right)\right)
            \right)
        \right)\\
    %
\end{align*}
Since the first two terms cancel each other and $\sum_{i=1}^n q^*_i(\vtheta)=1$, we obtain
\begin{align*}
    \min_{\vtheta}\,& \max_{\vq \in \Delta_n}
        \left(
        \sum_{i=1}^n q_i \cL\left(h(\vx_i; \vtheta), \vy_i\right)
        - \frac{1}{\de} D_{KL}\left(\vq\, \biggr\Vert\, \frac{1}{n}\mathbf{1}\right)
        \right)\\
    = \min_{\vtheta}\,& \frac{1}{\de} \log\left(
            \sum_{j=1}^n\exp\left(\de \cL\left(h(\vx_j; \vtheta), \vy_j\right)\right)
            \right)
            - \frac{1}{\de} \log\left(n\right)\\
    = \min_{\vtheta}\,& \frac{1}{\de} \log\left(
            \sum_{j=1}^n\exp\left(\de \cL\left(h(\vx_j; \vtheta), \vy_j\right)\right)
            \right)
\end{align*}
which is equivalent to the optimization problem \eqref{eq:expvar} because the term $\frac{1}{\de} \log\left(n\right)$ above and the term $\frac{1}{\de} \log\left(\alpha n\right)$ in \eqref{eq:expvar} are independent of $\vtheta$ $\blacksquare$
\section{Proof of Theorem~\ref{th:convergence_dro}: convergence of SGD with Hardness Weighted Sampling for Over-parameterized Deep Neural Networks with ReLU}\label{s:convergence_detailed}

In this section, we provide the proof of Theorem~\ref{th:convergence_dro}.
This generalizes the convergence of SGD for empirical risk minimization in~\citep[Theorem 2]{allen-zhu19a} to the convergence of SGD and our proposed hardness weighted sampler for distributionally robust optimization.

We start by describing in details the assumptions made for our convergence result in Section~\ref{s:assumptions}.

In Section~\ref{s:convergence_theorem2}, we restate Theorem~\ref{th:convergence_dro} using the assumptions and notations previously introduced in Section~\ref{s:notations}.

In Section~\ref{s:proof_convergence}, we give the proof of the convergence theorem. 
We focus on providing theoretical tools that could be used to generalize any convergence result for ERM using SGD to DRO using SGD with hardness weighted sampling as described in \Algref{alg:1}.

\subsection{Assumptions}\label{s:assumptions}
Our analysis is based on the results developed in~\citep{allen-zhu19a} which is a simplified version of~\citep{allen2018convergence}.
Improving on those theoretical results would automatically improve our results as well.

In the following we state our assumptions on the neural network $h$, and the per-example loss function $\cL$.
\begin{assumption}[Deep Neural Network]
\label{as:2}
In this section, we use the following notations and assumptions similar to \citep{allen-zhu19a}:
    \begin{itemize}
        \item h is a fully connected neural network with $L+2$ layers, $\relu$ as activation functions, and $m$ nodes in each hidden layer
        \item For all $i \in \{1, \ldots, n\}$, we denote $h_i: \vtheta \mapsto h_i(\rx_i;\vtheta)$ the $d$-dimensional output scores of $h$ applied to example $\rx_i$ of dimension $\gd$.
        \item For all $i \in \{1, \ldots, n\}$, we denote $\cL_i: h \mapsto \cL\left(h, \ry_i\right)$ where $\ry_i$ is the ground truth associated to example $i$.
        \item $\vtheta=\left(\vtheta_l\right)_{l=0}^{L+1}$ is the set of parameters of the neural network h, where $\vtheta_l$ is the set of weights for layer $l$ with $\vtheta_0 \in \sR^{\gd \times m}$, $\vtheta_{L+1} \in \sR^{m \times d}$, and $\vtheta_l \in \sR^{m \times m}$ for any other $l$.
        \item (Data separation) It exists $\delta > 0$ such that for all $i,j \in \{1, \ldots, n\}$, if $i \neq j, \norm{x_i - x_j} \geq \delta$.
        \item We assume $m \geq \Omega(d \times \textup{poly}(n,L,\delta^{-1}))$ for some sufficiently large polynomial $\textup{poly}$, and $\delta \geq O\left(\frac{1}{L}\right)$. We refer the reader to \citep{allen-zhu19a} for details about the polynomial $\textup{poly}$.
        \item The parameters $\vtheta=\left(\vtheta_l\right)_{l=0}^{L+1}$ are initialized at random such that:
        \begin{itemize}
            \item $\left[\vtheta_0\right]_{i,j} \sim \mathcal{N}\left(0, \frac{2}{m}\right)$ for every $(i,j) \in \{1, \ldots, m\}\times\{1, \ldots, \gd\}$
            \item $\left[\vtheta_l\right]_{i,j} \sim \mathcal{N}\left(0, \frac{2}{m}\right)$ for every $(i,j) \in \{1, \ldots, m\}^2$ and $l \in \{1, \ldots, L\}$
            \item $\left[\vtheta_{L+1}\right]_{i,j} \sim \mathcal{N}\left(0, \frac{1}{d}\right)$ for every $(i,j) \in \{1, \ldots, d\}\times\{1, \ldots, m\}$
        \end{itemize}
    \end{itemize}
\end{assumption}

\begin{assumption}[Regularity of $\cL$]
    \label{as:3}
    \textcolor{black}{There exists $C(\nabla \cL)>0$ and $C(\cL)>0$ such that} for all i, $\cL_i$ is a $C(\nabla \cL)$-gradient Lipschitz continuous, $C(\cL)$-Lipschitz continuous, and bounded (potentially non-convex) function.
    \textcolor{black}{
    When the optimization is performed on a closed convex set, the existence of $C(\nabla \cL)$ implies that there exists a constant $A(\nabla \cL)>0$ that bounds the gradients of $\cL_i$ for all i.}
\end{assumption}

\subsection{Convergence theorem (restated)}\label{s:convergence_theorem2}

In this section, we restate the convergence Theorem~\ref{th:convergence_dro} for SGD with hardness weighted sampling and stale per-example loss vector.

As an intermediate step, we will first 
generalize the convergence of SGD in \citep[Theorem 2]{allen-zhu19a} to the minimization of the distributionally robust loss using SGD and an \emph{exact} hardness weighted sampling~\eqref{eq:p}, i.e. with an exact per-example loss vector.
\begin{theorem}[Convergence with exact per-example loss vector]
    \label{th:conv_sgd_exact_loss_history}
    Let batch size $1 \leq b \leq n$, and $\epsilon > 0$.
    \textcolor{black}{Under assumption \ref{as:2} and assumption \ref{as:3},}
    suppose there exists constants $C_1,\, C_2,\, C_3 > 0$ such that
    the number of hidden units satisfies 
    $m \geq C_1 (d \epsilon^{-1} \times \textup{poly}(n,L,\delta^{-1}))$,
    $\delta \geq \left(\frac{C_2}{L}\right)$,
    and the learning rate be 
    $\eta_{exact} = C_3\left(
        \min \left(1,\,
        \frac{\alpha n^2 \rho}{\de C(\cL)^2+ 2 n \rho C(\nabla \cL)}
        \right)
        \times \frac{b\delta d}{\textup{poly}(n,L)m\log^2(m)}
        \right)$.
    There exists constants $C_4,\,C_5 >0$ such that
    with probability at least $1 - \exp\left(-C_4(\log^2(m))\right)$ over the randomness of the initialization and the mini-batches, SGD with hardness weighted sampling and exact per-example loss vector guarantees
    $\norm{\nabla_{\vtheta} (R\circ \vL \circ h)(\vtheta)} \leq \epsilon$ after 
    $T=C_5\left(\frac{L n^3}{\eta_{exact} \delta \epsilon^2}\right)$ iterations.
\end{theorem}
The proof can be found in Appendix~\ref{s:proof_exact_loss_history}.

$\alpha = \min_{\vtheta} \min_{i} \bar{p}_i(\vL(\vtheta))$ is a lower bound on the sampling probabilities.
For the Kullback-Leibler $\phi$-divergence, and for any $\phi$-divergence satisfying Definition~\ref{def:phi_divergence} with a robustness parameter $\de$ small enough, we have $\alpha > 0$.
We refer the reader to \citep[Theorem 2]{allen-zhu19a} for the values of the constants $C_1,\, C_2,\, C_3,\, C_4,\, C_5$ and the definitions of the polynomials.

Compared to \citep[Theorem 2]{allen-zhu19a} only the learning rate differs.
The $\min(1, \,.\,)$ operation in the formula for $\eta_{exact}$ allows us to guarantee that $\eta_{exact} \leq \eta '$ where $\eta '$ is the learning rate of \citep[Theorem 2]{allen-zhu19a}.

It is worth noting that for the KL $\phi$-divergence, $\rho=\frac{1}{n}$.
In addition, in the limit $\de \rightarrow 0$, which corresponds to ERM, we have $\alpha \rightarrow \frac{1}{n}$.
As a result, we recover exactly Theorem 2 of \citep{allen-zhu19a} as extended in their Appendix A for any smooth loss function $\cL$ that satisfies assumption $\ref{as:3}$ with $C(\nabla\cL)=1$.

%

%
We now restate the convergence of SGD with hardness weighted sampling and a stale per-example loss vector as in \Algref{alg:1}.
\begin{theorem}[Convergence with a stale per-example loss vector]
\label{th:conv_sgd_stale_loss_history}
Let batch size $1 \leq b \leq n$, and $\epsilon > 0$.
Under the conditions of Theorem \ref{th:conv_sgd_exact_loss_history}, 
the same notations,
and with the learning rate
$\eta_{stale} = C_6
\min \left(1,\,
\frac{\alpha \rho d^{3/2} \delta b \log\left(\frac{1}{1-\alpha}\right)}{\de C(\cL) A(\nabla \cL)L m^{3/2} n^{3/2} \log^2(m)}
\right)
\times \eta_{exact}$ for a constant $C_6 > 0$.
With probability at least $1 - \exp\left(-C_4(\log^2(m))\right)$ over the randomness of the initialization and the mini-batches, SGD with hardness weighted sampling and stale per-example loss vector guarantees
$\norm{\nabla_{\vtheta} (R\circ \vL \circ h)(\vtheta)} \leq \epsilon$ after 
$T=C_5\left(\frac{L n^3}{\eta_{stale} \delta \epsilon^2}\right)$ iterations.
\end{theorem}
The proof can be found in Appendix~\ref{s:proof_stale_loss_history}.

$C(\cL) > 0$ is a constant such that $\cL$ is $C(\cL)$-Lipschitz continuous,
and $A(\nabla \cL) > 0$ is a constant that bounds the gradient of $\cL$ with respect to its input.
$C(\cL)$ and $A(\nabla \cL)$ are guaranteed to exist under assumptions \ref{as:2}.

Compared to Theorem \ref{th:conv_sgd_exact_loss_history} only the learning rate differs.
Similarly to Theorem \ref{th:conv_sgd_exact_loss_history}, when $\de$ tends to zero we recover Theorem 2 of \citep{allen-zhu19a}.

It is worth noting that when $\de$ increases, $\frac{\alpha \rho d^{3/2} \delta b \log\left(\frac{1}{1-\alpha}\right)}{\de C(\cL) A(\nabla \cL) L m^{3/2} n^{3/2} \log^2(m)}$ decreases.
This implies that $\eta_{stale}$ decreases faster than $\eta_{exact}$ when $\de$ increases.
This was to be expected since the error that is made by using the stale per-example loss vector instead of the exact loss increases when $\de$ increases.

\subsection{Proofs of convergence}\label{s:proof_convergence}

In this section, we prove the results of Therem \ref{th:conv_sgd_exact_loss_history} and \ref{th:conv_sgd_stale_loss_history}.

For the ease of reading the proof, we remind here the chain rules for the distributionally robust loss that we are going to use intensively in the following proofs.

\paragraph{\textbf{Chain rule for the derivative of $R \circ \vL$ with respect to the network outputs $h$:}}\ \\
\begin{equation}
    \label{eq:reminder_chain_rule_h}
    \begin{aligned}
        \nabla_h (R \circ \vL)(h(\vtheta)) 
        &= \left(\nabla_{h_i} (R \circ \vL)(h(\vtheta))\right)_{i=1}^n\\
        \forall i \in \{1,\ldots n\},\quad \nabla_{h_i} (R \circ \vL)(h(\vtheta))
        &=\sum_{j=1}^n \frac{\partial R}{\partial v_j}(\vL(h(\vtheta))) \nabla_{h_i}\textcolor{black}{\mathcal{L}_j}(h_j(\vtheta)) \\
        &= \bar{p}_i(\vL(h(\vtheta)))\nabla_{h_i}\textcolor{black}{\mathcal{L}_i}(h_i(\vtheta))
    \end{aligned}
\end{equation}

\paragraph{\textbf{Chain rule for the derivative of $R \circ \vL \circ h$ with respect to the network parameters $\vtheta$:}}\ \\
\begin{equation}
\label{eq:reminder_chain_rule_theta}
    \begin{aligned}
        \nabla_{\vtheta} (R \circ \vL \circ h)(\vtheta) 
        &= \sum_{i=1}^n \nabla_\theta h_i(\vtheta) \nabla_{h_i}(R \circ \vL)(h(\vtheta))\\
        &= \sum_{i=1}^n \bar{p}_i(\vL(h(\vtheta))) \nabla_\theta h_i(\vtheta) \nabla_{h_i}\textcolor{black}{\mathcal{L}_i}(h_i(\vtheta))\\
        &= \sum_{i=1}^n \bar{p}_i(\vL(h(\vtheta))
        \nabla_{\vtheta}(\textcolor{black}{\mathcal{L}_i} \circ h_i)(\vtheta))\\
    \end{aligned}
\end{equation}
where for all $i \in \{1,\ldots n\}$,  $\nabla_\theta h_i(\vtheta)$ is the transpose of the Jacobian matrix of $h_i$ as a function of $\vtheta$.

\subsubsection{Proof that R o L is one-sided gradient Lipchitz}
This property that $R\circ \vL$ is one-sided gradient Lipschitz is a key element for the proof of the semi-smoothness theorem for the distributionally robust loss Theorem \ref{th:semi-smoothness}.

Under Definition~\ref{def:phi_divergence} for the $\phi$-divergence, we have shown that $R$ is $\frac{\de}{n\rho}$-gradient Lipchitz continuous (Lemma~\ref{lemma:R_property}).
And under assumption \ref{as:3}, for all $i$, $\cL_i$ is $C(\cL)$-Lipschitz continuous and $C(\nabla \cL)$-gradient Lipschitz continuous.

Let $\vz=(\vz_i)_{i=1}^n, \vz'=(\vz_i')_{i=1}^n \in \sR^{dn}$.

We want to show that $R\circ \vL$ is one-sided gradient Lipschitz, i.e. we want to prove the existence of a constant $C>0$, independent to $\vz$ and $\vz'$, such that:
\[
\langle \nabla_{\vz} (R\circ \vL)(\vz) - \nabla_z (R\circ \vL)(\vz'), \vz - \vz'\rangle 
\leq C \norm{\vz - \vz'}^2
\]
We have
\begin{equation}
	\label{eq:one-sided1}
	\begin{aligned}
	    \langle \nabla_{\vz} (R\circ \vL)(\vz) - \nabla_{\vz} (&R\circ \vL)(\vz'), \vz - \vz'\rangle \\
	    & = \sum_{i=1}^n \langle \nabla_{\vz_i} (R\circ \vL)(\vz) - \nabla_{\vz_i} (R\circ \vL)(\vz'), \vz_i - \vz_i'\rangle\\
	    & = \sum_{i=1}^n \langle \bar{p}_i(\vL(\vz))\nabla_{\vz_i} \textcolor{black}{\mathcal{L}_i}(\vz_i) - \bar{p}_i(\vL(\vz'))\nabla_{\vz_i}\textcolor{black}{\mathcal{L}_i}(\vz_i'), \vz_i - \vz_i'\rangle\\
	    & = \sum_{i=1}^n \bar{p}_i(\vL(\vz))\langle \nabla_{\vz_i} \textcolor{black}{\mathcal{L}_i}(\vz_i) - \nabla_{\vz_i}\textcolor{black}{\mathcal{L}_i}(\vz_i'), \vz_i - \vz_i'\rangle\\
	    & \quad + \sum_{i=1}^n \left(\bar{p}_i(\vL(\vz)) - \bar{p}_i(\vL(\vz'))\right)\langle \nabla_{\vz_i}\textcolor{black}{\mathcal{L}_i}(\vz_i'), \vz_i - \vz_i'\rangle\\
	\end{aligned}
\end{equation}

Where for all $i \in \{1,\ldots,n\}$ we have used the chain rule 
\[
\nabla_{\vz_i} (R\circ \vL)(\vz) = \sum_{j=1}^n \frac{\partial R}{\partial \vv_j}(\cL(\vz)) \nabla_{\vz_i}\textcolor{black}{\mathcal{L}_j}(\vz_j) = \bar{p}_i(\vL(\vz))\nabla_{\vz_i} \textcolor{black}{\mathcal{L}_i}(\vz_i)
\]

Let
\[
A = \left|\sum_{i=1}^n \bar{p}_i(\vL(\vz))\langle \nabla_{\vz_i} \textcolor{black}{\mathcal{L}_i}(\vz_i) - \nabla_{\vz_i}\textcolor{black}{\mathcal{L}_i}(\vz_i'), \vz_i - \vz_i'\rangle\right|
\]

For all $i$, $\cL_i$ is $C(\nabla \cL)$-gradient Lipchitz continuous, so using Cauchy-Schwarz inequality
\begin{equation}
	\label{eq:one-sided2}
	    A
	    \leq \sum_{i=1}^n C(\nabla \cL) \norm{\vz_i - \vz_i'}^2
	    = C(\nabla \cL) \norm{\vz - \vz'}^2
\end{equation}

Let 
\[
B = \left|\sum_{i=1}^n \left(\bar{p}_i(\vL(\vz)) - \bar{p}_i(\vL(\vz'))\right)\langle \nabla_{\vz_i}\textcolor{black}{\mathcal{L}_i}(\vz_i'), \vz_i - \vz_i'\rangle\right|
\]
Using the triangular inequality:
\begin{equation}
	\label{eq:one-sided3}
	\begin{aligned}
		B
		& \leq \left|\sum_{i=1}^n \left(\bar{p}_i(\vL(\vz)) - \bar{p}_i(\vL(\vz'))\right)(\textcolor{black}{\mathcal{L}_i}(\vz_i) - \textcolor{black}{\mathcal{L}_i}(\vz_i')\right|\\
		& \quad + \left|\sum_{i=1}^n \left(\bar{p}_i(\vL(\vz)) - \bar{p}_i(\vL(\vz'))\right)(\textcolor{black}{\mathcal{L}_i}(\vz_i') + \langle \nabla_{\vz_i}\textcolor{black}{\mathcal{L}_i}(\vz_i'), \vz_i - \vz_i'\rangle - \textcolor{black}{\mathcal{L}_i}(\vz_i)\right|\\
		& \leq \left\langle \nabla_{\textcolor{black}{\vv}} R(\vL(\vz)) - \nabla_{\textcolor{black}{\vv}} R(\vL(\vz')), \cL(\vz) - \cL(\vz')\right\rangle\\
		& \quad + 2 \sum_{i=1}^n\left|\textcolor{black}{\mathcal{L}_i}(\vz_i') + \langle \nabla_{\vz_i}\textcolor{black}{\mathcal{L}_i}(\vz_i'), \vz_i - \vz_i'\rangle - \cL_i(\vz_i)\right|\\
		& \leq \frac{\de}{n \rho}\norm{\vL(\vz) - \vL(\vz')}^2 
		+ 2\frac{C(\nabla \cL)}{2}\norm{\vz - \vz'}^2\\
		& \leq \left(\frac{\de C(\cL)^2}{n \rho}
		+ C(\nabla \cL)\right)\norm{\vz - \vz'}^2
	\end{aligned}
\end{equation}

Combining equations \eqref{eq:one-sided1}, \eqref{eq:one-sided2} and \eqref{eq:one-sided3} we finally obtain
\begin{equation}
	\langle \nabla_{\vz} (R\circ \vL)(\vz) - \nabla_{\vz} (R\circ \vL)(\vz'), \vz - \vz'\rangle
	\leq \left(\frac{\de C(\cL)^2}{n \rho}
	+ 2C(\nabla \cL)\right)\norm{\vz - \vz'}^2
\end{equation}

From there, we can obtain the following inequality that will be used for the proof of the semi-smoothness property in Theorem \ref{th:semi-smoothness}
\begin{equation}
    \label{eq:lipchitz}
	\begin{aligned}
	    &R(\vL(\vz')) - R(\vL(\vz)) - \langle \nabla_{\vz} (R\circ \vL)(\vz), \vz' - \vz \rangle\\
	    &\quad = \int_{t=0}^{1} \langle\nabla_{\vz} (R\circ \vL)\left(\vz + t(\vz' - \vz)\right) - \nabla_{\vz} (R\circ \vL)(\vz), \vz' - \vz\rangle dt\\
	    &\quad \leq \frac{1}{2}\left(\frac{\de C(\cL)^2}{n \rho}
	    + 2C(\nabla \cL)\right)\norm{\vz - \vz'}^2
	\end{aligned}
\end{equation}

\subsubsection{Semi-smoothness property of the distributionally robust loss}

We prove the following lemma which is a generalization of Theorem 4 in \citep{allen-zhu19a} for the distributionally robust loss.

\begin{theorem}[Semi-smoothness of the distributionally robust loss]\ \\
    \label{th:semi-smoothness}
    Let $\omega \in \left[\Omega\left(\frac{d^{3/2}}{m^{3/2}L^{3/2}\log^{3/2}(m)}\right),
    O\left(\frac{1}{L^{4.5}\log^{3}(m)}\right)\right]$,
    and the $\vtheta^{(0)}$ being initialized randomly as described in assumption \ref{as:2}.
    With probability as least $1 - \exp{(-\Omega(m\omega^{3/2}L))}$ over the initialization, we have for all $\vtheta, \vtheta' \in \left(\sR^{m \times m}\right)^L$ with $\norm{\vtheta - \vtheta^{(0)}}_2 \leq \omega$,
    and $\norm{\vtheta - \vtheta'}_2 \leq \omega$
    \begin{equation}
        \begin{aligned}
            R(\vL(h(\vtheta ')) & 
            \leq
            R(\vL(h(\vtheta)) + 
            \langle 
            \nabla_{\vtheta} (R \circ \vL \circ h)(\vtheta), \vtheta' - \vtheta
            \rangle \\
            & + \norm{\nabla_h (R \circ \vL)(h(\vtheta))}_{2,1} O\left(\frac{L^2\omega^{1/3}\sqrt{m\log(m)}}{\sqrt{d}}\right)\norm{\vtheta' - \vtheta}_{2, \infty}\\
            & + O\left(\left(\frac{\de C(\cL)^2}{n \rho}
	        + 2C(\nabla \cL)\right)\frac{n L^2 m}{d}\right) \norm{\vtheta' - \vtheta}_{2, \infty}^2\\
	    %
        \end{aligned}
    \end{equation}
\end{theorem}

where for all layer $l \in \{1,\ldots, L\}$, $\vtheta_l$ is the vector of parameters for layer $l$, and
\[
\begin{aligned}
    \norm{\vtheta' - \vtheta}_{2, \infty} 
    &= \max_{l} \norm{\vtheta_l' - \vtheta_l}_2\\
    \norm{\vtheta' - \vtheta}_{2, \infty}^2 
    &= \left(\max_{l} \norm{\vtheta_l' - \vtheta_l}_2^2\right)^2= \max_{l} \norm{\vtheta_l' - \vtheta_l}_2^2\\
    \norm{\nabla_h (R \circ \vL)(h(\vtheta))}_{2,1} 
    &=
        \sum_{i=1}^n \norm{\nabla_{h_i} (R \circ \vL)(h(\vtheta))}_2\\
    &= \sum_{i=1}^n \norm{\bar{p}_i(\vL(h(\vtheta)))\nabla_{h_i}\textcolor{black}{\mathcal{L}_i}(h_i(\vtheta))}_2 \quad \text{ (chain rule \eqref{eq:reminder_chain_rule_h}) }\\
    %
\end{aligned}
\]

To compare this semi-smoothness result to the one in \citep[Theorem 4]{allen-zhu19a}, let us first remark that
\[
\begin{aligned}
    \norm{\nabla_h (R \circ \vL)(h(\vtheta))}_{2,1}
    & \leq \sqrt{n} \norm{\nabla_h (R \circ \vL)(h(\vtheta))}_{2,2}\\
\end{aligned}
\]

As a result, our result is analogous to \citep[Theorem 4]{allen-zhu19a}, up to an additional multiplicative factor $\left(\frac{\de C(\cL)^2}{n \rho}+ 2C(\nabla \cL)\right)$ in the last term of the right-hand side.
It is worth noting that there is also implicitly an additional multiplicative factor $C(\nabla \cL)$ in Theorem 3 of \citep{allen-zhu19a} since \citep{allen-zhu19a} make the assumption that $C(\nabla \cL)=1$ \citep[see][Appendix A]{allen-zhu19a}.

Let $\vtheta, \vtheta' \in \left(\sR^{m \times m}\right)^L$ verifying the conditions of Theorem \ref{th:semi-smoothness}.

Let
$
A = R(\vL(h(\vtheta ')) - R(\vL(h(\vtheta)) 
    - \langle \nabla_{\vtheta} (R \circ \vL \circ h)(\vtheta), \vtheta' - \vtheta\rangle
$
, the quantity we want to bound.

Using \eqref{eq:lipchitz} for $\vz=h(\vtheta)$ and $\vz'=h(\vtheta')$, we obtain
\begin{equation}
\begin{aligned}
    A
    & \leq \frac{1}{2}\left(\frac{\de C(\cL)^2}{n \rho}
	    + 2C(\nabla \cL)\right)\norm{h(\vtheta') - h(\vtheta)}_2^2\\
	& \quad + \langle \nabla_h (R\circ \vL)(h(\vtheta)), h(\vtheta') - h(\vtheta) \rangle\\
	& \quad - \langle \nabla_{\vtheta} (R \circ \vL \circ h)(\vtheta), \vtheta' - \vtheta\rangle\\
\end{aligned}
\end{equation}

Then using the chain rule \eqref{eq:reminder_chain_rule_theta}
\begin{equation}
\label{eq:11_3}
\begin{aligned}
    A
    & \leq \frac{1}{2}\left(\frac{\de C(\cL)^2}{n \rho}
	    + 2C(\nabla \cL)\right)\norm{h(\vtheta') - h(\vtheta)}_2^2\\
	& \quad + \sum_{i=1}^n\langle \nabla_{h_i} (R\circ \vL)(h(\vtheta)), 
	h_i(\vtheta') - h_i(\vtheta) - \left(\nabla_{\vtheta}h_i(\vtheta)\right)^T(\vtheta'-\vtheta)\rangle\\
\end{aligned}
\end{equation}

For all $i \in \{1, \ldots, n\}$, let us denote
$\breve{loss}_i := \nabla_{h_i} (R\circ \vL)(h(\vtheta))$
to match the notations used in \citep{allen-zhu19a} for the derivative of the loss with respect to the output of the network for example i of the training set.

With this notation, we obtain exactly equation (11.3) in \citep{allen-zhu19a} up to the multiplicative factor $\left(\frac{\de C(\cL)^2}{n \rho} + 2C(\nabla \cL)\right)$ for the distributionally robust loss.

From there the proof of Theorem 4 in \citep{allen-zhu19a} being independent to the formula for $\breve{loss}_i$, we can conclude the proof of our Theorem \ref{th:semi-smoothness} as in \citep[][Appendix A]{allen-zhu19a}.

\subsubsection{Gradient bounds for the distributionally robust loss}

We prove the following lemma which is a generalization of Theorem 3 in \citep{allen-zhu19a} for the distributionally robust loss.

\begin{theorem}[Gradient Bounds for the Distributionally Robust Loss]\ \\
\label{th:gradient_bound}
    Let $\omega \in O\left(\frac{\delta^{3/2}}{n^{9/2}L^6\log^3(m)}\right)$,  and $\vtheta^{(0)}$ being initialized randomly as described in assumption~\ref{as:2}.
    With probability as least $1 - \exp{(-\Omega(m\omega^{3/2}L))}$ over the initialization, we have for all $\vtheta \in \left(\sR^{m \times m}\right)^L$ with $\norm{\vtheta - \vtheta^{(0)}}_2 \leq \omega$
    \begin{equation}
        \begin{aligned}
            &\forall i \in \{1,\ldots,n\},\,\, \forall l \in \{1, \ldots, L\},\,\, \forall \hat{\vL} \in \sR^n\\
            &\norm{\bar{p}_i(\hat{\vL}) \nabla_{\vtheta_l}(\textcolor{black}{\mathcal{L}_i}\circ h_i)(\vtheta)}_2^2 
            \leq O\left(\frac{m}{d}\norm{\bar{p}_i(\hat{\vL}) \nabla_{h_i}\textcolor{black}{\mathcal{L}_i} (h_i(\vtheta))}_2^2\right)\\
            & \forall l \in \{1, \ldots, L\},\,\, \forall \hat{\vL} \in \sR^n\\
            &\norm{\sum_{i=1}^n\bar{p}_i(\hat{\vL}) \nabla_{\vtheta_l}(\textcolor{black}{\mathcal{L}_i} \circ h_i)(\vtheta)}_2^2 
            \leq O\left(\frac{m n}{d}\sum_{i=1}^n\norm{\bar{p}_i(\hat{\vL}) \nabla_{h_i}\textcolor{black}{\mathcal{L}_i} (h_i(\vtheta))}_2^2\right)\\
            & \norm{\sum_{i=1}^n\bar{p}_i(\hat{\vL}) \nabla_{\vtheta_L}(\textcolor{black}{\mathcal{L}_i} \circ h_i)(\vtheta)}_2^2
            \geq \Omega\left(\frac{m \delta}{d n^2}\sum_{i=1}^n\norm{\bar{p}_i(\hat{\vL}) \nabla_{h_i}\textcolor{black}{\mathcal{L}_i} (h_i(\vtheta))}_2^2\right)\\
        \end{aligned}
    \end{equation}
\end{theorem}

It is worth noting that the loss vector $\hat{\vL}$ used for computing the robust probabilities $\bar{\vp}(\hat{\vL})=\left(\bar{p}_i(\hat{\vL})\right)_{i=1}^n$ does not have to be equal to $\vL(h(\vtheta))$.

We will use this for the proof of the Robust SGD with stale per-example loss vector.

The adaptation of the proof of Theorem 3 in \citep{allen-zhu19a} is straightforward.

Let $\vtheta \in \left(\sR^{m \times m}\right)^L$ satisfying the conditions of Theorem \ref{th:gradient_bound}, and $\hat{\vL} \in \sR^n$.

Let us denote $\vv := \left(\bar{p}_i(\hat{\vL}) \nabla_{h_i}\cL_i(h_i(\vtheta))\right)_{i=1}^n$, applying the proof of Theorem 3 in \citep{allen-zhu19a} to our $\vv$ gives:
\[
\begin{aligned}
    &\forall i \in \{1,\ldots,n\},\,\, \forall l \in \{1, \ldots, L\},\\
            &\norm{\bar{p}_i(\hat{\vL}) \nabla_{\vtheta_l}(\textcolor{black}{\mathcal{L}_i} \circ h_i)(\vtheta)}_2^2 
            \leq O\left(\frac{m}{d}\norm{\bar{p}_i(\hat{\vL}) \nabla_{h_i}\textcolor{black}{\mathcal{L}_i} (h_i(\vtheta))}_2^2\right)\\
            & \forall l \in \{1, \ldots, L\},\,\, \forall \hat{\vL} \in \sR^n\\
            &\norm{\sum_{i=1}^n\bar{p}_i(\hat{\vL}) \nabla_{\vtheta_l}(\textcolor{black}{\mathcal{L}_i} \circ h_i)(\vtheta)}_2^2 
            \leq O\left(\frac{m n}{d}\sum_{i=1}^n\norm{\bar{p}_i(\hat{\vL}) \nabla_{h_i}\textcolor{black}{\mathcal{L}_i} (h_i(\vtheta))}_2^2\right)\\
            & \norm{\sum_{i=1}^n\bar{p}_i(\hat{\vL}) \nabla_{\vtheta_L}(\textcolor{black}{\mathcal{L}_i} \circ h_i)(\vtheta)}_2^2
            \geq \Omega\left(\frac{m \delta}{d n}\max_{i}\left(\norm{\bar{p}_i(\hat{\vL}) \nabla_{h_i}\textcolor{black}{\mathcal{L}_i} (h_i(\vtheta))}_2^2\right)\right)\\
\end{aligned}
\]

In addition
\[
\max_{i}\left(\norm{\bar{p}_i(\hat{\vL}) \nabla_{h_i}\textcolor{black}{\mathcal{L}_i} (h_i(\vtheta))}_2^2\right)
\geq \frac{1}{n} \sum_{i=1}^n \norm{\bar{p}_i(\hat{\vL}) \nabla_{h_i}\textcolor{black}{\mathcal{L}_i} (h_i(\vtheta))}_2^2
\]
This allows us to conclude the proof of our Theorem \ref{th:gradient_bound}. $\blacksquare$

\subsubsection{Convergence of SGD with Hardness Weighted Sampling and exact per-example loss vector}\label{s:proof_exact_loss_history}

We can now prove Theorem \ref{th:conv_sgd_exact_loss_history}.

Similarly to the proof of the convergence of SGD for the mean loss (Theorem 2 in \citep{allen-zhu19a}), the convergence of SGD for the distributionally robust loss will mainly rely on the semi-smoothness property (Theorem \ref{th:semi-smoothness}) and the gradient bound (Theorem \ref{th:gradient_bound}) that we have proved previously for the distributionally robust loss.

Let $\vtheta \in \left(\sR^{m \times m}\right)^L$ satisfying the conditions of Theorem \ref{th:conv_sgd_exact_loss_history},
and $\hat{\vL}$ be the exact per-example loss vector at $\vtheta$, i.e.
\begin{equation}
    \label{eq:exact_loss_history}
    \hat{\vL}=\left(\textcolor{black}{\mathcal{L}_i}(h_i(\vtheta))\right)_{i=1}^n
\end{equation}
For the batch size $b \in \{1, \ldots, n\}$, let $S=\{i_j\}_{j=1}^b$ a batch of indices drawn from $\bar{\vp}(\hat{\vL})$ without replacement, i.e.
\begin{equation}
    \forall j \in \{1, \ldots b\}, \,\, i_j \overset{\text{i.i.d.}}{\sim} \bar{\vp}(\hat{\vL})
\end{equation}

Let $\vtheta' \in \left(\sR^{m \times m}\right)^L$ be the values of the parameters after a stochastic gradient descent step at $\vtheta$ for the batch $S$, i.e.
\begin{equation}
    \label{eq:next_theta}
    \vtheta' = \vtheta - \eta \frac{1}{b} \sum_{i \in S} \nabla_{\vtheta}(\textcolor{black}{\mathcal{L}_i} \circ h_i)(\vtheta) 
\end{equation}
where $\eta >0$ is the learning rate.

Assuming that $\vtheta$ and $\vtheta'$ satisfies the conditions of Theorem \ref{th:semi-smoothness}, we obtain
\begin{equation}
\label{eq:cv_sgd_exact_1}
    \begin{aligned}
            R(\vL(h(\vtheta ')) \leq &
            R(\vL(h(\vtheta)) 
            - \eta \langle \nabla_{\vtheta} (R \circ \vL \circ h)(\vtheta),\frac{1}{b} \sum_{i \in S} \nabla_{\vtheta}(\textcolor{black}{\mathcal{L}_i} \circ h_i)(\vtheta)\rangle \\
            & + \eta \sqrt{n}\norm{\nabla_h (R \circ \vL)(h(\vtheta))}_{2,2} O\left(\frac{L^2\omega^{1/3}\sqrt{m\log(m)}}{\sqrt{d}}\right)
            \norm{
            \frac{1}{b} \sum_{i \in S} \nabla_{\vtheta}(\textcolor{black}{\mathcal{L}_i} \circ h_i)(\vtheta)
            }_{2, \infty}\\
            & + \eta^2 O\left(\left(\frac{\de C(\cL)^2}{n \rho}
	    + 2C(\nabla \cL)\right)\frac{n L^2 m}{d}\right) 
	    \norm{
	    \frac{1}{b} \sum_{i \in S} \nabla_{\vtheta}(\textcolor{black}{\mathcal{L}_i} \circ h_i)(\vtheta)
	    }_{2, \infty}^2
        \end{aligned}
\end{equation}
where we refer to \eqref{eq:reminder_chain_rule_theta} for the form of $\nabla_{\vtheta} (R \circ \vL \circ h)(\vtheta)$ and to \eqref{eq:reminder_chain_rule_h} for the form of $\nabla_h (R \circ \vL)(h(\vtheta))$.

In addition, we make the assumption that for the set of values of $\vtheta$ considered the hardness weighted sampling probabilities admit an upper-bound
\begin{equation}
    \label{eq:alpha}
    \alpha = \min_{\vtheta} \min_{i} \bar{p}_i(\vL(\vtheta)) > 0
\end{equation}
Which is always satisfied under assumption \ref{as:3} for Kullback-Leibler $\phi$-divergence, and for any $\phi$-divergence satisfying Definition~\ref{def:phi_divergence} with a robustness parameter $\de$ small enough.

Let $\E_S$ be the expectation with respect to $S$.
Applying $\E_S$ to \eqref{eq:cv_sgd_exact_1}, we obtain
\begin{equation}
\label{eq:cv_sgd_exact_3}
    \begin{aligned}
        \E_S&\left[R(\vL(h(\vtheta '))\right] \\
        \leq &
        R(\vL(h(\vtheta)) 
        - \eta \norm{\nabla_{\vtheta} (R \circ \vL \circ h)(\vtheta)}_{2,2}^2 \\
        & + \eta
        \norm{
        \nabla_h (R \circ \vL)(h(\vtheta))
        }_{2,2}
        O\left(\frac{n L^2\omega^{1/3}\sqrt{m\log(m)}}{\sqrt{d}}\right)
        \sqrt{
        \sum_{i=1}^n \max_{l} \norm{\bar{p}_i(\hat{\vL})\nabla_{\vtheta_l}(\textcolor{black}{\mathcal{L}_i} \circ h_i)(\vtheta)}^2
        }\\
        & + \eta^2 O\left(\left(\frac{\de C(\cL)^2}{n \rho}
	    + 2C(\nabla \cL)\right)\frac{n L^2 m}{d}\right) 
	    \frac{1}{\alpha}\sum_{i=1}^n \max_{l} \norm{\bar{p}_i(\hat{\vL})\nabla_{\vtheta_l}(\textcolor{black}{\mathcal{L}_i} \circ h_i)(\vtheta)}^2
    \end{aligned}
\end{equation}
where we have used the following results:
\begin{itemize}
    \item For any integer $k \geq 1$, and all $\left(\va_i\right)_{i=1}^n \in \left(\sR^k\right)^n$, we have (see the proof in \ref{p:tech_lemma_1})
    \begin{equation}
        \label{eq:expectation}
        \begin{aligned}
            \E_S\left[\frac{1}{b}\sum_{i \in S} \va_i\right]
            &= \E_{\bar{p}(\hat{\vL})}\left[\va_{i}\right]
        \end{aligned}
    \end{equation}
    \item Using \eqref{eq:expectation} for $\left(\va_i\right)_{i=1}^n=\left(\nabla_{\vtheta}(\cL_i \circ h_i)(\vtheta)\right)_{i=1}^n$, and the chain rule \eqref{eq:reminder_chain_rule_theta}
    \begin{equation}
        \E_S\left[\frac{1}{b}\sum_{i \in S} \nabla_{\vtheta}(\textcolor{black}{\mathcal{L}_i} \circ h_i)(\vtheta)\right]
            = \sum_{i=1}^n\bar{p}_i(\hat{\vL})\nabla_{\vtheta}(\textcolor{black}{\mathcal{L}_i} \circ h_i)(\vtheta)
            = \nabla_{\vtheta} (R \circ \vL \circ h)(\vtheta)
    \end{equation}
    \item Using the triangular inequality
    \begin{equation}
    \label{eq:tr_ineq}
    \begin{aligned}
        \norm{
        \frac{1}{b} \sum_{i \in S} \nabla_{\vtheta}(\textcolor{black}{\mathcal{L}_i} \circ h_i)(\vtheta)
        }_{2, \infty}
        & \leq \frac{1}{b} \sum_{i \in S} \norm{\nabla_{\vtheta}(\textcolor{black}{\mathcal{L}_i} \circ h_i)(\vtheta)}_{2, \infty}
    \end{aligned}
    \end{equation}
    And using \eqref{eq:expectation} for $\left(a_i\right)_{i=1}^n=\left(\norm{\nabla_{\vtheta}(\cL_i \circ h_i)(\vtheta)}_{2,\infty}\right)_{i=1}^n$,
    \begin{equation}
    \begin{aligned}
        \E_S\left[
        \norm{
        \frac{1}{b} \sum_{i \in S} \nabla_{\vtheta}(\textcolor{black}{\mathcal{L}_i} \circ h_i)(\vtheta)
        }_{2, \infty}
        \right]
        & \leq \sum_{i=1}^n\bar{p}_i(\hat{\vL}) \norm{\nabla_{\vtheta}(\textcolor{black}{\mathcal{L}_i} \circ h_i)(\vtheta)}_{2, \infty}\\
        & \leq \sum_{i=1}^n 
        \max_{l}\norm{\nabla_{\vtheta_l}(\bar{p}_i(\hat{\vL})\textcolor{black}{\mathcal{L}_i} \circ h_i)(\vtheta)}_{2}\\
        & \leq \sqrt{n}\sqrt{
        \sum_{i=1}^n\max_{l}\norm{\nabla_{\vtheta_l}(\bar{p}_i(\hat{\vL})\textcolor{black}{\mathcal{L}_i} \circ h_i)(\vtheta)}_{2}^2
        }
    \end{aligned}
    \end{equation}
    where we have used Cauchy-Schwarz inequality for the last inequality.
    \item Using \eqref{eq:tr_ineq} and the convexity of the function $x \mapsto x^2$
    \begin{equation}
    \begin{aligned}
        \norm{
        \frac{1}{b} \sum_{i \in S} \nabla_{\vtheta}(\textcolor{black}{\mathcal{L}_i} \circ h_i)(\vtheta)
        }_{2, \infty}^2
        & \leq \frac{1}{b} \sum_{i \in S} \norm{\nabla_{\vtheta}(\textcolor{black}{\mathcal{L}_i} \circ h_i)(\vtheta)}_{2, \infty}^2
    \end{aligned}
    \end{equation}
    And using \eqref{eq:expectation} for $\left(a_i\right)_{i=1}^n=\left(\norm{\nabla_{\vtheta}(\cL_i \circ h_i)(\vtheta)}_{2,\infty}^2\right)_{i=1}^n$,
    \begin{equation}
    \label{eq:bad_ineq}
    \begin{aligned}
        \E_S\left[
        \norm{
        \frac{1}{b} \sum_{i \in S} \nabla_{\vtheta}(\textcolor{black}{\mathcal{L}_i} \circ h_i)(\vtheta)
        }_{2, \infty}^2
        \right]
        & \leq \sum_{i=1}^n\bar{p}_i(\hat{\vL}) \norm{\nabla_{\vtheta}(\textcolor{black}{\mathcal{L}_i} \circ h_i)(\vtheta)}_{2, \infty}^2\\
        & \leq \sum_{i=1}^n
        \frac{1}{\bar{p}_i(\hat{\vL})}
        \max_{l}\norm{\nabla_{\vtheta_l}(\bar{p}_i(\hat{\vL})\textcolor{black}{\mathcal{L}_i} \circ h_i)(\vtheta)}_{2}^2\\
        & \leq \frac{1}{\alpha}
        \sum_{i=1}^n\max_{l}\norm{\nabla_{\vtheta_l}(\bar{p}_i(\hat{\vL})\textcolor{black}{\mathcal{L}_i} \circ h_i)(\vtheta)}_{2}^2
    \end{aligned}
    \end{equation}
\end{itemize}

\paragraph{\textbf{Important Remark}:}\label{rk:important}
It is worth noting in \eqref{eq:bad_ineq} the apparition of $\alpha$ defined in \eqref{eq:alpha}.
If we were using a uniform sampling as for ERM (i.e. for DRO in the limit $\de \rightarrow 0$), we would have $\alpha = \frac{1}{n}$.
So although our inequality \eqref{eq:bad_ineq} may seem crude, it is consistent with equation (13.2) in \citep{allen-zhu19a} and the corresponding inequality in the case of ERM.

The rest of the proof of convergence will consist in proving that $\eta \norm{\nabla_{\vtheta} (R \circ \vL \circ h)(\vtheta)}_{2,2}^2$ dominates the two last terms in \eqref{eq:cv_sgd_exact_1}.
As a result, we can already state that either the robustness parameter $\de$, or the learning rate $\eta$ will have to be small enough to control $\alpha$. 
%

Indeed, combining \eqref{eq:cv_sgd_exact_1} with the chain rule \eqref{eq:reminder_chain_rule_theta}, and the gradient bound Theorem \ref{th:gradient_bound} where we use our $\hat{\vL}$ defined in \eqref{eq:exact_loss_history}
\begin{equation}
\label{eq:cv_sgd_exact_2}
    \begin{aligned}
        \E_S\left[R(\vL(h(\vtheta '))\right]  
        & \leq
        R(\vL(h(\vtheta)) 
        - \Omega\left(\frac{\eta m \delta}{d n^2}\right)
        \sum_{i=1}^n 
            \norm{
            \bar{p}_i(\hat{\vL})\nabla_{h_i}\textcolor{black}{\mathcal{L}_i}(h_i(\vtheta))
            }_{2}^2 \\
        & + \eta
        O\left(\frac{n L^2\omega^{1/3}\sqrt{m\log(m)}}{\sqrt{d}}\right)
        O\left(\sqrt{\frac{m}{d}}\right)
        \sum_{i=1}^n 
            \norm{
            \bar{p}_i(\hat{\vL})\nabla_{h_i}\textcolor{black}{\mathcal{L}_i}(h_i(\vtheta))
            }_{2}^2 \\
        & + \eta^2 O\left(\left(\frac{\de C(\cL)^2}{n \rho}
	    + 2C(\nabla \cL)\right)\frac{n L^2 m}{d}\right) 
	     O\left(\frac{m}{d\alpha}\right)
	    \sum_{i=1}^n 
            \norm{
            \bar{p}_i(\hat{\vL})\nabla_{h_i}\textcolor{black}{\mathcal{L}_i}(h_i(\vtheta))
            }_{2}^2 \\
        & \leq
        R(\vL(h(\vtheta)) 
        - \Omega\left(\frac{\eta m \delta}{d n^2}\right)
        \sum_{i=1}^n 
            \norm{
            \bar{p}_i(\hat{\vL})\nabla_{h_i}\textcolor{black}{\mathcal{L}_i}(h_i(\vtheta))
            }_{2}^2 \\
        & +
        O\left(
        \frac{\eta n L^2 m \omega^{1/3}\sqrt{\log(m)}}{d}
        + K \frac{\eta^2 (n / \alpha) L^2 m^2}{d^2}
        \right)
        \sum_{i=1}^n 
            \norm{
            \bar{p}_i(\hat{\vL})\nabla_{h_i}\textcolor{black}{\mathcal{L}_i}(h_i(\vtheta))
            }_{2}^2 \\
    \end{aligned}
\end{equation}
where we have used
\begin{equation}
    K := \frac{\de C(\cL)^2}{n \rho} + 2C(\nabla \cL)
\end{equation}
There are only two differences \textcolor{black}{compared to} equation (13.2) in \citep{allen-zhu19a}:
\begin{itemize}
    \item in the last fraction we have $n / \alpha$ instead of $n^2$ (see remark \ref{rk:important} for more details), and an additional multiplicative term $K$.
    So in total, this term differs by a multiplicative factor $\frac{\alpha n}{K}$ from the analogous term in the proof of \citep{allen-zhu19a}.
    \item we have $\sum_{i=1}^n 
            \norm{
            \bar{p}_i(\hat{\vL})\nabla_{h_i}\cL_i(h_i(\vtheta))
            }_{2}^2$ instead of $F(\mathbf{W}^{(t)})$. 
            In fact they are analogous since in  equation (13.2) in \citep{allen-zhu19a}, $F(\mathbf{W}^{(t)})$ is the squared norm of the mean loss for the $L^2$ loss.
            We don't make such a strong assumption on the choice of $\cL$ (see assumption \ref{as:3}).
            It is worth noting that the same analogy is used in \citep[Appendix A]{allen-zhu19a} where they extend their result to the mean loss with other objective function than the $L^2$ loss.
\end{itemize}


%
Our choice of learning rate in Theorem \ref{th:conv_sgd_stale_loss_history} can be rewritten as
\begin{equation}
    \begin{aligned}
         \eta_{exact} 
         &= \Theta\left(
        \frac{\alpha n^2 \rho}{\de C(\cL)^2+ 2 n \rho C(\nabla \cL)} 
        \times \frac{b\delta d}{\textup{poly}(n,L)m\log^2(m)}
        \right)\\
        &= \Theta\left(
        \frac{\alpha n}{K} 
        \times \frac{b\delta d}{\textup{poly}(n,L)m\log^2(m)}
        \right)\\
        & \leq \frac{\alpha n}{K} \times \eta'\\
    \end{aligned}
\end{equation}
And we also have
\begin{equation}
    \eta_{exact} \leq \eta'
\end{equation}
%
where $\eta'$ is the learning rate chosen in the proof of Theorem 2 in \citep{allen-zhu19a}.
We refer the reader to \citep{allen-zhu19a} for the details of the constant in "$\Theta$" and the exact form of the polynomial $\textup{poly}(n,L)$.

As a result, for $\eta=\eta_{exact}$, the term $\Omega\left(\frac{\eta m \delta}{d n^2}\right)$ dominates the other term of the right-hand side of inequality \eqref{eq:cv_sgd_exact_2}
as in the proof of Theorem 2 in \citep{allen-zhu19a}.

This implies that the conditions of Theorem \ref{th:gradient_bound} are satisfied for all  $\vtheta^{(t)}$, and that we have for all iteration $t > 0$
\begin{equation}
    \label{eq:cv_sgd_exact_4}
    \E_{S_t}\left[R(\vL(h(\vtheta^{(t+1)}))\right]  
        \leq
        R(\vL(h(\vtheta^{(t)})) 
        - \Omega\left(\frac{\eta m \delta}{d n^2}\right)
        \sum_{i=1}^n 
            \norm{
            \bar{p}_i(\hat{\vL})\nabla_{h_i}\textcolor{black}{\mathcal{L}_i}(h_i(\vtheta^{(t)}))
            }_{2}^2
\end{equation}

And using a result in Appendix A of \citep{allen-zhu19a}, since under assumption \ref{as:3} the distributionally robust loss is non-convex and bounded, we obtain for all $\epsilon'>0$
\begin{equation}
    \norm{\nabla_h (R\circ \vL)(h(\vtheta^{(T)}))}_{2,2} \leq \epsilon'
    \quad \textup{if} \quad
    T = O\left(\frac{d n^2}{\eta \delta m \epsilon'^2}\right)
\end{equation}
where according to \eqref{eq:reminder_chain_rule_h}
\begin{equation}
    \norm{\nabla_h (R\circ \vL)(h(\vtheta^{(T)}))}_{2,2}
    =
    \sum_{i=1}^n 
            \norm{
            \bar{p}_i(\hat{\vL})\nabla_{h_i}\textcolor{black}{\mathcal{L}_i}(h_i(\vtheta^{(t)}))
            }_{2}^2
\end{equation}
However, we are interested in a bound on $\norm{\nabla_{\vtheta} (R\circ \vL \circ h)(\vtheta^{(T)})}_{2,2}$, rather than a bound on $\norm{\nabla_h (R\circ \vL)(h(\vtheta^{(T)}))}_{2,2}$.
Using the gradient bound of Theorem \ref{th:gradient_bound} and the chain rules \eqref{eq:reminder_chain_rule_theta} and \eqref{eq:reminder_chain_rule_h}
\begin{equation}
    \norm{\nabla_{\vtheta} (R\circ \vL \circ h)(\vtheta^{(T)})}_{2,2}
    \leq c_1 \sqrt{\frac{L m n}{d}} \norm{\nabla_h (R\circ \vL)(h(\vtheta^{(T)}))}_{2,2}
\end{equation}
where $c_1 > 0$ is the constant hidden in $O\left(\sqrt{\frac{L m n}{d}}\right)$.

So with $\epsilon' = \frac{1}{c_1}\sqrt{\frac{d}{L m n}}\epsilon$, we finally obtain
\begin{equation}
    \begin{aligned}
        \norm{\nabla_{\vtheta} (R\circ \vL \circ h)(\vtheta^{(T)})}_{2,2}
        &\leq c_1 \sqrt{\frac{L m n}{d}} 
        \norm{\nabla_h (R\circ \vL)(h(\vtheta^{(T)}))}_{2,2}\\
        &\leq c_1 \sqrt{\frac{L m n}{d}} \epsilon'\\
        & \leq \epsilon
    \end{aligned}
\end{equation}
If
\begin{equation}
    T = O\left(\frac{d n^2}{\eta \delta m \epsilon'^2}\right)
    = O\left(\frac{d n^2}{\eta \delta m}\frac{L m n}{d \epsilon^2}\right)
    = O\left(\frac{L n^3}{\eta \delta \epsilon^2}\right)
\end{equation}
which concludes the proof. $\blacksquare$


\subsubsection{Proof of technical lemma 1}\label{p:tech_lemma_1}\ \\
For any integer $k \geq 1$, and all $\left(\va_i\right)_{i=1}^n \in \left(\sR^k\right)^n$, we have 
\begin{equation}
    \begin{aligned}
        \E_S\left[\frac{1}{b}\sum_{i \in S} \va_i\right] 
        &= \sum_{1 \leq i_1,\ldots, i_b \leq n} 
        \left[ 
        \left(\prod_{k=1}^n \bar{p}_{i_k}(\hat{\vL})\right)
        \frac{1}{b}\sum_{j=1}^b \va_{i_j}
        \right]\\
        &= \frac{1}{b} \sum_{1 \leq i_1,\ldots, i_b \leq n} 
        \left[ 
        \sum_{j=1}^b \bar{p}_{i_j}(\hat{\vL})\, \va_{i_j}
        \left(\prod_{\substack{k=1\\ k\neq j}}^n \bar{p}_{i_k}(\hat{\vL})\right)
        \right]\\
        &= \frac{1}{b}\sum_{j=1}^b 
        \left[
        \sum_{1 \leq i_1,\ldots, i_b \leq n}\bar{p}_{i_j}(\hat{\vL})\, \va_{i_j}
        \left(\prod_{\substack{k=1\\ k\neq j}}^n \bar{p}_{i_k}(\hat{\vL})\right)
        \right]\\
        &= \frac{1}{b}\sum_{j=1}^b 
        \left[
        \left(
        \sum_{i_j=1}^n\bar{p}_{i_j}(\hat{\vL})\, \va_{i_j}
        \right)
        \prod_{\substack{k=1\\ k\neq j}}^n
        \left(
        \sum_{i_k=1}^n\bar{p}_{i_k}(\hat{\vL})
         \right)
        \right]\\
        &= \frac{1}{b}\sum_{j=1}^b
        \left(
        \sum_{i=1}^n\bar{p}_{i}(\hat{\vL})\, \va_{i}
        \right)\\
        &= \sum_{i=1}^n\bar{p}_{i}(\hat{\vL})\, \va_{i}\\
        &= \E_{\bar{\vp}(\hat{\vL})}\left[\va_{i}\right]
    \end{aligned}
\end{equation}

\subsection{Convergence of SGD with Hardness Weighted Sampling and stale per-example loss vector}\label{s:proof_stale_loss_history}

The proof of the convergence of \Algref{alg:1} under the conditions of Theorem \ref{th:conv_sgd_stale_loss_history} follows the same structure as the proof of the convergence of Robust SGD with exact per-example loss vector \ref{s:proof_exact_loss_history}.
We will reuse the intermediate results of \ref{s:proof_exact_loss_history} when possible and focus on the differences between the two proofs due to the inexactness of the per-example loss vector.

Let $t$ be the iteration number, and let $\vtheta^{(t)} \in \left(\sR^{m \times m}\right)^L$ be the parameters of the deep neural network at iteration $t$.
We define the stale per-example loss vector at iteration $t$ as
\begin{equation}
    \label{eq:stale_loss_history}
    \hat{\vL}=\left(\textcolor{black}{\mathcal{L}_i}(h_i(\vtheta^{(t_i(t))}))\right)_{i=1}^n
\end{equation}
where for all $i$, $t_i(t) < t$ corresponds to the latest iteration before $t$ at which the per-example loss value for example $i$ has been updated. 
Or equivalently, it corresponds to the last iteration before $t$ when example $i$ was drawn to be part of a mini-batch.

We also define the exact per-example loss vector that is unknown in \Algref{alg:1}, as
\begin{equation}
    \label{eq:ecat_loss_history2}
    \breve{\vL}=\left(\textcolor{black}{\mathcal{L}_i}(h_i(\vtheta^{(t)}))\right)_{i=1}^n
\end{equation}
Similarly to \eqref{eq:next_theta} we define
\begin{equation}
    \label{eq:next_theta2}
    \vtheta^{(t+1)} = \vtheta^{(t)} - \eta \frac{1}{b} \sum_{i \in S} \nabla_{\vtheta}(\textcolor{black}{\mathcal{L}_i} \circ h_i)(\vtheta^{(t)}) 
\end{equation}
and using Theorem \ref{th:semi-smoothness}, similarly to \eqref{eq:cv_sgd_exact_1}, we obtain 
\begin{equation}
\label{eq:cv_sgd_stale_1}
    \begin{aligned}
            R(\vL(h(\vtheta^{(t+1)})) \leq &
            R(\vL(h(\vtheta^{(t)})) 
            - \eta \langle \nabla_{\vtheta} (R \circ \vL \circ h)(\vtheta^{(t)}),\frac{1}{b} \sum_{i \in S} \nabla_{\vtheta}(\textcolor{black}{\mathcal{L}_i} \circ h_i)(\vtheta^{(t)})\rangle \\
            %
            & + \eta \norm{\nabla_h (R \circ \vL)(h(\vtheta^{(t)}))}_{1,2} O\left(\frac{L^2\omega^{1/3}\sqrt{m\log(m)}}{\sqrt{d}}\right)
            \norm{
            \frac{1}{b} \sum_{i \in S} \nabla_{\vtheta}(\textcolor{black}{\mathcal{L}_i} \circ h_i)(\vtheta^{(t)})
            }_{2, \infty}\\
            & + \eta^2 O\left(\left(\frac{\de C(\cL)^2}{n \rho}
	    + 2C(\nabla \cL)\right)\frac{n L^2 m}{d}\right) 
	    \norm{
	    \frac{1}{b} \sum_{i \in S} \nabla_{\vtheta}(\textcolor{black}{\mathcal{L}_i} \circ h_i)(\vtheta^{(t)})
	    }_{2, \infty}^2
        \end{aligned}
\end{equation}
We can still define $\alpha$ as in \eqref{eq:alpha}
\begin{equation}
    \label{eq:alpha2}
    \alpha = \min_{\vtheta} \min_{i} \bar{p}_i(\vL(\vtheta)) > 0
\end{equation}
where we are guaranteed that $\alpha > 0$ under assumptions \ref{as:2}.

Since Theorem \ref{th:gradient_bound} is independent to the choice of $\hat{\vL}$, taking the expectation with respect to $S$, similarly to \eqref{eq:cv_sgd_exact_2}, we obtain

\begin{equation}
\label{eq:cv_sgd_stale_2}
    \begin{aligned}
        \E_S\left[R(\vL(h(\vtheta^{(t+1)}))\right]  
        & \leq
        R(\vL(h(\vtheta^{(t)})) 
        - \eta 
        \langle \nabla_{\vtheta} (R \circ \vL \circ h)(\vtheta^{(t)}),
        \sum_{i=1}^n 
            \bar{p}_i(\hat{\vL})\nabla_{\vtheta}(\textcolor{black}{\mathcal{L}_i} \circ h_i)(\vtheta^{(t)}))
        \rangle \\
        & + \eta
         \norm{
        \nabla_h (R \circ \vL)(h(\vtheta^{(t)}))
        }_{1,2}
        O\left(\frac{L^2\omega^{1/3}\sqrt{nm\log(m)}}{\sqrt{d}}\right)
        \sqrt{
        \sum_{i=1}^n 
            \norm{
            \bar{p}_i(\hat{\vL})\nabla_{h_i}\textcolor{black}{\mathcal{L}_i}(h_i(\vtheta^{(t)}))
            }_{2}^2
        }\\
        & + \eta^2 O\left(\left(\frac{\de C(\cL)^2}{n \rho}
	    + 2C(\nabla \cL)\right)\frac{n L^2 m}{d}\right) 
	     O\left(\frac{m}{d\alpha}\right)
	    \sum_{i=1}^n 
            \norm{
            \bar{p}_i(\hat{\vL})\nabla_{h_i}\textcolor{black}{\mathcal{L}_i}(h_i(\vtheta^{(t)}))
            }_{2}^2 \\
    \end{aligned}
\end{equation}
where the differences with respect to \eqref{eq:cv_sgd_exact_2} comes from the fact that $\hat{\vL}$ is not the exact per-example loss vector here, i.e. $\hat{\vL} \neq \breve{\vL}$,
which leads to
\begin{equation}
    \begin{aligned}
        \nabla_{\vtheta} (R \circ \vL \circ h)(\vtheta^{(t)}) 
        &=
        \sum_{i=1}^n 
            \hat{p}_i(\breve{\vL})\nabla_{\vtheta}(\textcolor{black}{\mathcal{L}_i} \circ h_i)(\vtheta^{(t)}))\\
        & \neq
        \sum_{i=1}^n 
            \bar{p}_i(\hat{\vL})\nabla_{\vtheta}(\textcolor{black}{\mathcal{L}_i} \circ h_i)(\vtheta^{(t)}))
    \end{aligned}
\end{equation}
and
\begin{equation}
    \begin{aligned}
        \norm{
            \nabla_{h} (R \circ \vL)(h(\vtheta^{(t)}))
        }_{1,2}
        &=
        \sum_{i=1}^n
            \norm{
                \hat{p}_i(\breve{\vL})
                \nabla_{h_i}\textcolor{black}{\mathcal{L}_i}(h_i(\vtheta^{(t)})))
            }_2\\
        & \neq
        \sum_{i=1}^n
            \norm{
                \hat{p}_i(\hat{\vL})
                \nabla_{h_i}\textcolor{black}{\mathcal{L}_i}(h_i(\vtheta^{(t)})))
            }_2\\
    \end{aligned}
\end{equation}
Let 
\begin{equation}
    K' = C(\cL) A(\nabla \cL)
    \,O\left(
    \frac{\de L m^{3/2} \log^2(m)}{\alpha n^{1/2} \rho d^{3/2} b \log\left(\frac{1}{1-\alpha}\right)}
    \right)
\end{equation}
Where $C(\cL) > 0$ is a constant such that $\cL$ is $C(\cL)$-Lipschitz continuous,
and $A(\nabla \cL) > 0$ is a constant that bound the gradient of $\cL$ with respect to its input.
$C(\cL)$ and $A(\nabla \cL)$ are guaranteed to exist under assumptions \ref{as:2}.

We can prove that, with probability at least 
$1 - \exp\left(-\Omega\left(\log^2(m)\right)\right)$,
\begin{itemize}
    \item according to lemma~\ref{p:tech_lemma_2}
    \begin{equation}
    \norm{\hat{p}(\hat{\vL}) - \hat{p}(\breve{\vL})}_2
    = 
    \sqrt{
    \sum_{i=1}^n \left(\hat{p}_i(\hat{\vL}) - \hat{p}_i(\breve{\vL})\right)^2
    }
    \leq 
    \eta \alpha K'
    \end{equation}
    \item according to lemma~\ref{p:tech_lemma_3}
    \begin{equation}
        \begin{aligned}
            \left|
            \langle
            \nabla_{\vtheta} (R \circ \vL \circ h)(\vtheta^{(t)})
            - \sum_{i=1}^n 
            \bar{p}_i(\hat{\vL})\nabla_{\vtheta}(\textcolor{black}{\mathcal{L}_i} \circ h_i)(\vtheta^{(t)})),
            \sum_{i=1}^n 
            \bar{p}_i(\hat{\vL})\nabla_{\vtheta}(\textcolor{black}{\mathcal{L}_i} \circ h_i)(\vtheta^{(t)}))
            \rangle
            \right|\\
            \leq 
            \eta \frac{m}{d} K'
            \sum_{i=1}^n
            \norm{
            \bar{p}_i(\hat{\vL})\nabla_{\vtheta}(\textcolor{black}{\mathcal{L}_i} \circ h_i)(\vtheta^{(t)}))}_2^2
        \end{aligned}
    \end{equation}
    \item according to lemma~\ref{p:tech_lemma_4}
    \begin{equation}
        \norm{
            \nabla_{h} (R \circ \vL)(h(\vtheta^{(t)}))
        }_{1,2}
        \leq
        \left(\sqrt{n} + \eta K'\right)
        \sqrt{
            \sum_{i=1}^n
            \norm{
            \bar{p}_i(\hat{\vL})\nabla_{\vtheta}(\textcolor{black}{\mathcal{L}_i} \circ h_i)(\vtheta^{(t)}))}_2^2
        }
    \end{equation}
\end{itemize}
Combining those three inequalities with \eqref{eq:cv_sgd_stale_2} we obtain
\begin{equation}
    \begin{aligned}
        \E_S\left[R(\vL(h(\vtheta^{(t+1)}))\right]& - R(\vL(h(\vtheta^{(t)}))
        \leq
        \\
        \eta& \left[
            - \Omega\left(\frac{m \delta}{d n^2}\right)
            + O\left(
            \frac{n L^2 m \omega^{1/3}\sqrt{\log(m)}}{d}
            \right)
            \right]
            \sum_{i=1}^n 
            \norm{
            \bar{p}_i(\hat{\vL})\nabla_{h_i}\textcolor{black}{\mathcal{L}_i}(h_i(\vtheta^{(t)}))
            }_{2}^2
            \\
        \eta^2 &
            O\left(
            K \frac{(n / \alpha) L^2 m^2}{d^2}
            + \left(1 + \frac{m}{d}\right) K'
            \right)
        \sum_{i=1}^n 
            \norm{
            \bar{p}_i(\hat{\vL})\nabla_{h_i}\textcolor{black}{\mathcal{L}_i}(h_i(\vtheta^{(t)}))
            }_{2}^2
            \\
    \end{aligned}
\end{equation}
One can see that compared to \eqref{eq:cv_sgd_exact_2}, there is only the additional term $\left(1 + \frac{m}{d}\right) K'$.

Using our choice of $\eta$,
\begin{equation}
    \eta = \eta_{stale} \leq O \left(
                        \frac{\delta}{n^2 K'}\eta_{exact}
                    \right)
\end{equation}
where $\eta_{exact}$ is the learning rate of Theorem \ref{th:conv_sgd_exact_loss_history},
we have
\begin{equation}
    \Omega\left(\frac{\eta m \delta}{d n^2}\right) 
    \geq 
    O\left(\eta ^2 \left(1 + \frac{m}{d}\right) K'\right)
\end{equation}
As a result, $\eta^2 \left(1 + \frac{m}{d}\right) K'$ is dominated by the term $\Omega\left(\frac{\eta m \delta}{d n^2}\right)$ 

In addition, since $\eta_{stale} \leq \eta_{exact}$, $\Omega\left(\frac{\eta m \delta}{d n^2}\right)$ still dominates also the ther terms as in the proof of Theorem \ref{th:conv_sgd_exact_loss_history}.

As a consequence, we obtain as in \eqref{eq:cv_sgd_exact_4} that for any iteration $t > 0$
\begin{equation}
    \label{eq:cv_sgd_stale_4}
    \E_{S_t}\left[R(\vL(h(\vtheta^{(t+1)}))\right]
        \leq
        R(\vL(h(\vtheta^{(t)})) 
        - \Omega\left(\frac{\eta m \delta}{d n^2}\right)
        \sum_{i=1}^n 
            \norm{
            \bar{p}_i(\hat{\vL})\nabla_{h_i}\textcolor{black}{\mathcal{L}_i}(h_i(\vtheta^{(t)}))
            }_{2}^2
\end{equation}
This concludes the proof using the same arguments as in the end of the proof of Theorem \ref{th:conv_sgd_exact_loss_history} starting from \eqref{eq:cv_sgd_exact_4}. $\blacksquare$

\subsubsection{Proof of technical lemma 2}\label{p:tech_lemma_2}\ \\
Using Lemma~\ref{lemma:robust_loss_sg} and Lemma~\ref{lemma:R_property} we obtain
\begin{equation}
    \begin{aligned}
        \norm{\hat{\vp}(\hat{\vL}) - \hat{\vp}(\breve{\vL})}_2
        &= 
        \norm{\nabla_v R(\hat{\vL}) - \nabla_v R(\breve{\vL})}_2\\
        & \leq \frac{\de}{n \rho} \norm{\hat{\vL} - \breve{\vL}}_2
    \end{aligned}
\end{equation}
Using assumptions~\ref{as:3} and \citep[Claim 11.2]{allen-zhu19a}
\begin{equation}
    \begin{aligned}
        \norm{\hat{\vp}(\hat{\vL}) - \hat{\vp}(\breve{\vL})}_2
        &\leq 
        \frac{\de}{n \rho} 
        \sqrt{
        \sum_{i=1}^n \left(\textcolor{black}{\mathcal{L}_i}\circ~h_i(\vtheta^{(t)}) - \textcolor{black}{\mathcal{L}_i}\circ~h_i(\vtheta^{(t_i(t))})\right)^2
        }\\
        & \leq 
        \frac{\de}{n \rho}
        C(\cL)C(h)
        \sqrt{
        \sum_{i=1}^n \norm{\vtheta^{(t)} - \vtheta^{(t_i(t))}}_{2,2}^2
        }\\
        & \leq 
        C(\cL)
        O\left(\frac{\de L m^{1/2}}{n \rho d^{1/2}}\right)
        \sqrt{
        \sum_{i=1}^n \norm{\vtheta^{(t)} - \vtheta^{(t_i(t))}}_{2,2}^2
        }\\
    \end{aligned}
\end{equation}
Where $C(\cL)$ is the constant of Lipschitz continuity of the per-example loss $\cL$ (see assumptions~\ref{as:3}) and $C(h)$ is the constant of Lipschitz continuity of the deep neural network $h$ with respect to its parameters $\vtheta$.

By developing the recurrence formula of $\vtheta^{(t)}$~\eqref{eq:next_theta2}, we obtain
\begin{equation*}
    \begin{aligned}
        \norm{\hat{\vp}(\hat{\vL}) - \hat{\vp}(\breve{\vL})}_2
        & \leq 
        C(\cL)
        O\left(\frac{\de L m^{1/2}}{n \rho d^{1/2}}\right)
        \sqrt{
        \sum_{i=1}^n \norm{
        \vtheta^{(t_i(t))} 
        - \left(\sum_{\tau = t_i(t)}^{t-1} \frac{\eta}{b} \sum_{j \in S_{\tau}} \nabla_{\vtheta} (\textcolor{black}{\mathcal{L}_j} \circ~h_j)(\vtheta^{(\tau)})
        \right)
        - \vtheta^{(t_i(t))}
        }_{2,2}^2
        }\\
         & \leq 
         \eta
        C(\cL)
        O\left(\frac{\de L m^{1/2}}{n \rho d^{1/2}}\right)
        \sqrt{
        \sum_{i=1}^n \norm{
        \sum_{\tau = t_i(t)}^{t-1} \frac{1}{b} \sum_{j \in S_{\tau}} \nabla_{\vtheta} (\textcolor{black}{\mathcal{L}_j} \circ~h_j)(\vtheta^{(\tau)})
        }_{2,2}^2
        }\\
    \end{aligned}
\end{equation*}
Let $A(\nabla \cL)$ a bound on the gradient of the per-example loss function.
Using Theorem~\ref{th:gradient_bound} and the chain rule
\begin{equation}
    \forall j,\, \forall \tau \quad 
    \norm{
    \nabla_{\vtheta} (\textcolor{black}{\mathcal{L}_j} \circ h_j)(\vtheta^{(\tau)})
    }_{2,2}
    \leq 
    A(\nabla \cL) O \left( \frac{m}{d}\right)
\end{equation}
And using the \textcolor{black}{triangle} inequality
\begin{equation}
    \begin{aligned}
        \norm{
        \sum_{\tau = t_i(t)}^{t-1} \frac{1}{b} \sum_{j \in S_{\tau}} \nabla_{\vtheta} (\textcolor{black}{\mathcal{L}_j} \circ h_j)(\vtheta^{(\tau)})
        }_{2,2}
        &\leq
        \sum_{\tau = t_i(t)}^{t-1} \frac{1}{b} \sum_{j \in S_{\tau}} \norm{\nabla_{\vtheta} (\textcolor{black}{\mathcal{L}_j} \circ h_j)(\vtheta^{(\tau)})}_{2,2}\\
        & \leq
        \sum_{\tau = t_i(t)}^{t-1} A(\nabla \cL) O \left( \frac{m}{d}\right)\\
        & \leq 
        A(\nabla \cL) O \left( \frac{m}{d}\right) (t - t_i(t))
    \end{aligned}
\end{equation}
As a result, we obtain
\begin{equation}
    \begin{aligned}
        \norm{\hat{\vp}(\hat{\vL}) - \hat{\vp}(\breve{\vL})}_2
         & \leq 
         \eta
        C(\cL)
        A(\nabla \cL)
        O\left(\frac{\de L m^{3/2}}{n \rho d^{3/2}}\right)
        \sqrt{
        \sum_{i=1}^n (t - t_i(t))^2
        }\\
    \end{aligned}
\end{equation}
For all $i$ and for any $\tau$ the probability that the sample $i$ is not in batch $S_{\tau}$ is lesser than 
$\left(1 - \alpha\right)^b$.

Therefore, for any $k \geq 1$ and for any $t$,
\begin{equation}
    P\left(t - t_i(t) \geq k\right) \leq \left(1 - \alpha\right)^{kb}
\end{equation}
For 
$k\geq \frac{1}{b}\Omega\left(\frac{\log^2(m)}{\log\left(\frac{1}{1-\alpha}\right)}\right)$,
we have
$\left(1 - \alpha\right)^{kb} \leq \exp\left(-\Omega\left(\log^2(m)\right)\right)$,
and thus with probability at least 
$1 - \exp\left(-\Omega\left(\log^2(m)\right)\right)$,
\begin{equation}
    \forall t,\quad t - t_i(t) \leq O\left(\frac{\log^2(m)}{b\log\left(\frac{1}{1-\alpha}\right)}\right)
\end{equation}
As a result, we finally obtain that with probability at least 
$1 - \exp\left(-\Omega\left(\log^2(m)\right)\right)$,
\begin{equation}
    \begin{aligned}
        \norm{\hat{\vp}(\hat{\vL}) - \hat{\vp}(\breve{\vL})}_2
         & \leq 
         \eta
        C(\cL)
        A(\nabla \cL)
        O\left(\frac{\de L m^{3/2}}{n \rho d^{3/2}}\right)
        \sqrt{n}
        O\left(\frac{\log^2(m)}{b\log\left(\frac{1}{1-\alpha}\right)}\right)
        \\
        & \leq
        \eta \alpha 
        O\left(
        \frac{\de L m^{3/2} \log^2(m)}{\alpha n^{1/2} \rho d^{3/2} b \log\left(\frac{1}{1-\alpha}\right)}
        \right)\\
        & \leq
        \eta \alpha K'
    \end{aligned}
\end{equation}

\subsubsection{Proof of technical lemma 3}\label{p:tech_lemma_3}\ \\
Let us first denote
\begin{equation}
    \begin{aligned}
        A &= \left|
            \langle
            \nabla_{\vtheta} (R \circ \vL \circ h)(\vtheta^{(t)})
            - \sum_{i=1}^n 
            \bar{p}_i(\hat{\vL})\nabla_{\vtheta}(\textcolor{black}{\mathcal{L}_i} \circ h_i)(\vtheta^{(t)})),
            \sum_{i=1}^n 
            \bar{p}_i(\hat{\vL})\nabla_{\vtheta}(\textcolor{black}{\mathcal{L}_i} \circ h_i)(\vtheta^{(t)}))
            \rangle
            \right|\\
        &=
        \left|
            \langle
            \sum_{i=1}^n
            \left(
                \bar{p}_i(\breve{\vL})
                - \bar{p}_i(\hat{\vL})
            \right)
            \nabla_{\vtheta}(\textcolor{black}{\mathcal{L}_i} \circ h_i)(\vtheta^{(t)})),
            \sum_{i=1}^n 
            \bar{p}_i(\hat{\vL})\nabla_{\vtheta}(\textcolor{black}{\mathcal{L}_i} \circ h_i)(\vtheta^{(t)}))
            \rangle
            \right|\\
    \end{aligned}
\end{equation}
Using Cauchy-Schwarz inequality
\begin{equation}
    \begin{aligned}
        A 
        &=
        \left|
        \sum_{i=1}^n
        \left(
            \bar{p}_i(\breve{\vL})
            - \bar{p}_i(\hat{\vL})
        \right)
        \langle
            \nabla_{\vtheta}(\textcolor{black}{\mathcal{L}_i} \circ h_i)(\vtheta^{(t)})),
            \sum_{j=1}^n 
            \bar{p}_j(\hat{\vL})\nabla_{\vtheta}(\textcolor{black}{\mathcal{L}_j} \circ h_j)(\vtheta^{(t)}))
        \rangle
        \right|\\
        &\leq 
        \norm{\hat{\vp}(\hat{\vL}) - \hat{\vp}(\breve{\vL})}_2
        \sqrt{
            \sum_{i=1}^n\left(
                \langle
                \nabla_{\vtheta}(\textcolor{black}{\mathcal{L}_i} \circ h_i)(\vtheta^{(t)})),
                \sum_{j=1}^n 
                \bar{p}_j(\hat{\vL})\nabla_{\vtheta}(\textcolor{black}{\mathcal{L}_j} \circ h_j)(\vtheta^{(t)}))
                \rangle
            \right)^2
        }
    \end{aligned}
\end{equation}
Let 
\begin{equation}
    B = 
        \langle
            \nabla_{\vtheta}(\textcolor{black}{\mathcal{L}_i} \circ h_i)(\vtheta^{(t)})),
            \sum_{j=1}^n 
            \bar{p}_j(\hat{\vL})\nabla_{\vtheta}(\textcolor{black}{\mathcal{L}_j} \circ h_j)(\vtheta^{(t)}))
        \rangle
\end{equation}
Using again Cauchy-Schwarz inequality
\begin{equation}
    B
    \leq 
    \norm{\nabla_{\vtheta}(\textcolor{black}{\mathcal{L}_i} \circ h_i)(\vtheta^{(t)}))}_{2,2}
    \norm{\sum_{j=1}^n 
            \bar{p}_j(\hat{\vL})\nabla_{\vtheta}(\textcolor{black}{\mathcal{L}_j} \circ h_j)(\vtheta^{(t)}))}_{2,2}
\end{equation}
As a result, $A$ becomes
\begin{equation}
    \begin{aligned}
        A
        & \leq
        \norm{\hat{\vp}(\hat{\vL}) - \hat{\vp}(\breve{\vL})}_2
        \norm{\sum_{j=1}^n 
            \bar{p}_j(\hat{\vL})\nabla_{\vtheta}(\textcolor{black}{\mathcal{L}_j} \circ h_j)(\vtheta^{(t)}))
        }_{2,2}
        \sqrt{
            \sum_{i=1}^n
                \norm{\nabla_{\vtheta}(\textcolor{black}{\mathcal{L}_i} \circ h_i)(\vtheta^{(t)}))}_{2,2}^2
        }\\
        & \leq
        \norm{\hat{\vp}(\hat{\vL}) - \hat{\vp}(\breve{\vL})}_2
        \norm{\sum_{j=1}^n 
            \bar{p}_j(\hat{\vL})\nabla_{\vtheta}(\textcolor{black}{\mathcal{L}_j} \circ h_j)(\vtheta^{(t)}))
        }_{2,2}
        \sqrt{
            \sum_{i=1}^n
                \frac{1}{\alpha^2}\norm{\bar{p}_j(\hat{\vL})\nabla_{\vtheta}(\textcolor{black}{\mathcal{L}_i} \circ h_i)(\vtheta^{(t)}))}_{2,2}^2
        }\\
        & \leq
        \frac{1}{\alpha}
        \norm{\hat{\vp}(\hat{\vL}) - \hat{\vp}(\breve{\vL})}_2
        \norm{\sum_{j=1}^n 
            \bar{p}_j(\hat{\vL})\nabla_{\vtheta}(\textcolor{black}{\mathcal{L}_j} \circ h_j)(\vtheta^{(t)}))
        }_{2,2}^2
    \end{aligned}
\end{equation}
Using the triangular inequality, Theorem~\ref{th:gradient_bound}, and Lemma~\ref{p:tech_lemma_2}, we finally obtain
\begin{equation}
    \begin{aligned}
        A
        & \leq
        \frac{m}{\alpha d}
        \norm{\hat{\vp}(\hat{\vL}) - \hat{\vp}(\breve{\vL})}_2
        \sum_{j=1}^n 
            \norm{\bar{p}_j(\hat{\vL})\nabla_{h_j}\textcolor{black}{\mathcal{L}_j} (h_j(\vtheta^{(t)}))
        }_{2,2}^2\\
        &\leq
        \eta 
        \frac{m}{d}
        K'
        \sum_{j=1}^n 
            \norm{\bar{p}_j(\hat{\vL})\nabla_{h_j}\textcolor{black}{\mathcal{L}_j} (h_j(\vtheta^{(t)}))
        }_{2,2}^2
    \end{aligned}
\end{equation}

\subsubsection{Proof of technical lemma 4}\label{p:tech_lemma_4}\ \\
We have
\begin{equation}
    \begin{aligned}
        \norm{
            \nabla_{h} (R \circ \vL)(h(\vtheta^{(t)}))
        }_{1,2}
        &=
        \sum_{j=1}^n
            \bar{p}_j(\breve{\vL})
            \norm{\nabla_{h_j}\textcolor{black}{\mathcal{L}_j} (h_j(\vtheta^{(t)}))
        }_{2,2}\\
        &=
        \sum_{j=1}^n
            \bar{p}_j(\hat{\vL})
            \norm{\nabla_{h_j}\textcolor{black}{\mathcal{L}_j} (h_j(\vtheta^{(t)}))
        }_{2,2}\\
        &
        + \sum_{j=1}^n
            \left(\frac{\bar{p}_j(\breve{\vL}) - \bar{p}_j(\hat{\vL})}{\bar{p}_j(\hat{\vL})}\right)
            \bar{p}_j(\hat{\vL})
            \norm{\nabla_{h_j}\textcolor{black}{\mathcal{L}_j} (h_j(\vtheta^{(t)}))
        }_{2,2}\\
    \end{aligned}
\end{equation}
Using Cauchy-Schwarz inequality
\begin{equation}
    \begin{aligned}
        \norm{
            \nabla_{h} (R \circ \vL)(h(\vtheta^{(t)}))
        }_{1,2}
        &\leq
        \left(
        \sqrt{n} +
        \sqrt{
        \sum_{j=1}^n
            \left(\frac{\bar{p}_j(\breve{\vL}) - \bar{p}_j(\hat{\vL})}{\bar{p}_j(\hat{\vL})}\right)^2
            }
        \right)
        \sqrt{
        \sum_{j=1}^n
            \norm{\bar{p}_j(\hat{\vL})\nabla_{h_j}\textcolor{black}{\mathcal{L}_j} (h_j(\vtheta^{(t)}))
            }_{2,2}^2
        }\\
    \end{aligned}
\end{equation}
Using Lemma~\ref{p:tech_lemma_2}
\begin{equation}
    \begin{aligned}
        \sum_{j=1}^n
            \left(\frac{\bar{p}_j(\breve{\vL}) - \bar{p}_j(\hat{\vL})}{\bar{p}_j(\hat{\vL})}\right)^2
        & \leq 
        \frac{1}{\alpha}
        \norm{\hat{\vp}(\hat{\vL}) - \hat{\vp}(\breve{\vL})}_2\\
        & \leq
        \eta K'
    \end{aligned}
\end{equation}
Therefore, we finally obtain
\begin{equation}
    \begin{aligned}
        \norm{
            \nabla_{h} (R \circ \vL)(h(\vtheta^{(t)}))
        }_{1,2}
        &\leq
        \left(
        \sqrt{n} + \eta K'
        \right)
        \sqrt{
        \sum_{j=1}^n
            \norm{\bar{p}_j(\hat{\vL})\nabla_{h_j}\textcolor{black}{\mathcal{L}_j} (h_j(\vtheta^{(t)}))
            }_{2,2}^2
        }\\
    \end{aligned}
\end{equation}

\end{document}